%% file: paper.tex
\documentclass[10pt,journal,compsoc,hidelinks]{IEEEtran}
\pdfoutput=1
\ifCLASSOPTIONcompsoc
\usepackage[nocompress]{cite}
\else
\usepackage{cite}
\fi
                                 
\usepackage{amsmath,amsfonts,amssymb} 
\usepackage{subcaption}
\usepackage{algorithm}
\usepackage{algorithmic}
\usepackage{float}
\usepackage{caption}
\usepackage{balance}
\usepackage{graphicx,import}
\usepackage{colortbl}
\usepackage{tabularx}
\usepackage{bm}
\usepackage{booktabs}
\usepackage{textcomp}
\usepackage{pifont}
\usepackage{array,multirow,graphicx}

\definecolor{col1}{RGB}{232, 161, 148}
\definecolor{col2}{RGB}{148, 187, 232}
\definecolor{mygreen}{RGB}{0, 180, 0}
\definecolor{myred}{RGB}{200, 0, 0}

\newcommand{\cmark}{\ding{51}}
\newcommand{\xmark}{\ding{55}}
\newcommand{\bd}[1]{\mathbf{#1}}

\newcommand\mypara[1]{\vspace{1mm}\noindent\textbf{#1}}

\usepackage{tikz}

\begin{document}
	
\title{
	GenDepth: Generalizing Monocular Depth Estimation for Arbitrary Camera Parameters via Ground Plane Embedding
}

\author{Karlo Koledić, Luka Petrović, Ivan Petrović, Ivan Marković$^{1}$%
	\thanks{$^{1}$Authors are with University of Zagreb Faculty of Electrical Engineering and Computing, Laboratory for Autonomous Systems and Mobile Robotics, Zagreb, Croatia	{\tt\small \{name.surname@fer.hr\}. \newline}
		This research has been funded by the H2020 project AIFORS under Grant Agreement No 952275.
		}.
}

\IEEEtitleabstractindextext{%
\begin{abstract}
	Learning-based monocular depth estimation leverages geometric priors present in the training data to enable metric depth perception from a single image, a traditionally ill-posed problem. However, these priors are often specific to a particular domain, leading to limited generalization performance on unseen data. Apart from the well studied environmental domain gap, monocular depth estimation is also sensitive to the domain gap induced by varying camera parameters, an aspect that is often overlooked in current state-of-the-art approaches. This issue is particularly evident in autonomous driving scenarios, where datasets are typically collected with a single vehicle-camera setup, leading to a bias in the training data due to a fixed perspective geometry. In this paper, we challenge this trend and introduce GenDepth, a novel model capable of performing metric depth estimation for arbitrary vehicle-camera setups. To address the lack of data with sufficiently diverse camera parameters, we first create a bespoke synthetic dataset collected with different vehicle-camera systems. Then, we design GenDepth to simultaneously optimize two objectives: (i) equivariance to the camera parameter variations on synthetic data, (ii) transferring the learned equivariance to real-world environmental features using a single real-world dataset with a fixed vehicle-camera system. To achieve this, we propose a novel embedding of camera parameters as the ground plane depth and present a novel architecture that integrates these embeddings with adversarial domain alignment. We validate GenDepth on several autonomous driving datasets, demonstrating its state-of-the-art generalization capability for different vehicle-camera systems.
\end{abstract}

}

\maketitle

\IEEEdisplaynontitleabstractindextext

\IEEEraisesectionheading{\section{Introduction}}
\label{sec:intro}
\input{sections/introduction}

\section{Related Work}
\label{sec:work}
\input{sections/related_work}

\section{Challenges of camera parameter variations}
\label{sec:params}
\input{sections/parameters_effects}

\section{Method}
\label{sec:method}
\input{sections/method}

\section{Experimental evaluation}
\label{sec:experiments}
\input{sections/experiments}

\section{Conclusion and future work}
\label{sec:conclusion}
\input{sections/conclusion}
\bibliographystyle{IEEEtran}

\balance
\bibliography{IEEEabrv.bib,library.bib}

\section*{Appendix}
\input{sections/appendix.tex}

\end{document}

%% file: sections/introduction.tex
\IEEEPARstart{A}{ccurate} 3D perception is one of the fundamental challenges in computer vision, with numerous applications in fields such as robotics, virtual reality and autonomous driving. 
Due to the depth ambiguity of the monocular camera systems, traditional approaches, including Structure-from-Motion or Visual SLAM, estimate 3D structure via subsequent correspondence matching and triangulation. 
However, these solutions are prone to errors in challenging conditions such as occlusions or textureless regions and also require a complicated setup with multi-sensor calibration.

To remedy these issues, learning-based Monocular Depth Estimation (MDE) models use a large amount of data to directly regress per-pixel depth from a single image, usually by exploiting learnable semantic and geometric cues, such as the size of known objects or their position on the ground plane.
These models are consequently heavily reliant upon environments and objects present in the training data. 
Given that, supervised methods \cite{eigen2014depth, laina2016deeper, cs2018depthnet, mancini2017toward, li2015depth, yin2019enforcing, yuan2022newcrfs, bhat2021adabins, agarwal2023attention}, which require expensive and rigorously obtainable ground-truth data, are often limited to a narrow distribution of environments, thereby adversely affecting the generalization performance.
On the other hand, self-supervised methods \cite{godard2019digging, zhan2018unsupervised, li2018undeepvo, zhou_diffnet, guizilini20203d, wang2023planedepth, zhao2022monovit, zhou2017unsupervised} offer a promising alternative, by constructing a supervision signal via view synthesis of nearby frames. This alleviates the issue of ground-truth data acquisition and enables scalable deployment across diverse environments and scenarios.

\begin{figure*}[!t]
	\centering
	\input{figures/intro/intro_fig.tex}
	\caption{Inferred depth maps and corresponding error maps when evaluated on datasets unseen during training. Traditional methods such as Monodepth2 \cite{godard2019digging} and iDisc\cite{piccinelli2023idisc} overfit to the perspective geometry bias in the training data, resulting in poor performance for images captured with different vehicle-camera setups. In contrast, \textbf{GenDepth estimates the accurate \textbf{metric depth} for arbitrary camera parameters without retraining, fine-tuning or post-processing, using only a single real-world dataset without ground-truth labels}.}
	\label{fig:intro}
\end{figure*}
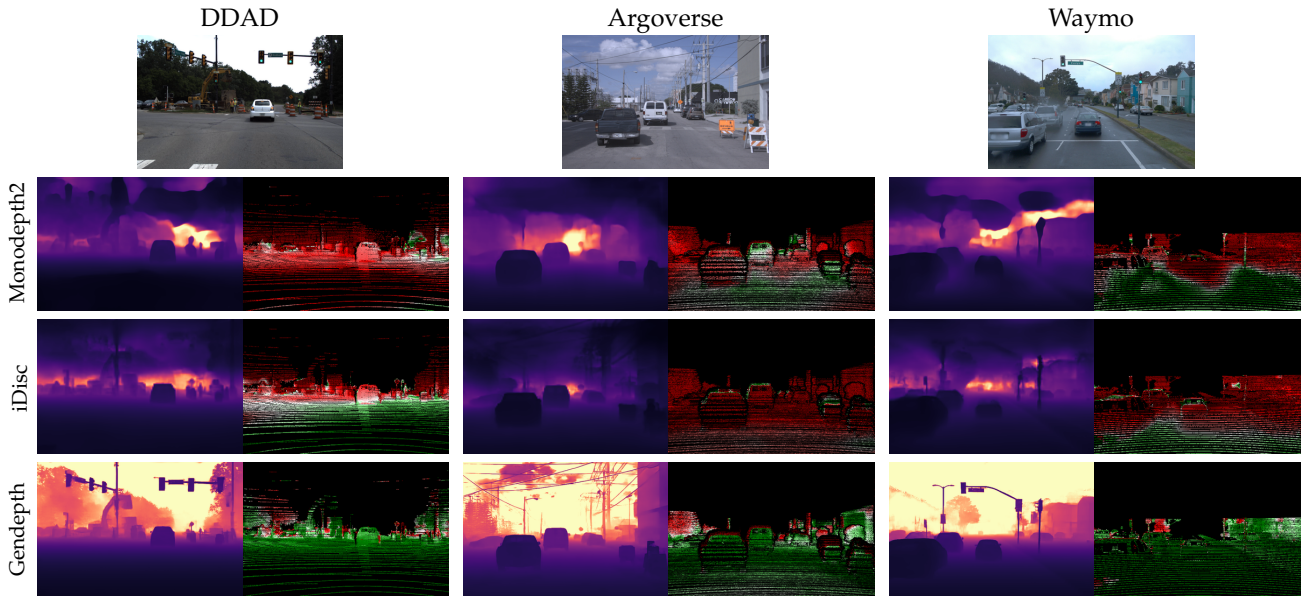

Self-supervised learning simplifies the gathering of data in various environments. However, the impact of different extrinsic and intrinsic camera parameters is frequently overlooked during testing. 
This issue is particularly evident in autonomous driving scenarios. In such cases, datasets are often collected with a single vehicle-camera setup, resulting in all images being captured from the same view relative to the ground plane. 
Models naturally overfit to this perspective geometry bias in the training data and exhibit critical performance degradation when inferring depth for images captured with different vehicle-camera setups \cite{koledic2023moft}.

The generation of synthetic data with diverse camera parameters, along with embedding of camera extrinsics \cite{zhao2021camera}, intrinsics \cite{facil2019cam} or focal length \cite{he2018learning}, has been proposed to solve this issue. 
However, these methods do not provide solutions that are applicable to a real-world scenario with an arbitrary vehicle-camera setup. 
Another line of work focuses on relative depth estimation \cite{chen2016single, yin2021learning, ranftl2020towards, xian2018monocular, yin2020diversedepth, wang2019web, xian2020structure, li2018megadepth}, which usually uses large-scale diverse stereo image or video data from the internet. 
Such an approach facilitates generalization, but abstracts away camera parameters during training, hence enabling depth estimation only up to an unknown scale and shift. 
In addition, these methods rely entirely on the relative object size cue and ignore the constraint between the fixed camera system and the ground plane, which is present in autonomous driving scenarios.

We believe that generalization for different vehicle-camera setups is an indispensable requirement for the scalable and applicable deployment of MDE in autonomous driving. 
Without such a capability, the models would need to be retrained or fine-tuned whenever a vehicle-camera setup is changed compared to the one used in training. 
This could either be due to the use of a camera with different intrinsic parameters or a discrepancy in the extrinsic parameters due to the mounting on the new vehicle model. 
In addition, the parameters can also change during vehicle operation due to physical shocks, impacts or extreme temperature fluctuations.

Motivated by the lack of MDE methods that possess such generalization capability, in this paper we propose GenDepth, a monocular depth estimation method specifically targeted for generalization to arbitrary vehicle-camera systems.
The GenDepth method enables generation of accurate metric depths across various real-world datasets with different vehicle-camera setups without the need for any form of retraining, fine-tuning or post-processing.
To the best of our knowledge, it is the first method with such capabilities, thereby advancing the scalability and applicability of MDE in autonomous driving.
To derive the GenDepth method we first thoroughly analyze the effects of camera parameter variation and identify the most plausible variations in vehicle-camera setup which can degrade MDE performance. 
Then, due to the shortage of data with sufficiently diverse camera parameters, we create a bespoke dataset in the CARLA simulator \cite{dosovitskiy2017carla}, which enables us to learn accurate depth for numerous vehicle-camera setups in the synthetic environment. 
Naturally, due to the significant sim2real gap, the performance of such a model trained exclusively on synthetic images degrades substantially when confronted with real-world data. 
To that end, we introduce a novel architecture that carefully incorporates camera parameters embeddings in conjunction with adversarial feature alignment.
We show that the GenDepth method learns depth that is equivariant to the variations of the camera parameters on synthetic data, while simultaneously adapting to the environmental features of the target real-world data. As visualized in Fig. \ref{fig:intro}, this results in accurate metric depth estimation for several real-world datasets unseen during training.

To summarize, in this paper we present the following contributions:
\begin{itemize}
	\item an in-depth analysis and discussion of the effect of variations in camera parameters on MDE,
	\item a large-scale annotated autonomous driving dataset collected in the CARLA simulator, which includes a wide distribution of vehicle-camera setups,
	\item a novel embedding of camera parameters, designed to enhance generalization and learn domain invariant features,
	\item a novel architecture with adversarial feature alignment, which in conjunction with careful incorporation of camera parameters embeddings, enables domain generalization to the real-world datasets unseen during training,
	\item a thorough evaluation on numerous autonomous driving datasets, which validates our approach and exhibits the ability of our method to estimate metric depth for arbitrary vehicle-camera setups.
\end{itemize}

%% file: figures/intro/intro_fig.tex
\newcommand{\newwidth}{0.15\textwidth}
\centering

\begin{tabular}{@{\hskip 1mm}c@{\hspace {\dimexpr 0.075\textwidth + 1mm}}c@{\hspace {\dimexpr 0.15\textwidth + 2mm}}c@{\hspace {\dimexpr 0.15\textwidth + 2mm}}c@{\hskip 0.075\textwidth}}	
	& DDAD & Argoverse & Waymo \\ 
	
&
	\includegraphics[width=\newwidth]{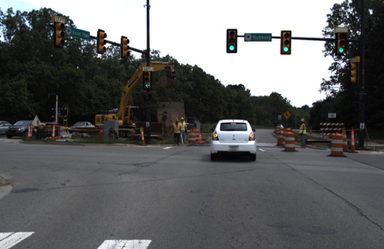}&
	\includegraphics[width=\newwidth]{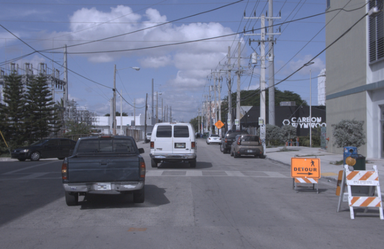}&
	\includegraphics[width=\newwidth]{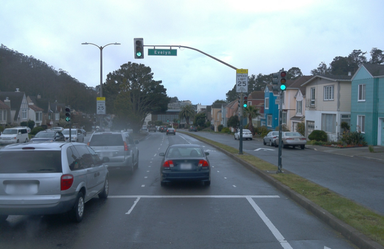} \\
\end{tabular}

\begin{tabular}{@{\hskip 0mm}c@{\hskip 1mm}c@{\hskip 0mm}c@{\hskip 2mm}c@{\hskip 0mm}c@{\hskip 2mm}c@{\hskip 0mm}c@{\hskip 0mm}}	
	{\rotatebox{90}{\hspace{1mm}\footnotesize Monodepth2}} &
	\includegraphics[width=\newwidth]{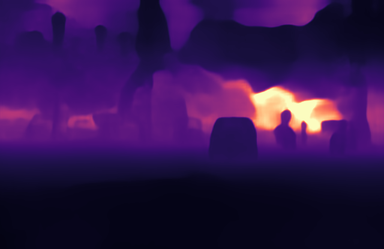}&
	\includegraphics[width=\newwidth]{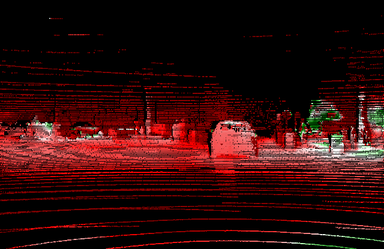}&
	\includegraphics[width=\newwidth]{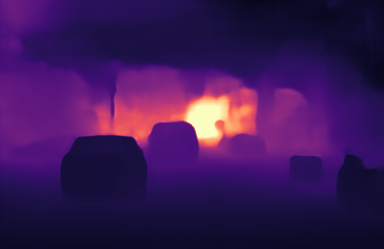}&
	\includegraphics[width=\newwidth]{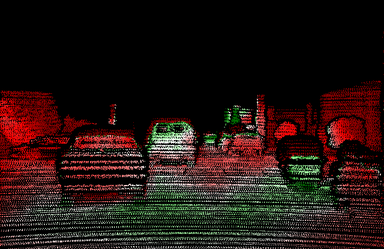}&
	\includegraphics[width=\newwidth]{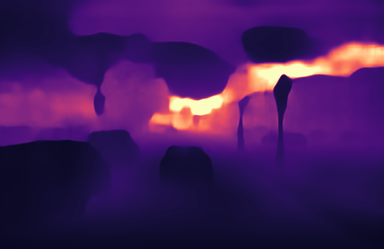}&
	\includegraphics[width=\newwidth]{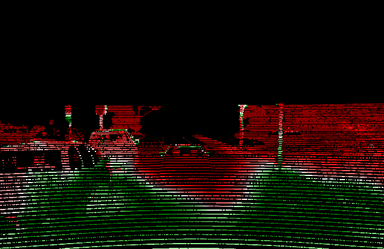} \\
	
	{\rotatebox{90}{\hspace{5.5mm}\footnotesize iDisc}}&
\includegraphics[width=\newwidth]{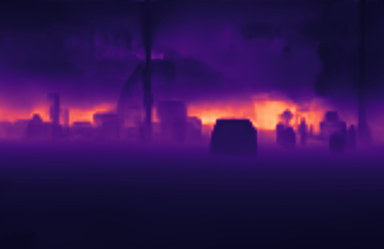}&
\includegraphics[width=\newwidth]{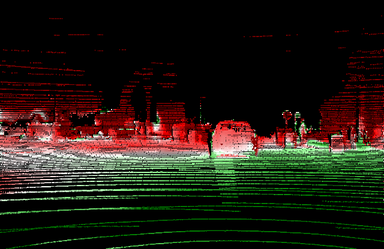}&
\includegraphics[width=\newwidth]{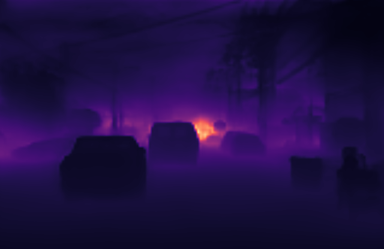}&
\includegraphics[width=\newwidth]{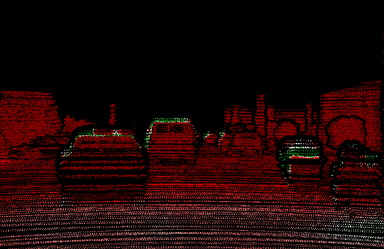}&
\includegraphics[width=\newwidth]{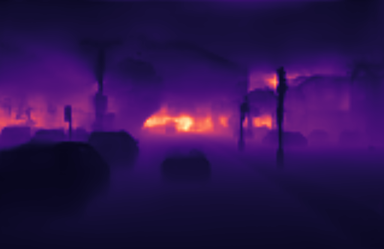}&
\includegraphics[width=\newwidth]{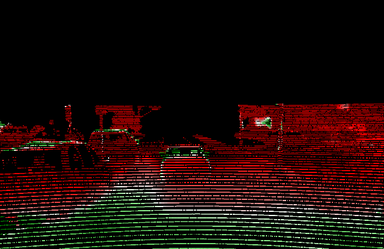} \\	

	{\rotatebox{90}{\hspace{2.5mm}\footnotesize Gendepth}}&
	\includegraphics[width=\newwidth]{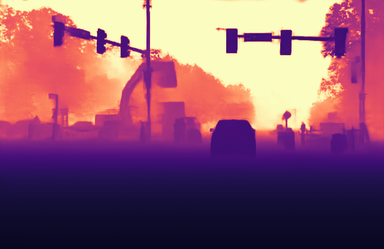}&
	\includegraphics[width=\newwidth]{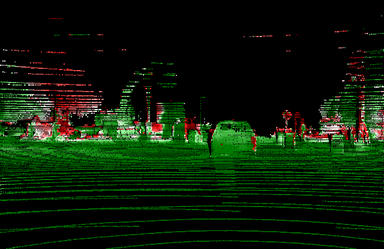}&
	\includegraphics[width=\newwidth]{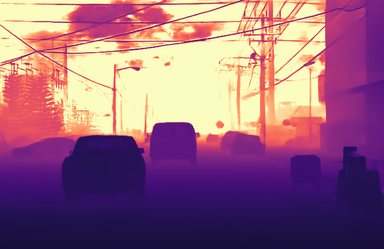}&
	\includegraphics[width=\newwidth]{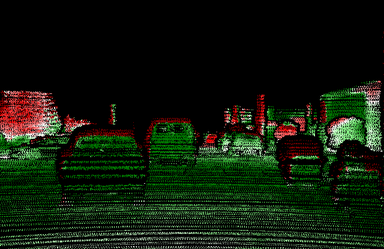}&
	\includegraphics[width=\newwidth]{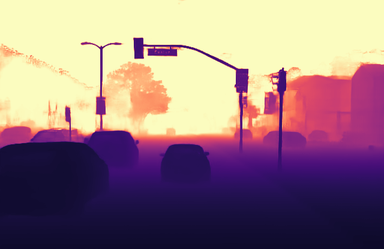}&
	\includegraphics[width=\newwidth]{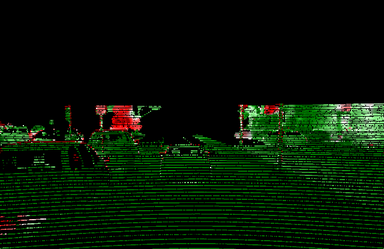} \\
	
\end{tabular}

%% file: sections/related_work.tex
\subsection{Monocular depth estimation}
Learning based monocular depth estimation was initially proposed as a dense estimation task, trained with supervised loss via ground-truth data obtained by reprojection of LiDAR scans \cite{eigen2014depth}. 
Inspired by the popularity of U-net-like architectures in semantic segmentation \cite{ronneberger2015u}, many works have incorporated a similar encoder-decoder architecture with skip connections \cite{laina2016deeper, godard2019digging, li2018undeepvo, zhou2017unsupervised}. 
Additionally, the inclusion of conditional random fields \cite{xu2018structured, li2015depth, liu2015deep, yuan2022newcrfs}, recurrent neural networks \cite{cs2018depthnet, mancini2017toward}, and adversarial training \cite{jung2017depth, lore2018generative} has been proposed to improve the accuracy.
Recently, with increased usage of transformers in computer vision \cite{dosovitskiy2020image}, many works have integrated self-attention into the architecture \cite{bhat2021adabins, agarwal2023attention, li2022depthformer, piccinelli2023idisc}, enabling the modeling of long-range dependencies and thus being naturally complementary to the locality of convolutions.
Finally, foundation models, which learn both low-level and high-level visual concepts from large-scale internet data, have been successfully applied to the problem of monocular depth estimation \cite{zhao2023unleashing, oquab2023dinov2}, achieving state-of-the-art results.

A fundamental issue with supervised methods is the requirement of ground-truth depth data, which necessitates an expensive sensor setup with challenging multi-sensor calibration. 
To improve scalability, self-supervised methods construct the training signal from multiple images of spatially overlapping camera frustums \cite{godard2019digging, zhan2018unsupervised, li2018undeepvo, zhou_diffnet, guizilini20203d, wang2023planedepth, zhao2022monovit, zhou2017unsupervised}.
The easiest way to achieve this is by stereo reconstruction \cite{garg2016unsupervised, godard2017unsupervised}, which enables learning of metric depths due to the known stereo camera baseline. 
In addition to the depth estimation network, SfMLearner \cite{zhou2017unsupervised} proposed using a separate network for relative pose estimation of nearby video frames, enabling view synthesis and joint training of both networks via reconstruction error. 
Further works explore the addition of stereo frames \cite{zhan2018unsupervised, godard2019digging}, IMU \cite{han2019deepvio}, or velocity measurements \cite{guizilini20203d} in order to solve the scale ambiguity problem of monocular video. 
Inspired by the success of transformer in supervised approaches, Monovit \cite{zhao2022monovit} introduces parallel self-attention and convolutional layers during feature encoding. 
Recently, PlaneDepth \cite{wang2023planedepth} has achieved state-of-the-art results by approximating the scene as a mixture of orthogonal planes, which distinctively improves the accuracy in highly planar regions.

\subsection{Domain generalization for monocular depth estimation}
The goal of domain generalization is to achieve satisfactory performance on an unseen test domain while training on various source domains, which may often be indeterminately different from the test domain.
To achieve this, relative depth estimation approaches train networks on diverse large-scale internet data collected across different environments and camera systems \cite{chen2016single, yin2021learning, ranftl2020towards, xian2018monocular, yin2020diversedepth, wang2019web, xian2020structure, li2018megadepth}. 
In a seminal work, Chen et al. \cite{chen2016single} introduced an ordinal relations-inspired loss, which instructs the network to focus on relative distances, thus enabling training on data collected with diverse camera systems. 
MegaDepth \cite{li2018megadepth} uses a scale-invariant loss \cite{eigen2014depth} for optimization on a novel large-scale dataset created from images of well-photographed landmarks.
MiDaS \cite{ranftl2020towards} further proposes an approach that is invariant to the unknown shift, often induced during stereoscopic post-processing from data sources such as 3D movies.
Unfortunately, predicting depths with unknown shifts leads to the inability to accurately recover 3D scene shape.
LeReS \cite{yin2021learning} alleviates this issue via a novel module designed for the refinement of distorted point clouds. 
The main issue of the aforementioned methods is their inability to estimate metrically correct depth due to the usage of scale and shift invariant losses.
Furthermore, owing to the lack of additional constraints, these methods rely exclusively on the relative object size cue.

Another line of work, which specifically addresses the domain gap induced by the usage of various camera systems, investigates the embedding of known camera parameters as an additional input at a certain stage of the network. 
In theory, this provides the network with useful information and enhances generalization to camera systems not present in the training data. 
To solve the focal length/depth ambiguity, He et al. \cite{he2018learning} collected new datasets with varying focal lengths while embedding the known focal length value as an additional input.
Instead of simple embedding via a fully-connected layer, CAM-Convs \cite{facil2019cam} proposes to embed the parameters in convolutional channels and fuse them in skip connections between the encoder and decoder, facilitating generalization for different sensor sizes, principal points, and focal lengths. 
On the other hand, Zhao et al. \cite{zhao2021camera} investigate the effects of extrinsic parameters, namely camera height and camera pitch, on the performance of MDE. 
They developed data augmentation and extrinsic encoding to mitigate the overfitting due to the perspective geometry bias induced by the dominant extrinsic parameters in the training dataset. Finally, related to this work, in \cite{koledic2023towards} we presented the ground-plane embeddings and demonstrated their effectiveness on the synthetic data.

\subsection{Domain adaptation for monocular depth estimation}
The task of domain adaptation is to adapt the model, which was previously trained on either known or unknown source data, to target data originating from a different high-level distribution. 
For this purpose, domain adaptation, unlike domain generalization, gives the model access to the target domain data during training.

The primary goal of research in this area is to enable training on synthetic data due to the easily obtainable dense ground-truth. 
Naturally, due to the significant sim2real gap, performance on real-world data degrades dramatically when deploying models trained purely with synthetic data. 
Most methods employ adversarial domain adaptation to reduce this distribution shift, either in the input space, feature space, or both \cite{hoffman2016fcns, tzeng2017adversarial}.
One of the most popular approaches assumes that the distribution shift is predominantly induced by the change of style between the two domains, while content remains largely the same. 
This is first examined in \cite{atapour2018real}, where the authors propose using cycle-consistency \cite{hoffman2018cycada} to translate the style, while simultaneously preserving the content. 
On the other hand, AdaDepth \cite{kundu2018adadepth} adversarially aligns encoded high-dimensional image representations, encouraging the network to learn domain-invariant features. 
T2Net \cite{zheng2018t2net} uses a combination of both approaches, further explored in \cite{chen2019crdoco, zhao2019geometry, akada2022self, lo2022learning}. 
Notably, LFDA \cite{lo2022learning} considers an advanced method for feature decomposition into content and style features, enabling the latter to be ignored during domain alignment.

Recently, the trend has been to include an additional loss of the target domain to facilitate domain alignment, usually by reconstructing the nearby view \cite{zhao2019geometry, lo2022learning, gurram2021monocular, pnvr2020sharingan, cheng2020s}.
For example, GASDA \cite{zhao2019geometry} enforces the epipolar constraint between the rectified stereo images, while MonoDEVSNet \cite{gurram2021monocular} trains an additional pose network to enable view reconstruction for images from unknown nearby views. 
Additionally, the incorporation \cite{ cheng2020s, lopez2023desc} or joint learning \cite{saha2021learning} of semantic information can lead to further performance improvements. For instance, DESC \cite{lopez2023desc} enforces the semantic consistency on the target domain via a pretrained segmentation network.

%% file: sections/parameters_effects.tex
\begin{figure}[!t]
	\centering
	\includegraphics[width=0.8 \hsize]{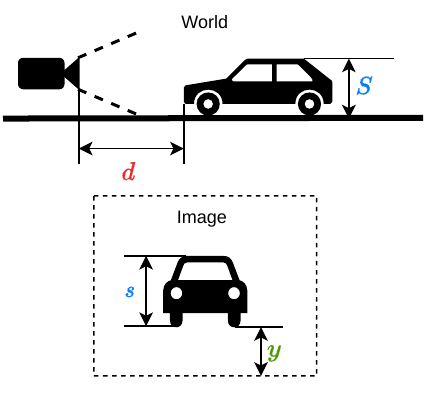}
	\caption{Illustration of the possible depth cues used in MDE. The most common cue for estimating depth $d$, is the object imaging size $s$ - bigger objects correspond to smaller depths. Another cue, unique to MDE for ground vehicles, is the vertical image position $y$ - lower objects correspond to smaller depths.}
	\label{fig:car}
\end{figure}

\begin{figure*}[!t]
	\includegraphics[width=\hsize]{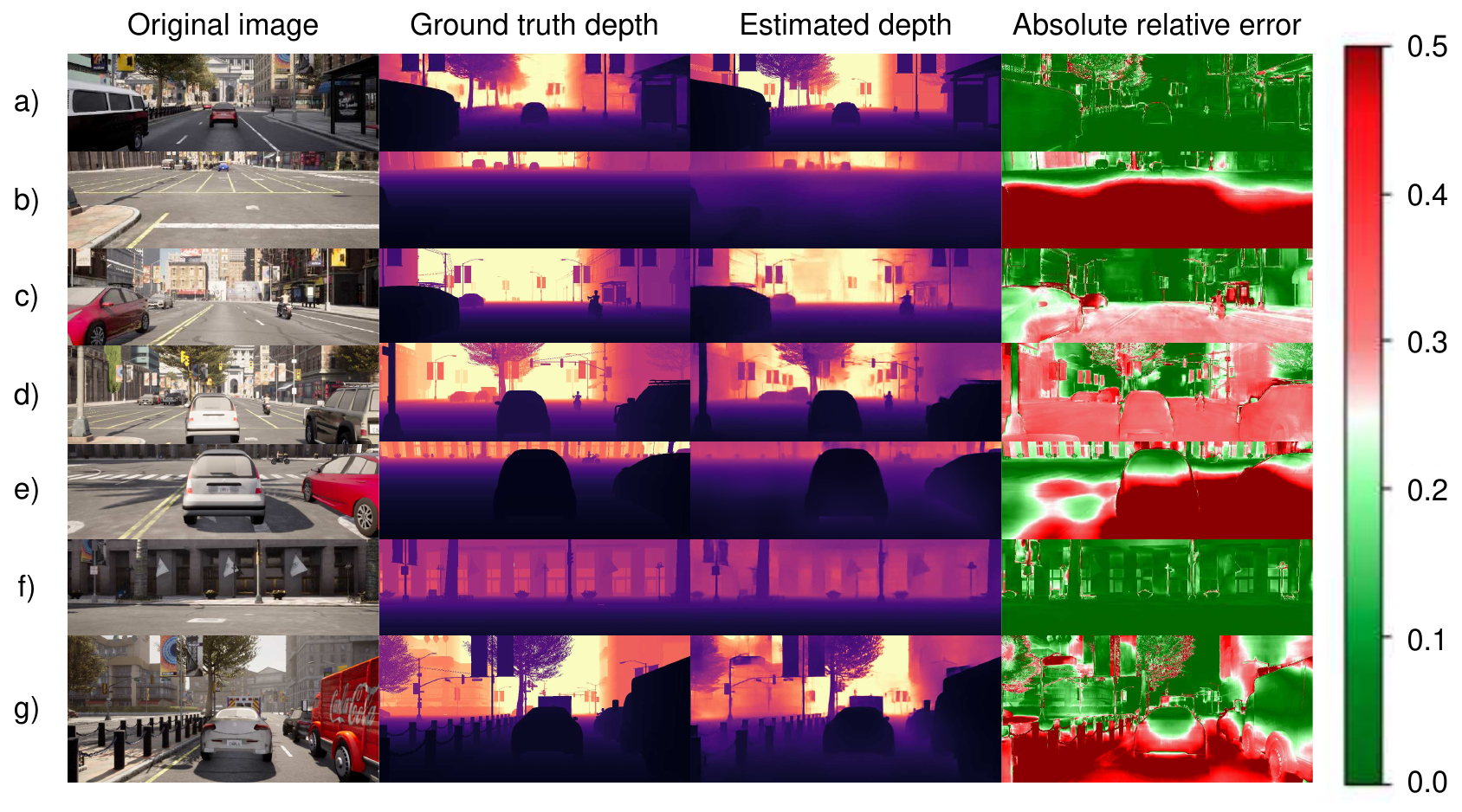}
	\caption{Effects of camera parameter variations. We train the baseline encoder-decoder architecture similar to \cite{godard2019digging} on a dataset with a fixed vehicle-camera system. We then change the parameters one at a time (here referred to as \textit{training value} $\to$ \textit{testing value}) and show the accuracy of the model for different image regions: \textbf{a)} training parameters, \textbf{b)} $\alpha$ - camera pitch  $(0^\circ \to -10^\circ)$, \textbf{c)} $h$ - camera height $ (1.5 \text{m} \to 1.2 \text{m})$, \textbf{d)} $f_v$ - vertical focal length $(570 \text{px} \to 750 \text{px})$, \textbf{e)} $c_v$ - vertical principal point coordinate $(160 \text{px} \to 120 \text{px})$, \textbf{f)} $f_h$ - horizontal focal length $(1824 \text{px} \to 2400 \text{px})$, \textbf{g)} $H$ - vertical sensor size $(320 \text{px} \to 480 \text{px})$. Since $f_v$ and $H$ are coupled via the equation $\theta_v = 2\arctan(H/(2f_v))$, where $\theta_v$ represents the vertical field of view, we vary $f_v$ and $H$ by changing the field of view, and thus keeping the other parameters fixed.}
	\label{fig:params}
\end{figure*}

The ability to generalize is an indispensable prerequisite for the scalable deployment of deep learning solutions in autonomous driving. 
Obviously, an autonomous vehicle should be able to function in various environments, whether it is employed in a rainy city or a sunny rural environment. 
This environmental domain gap is a major focus of research in many perception tasks, including MDE.
However, the domain gap can also be caused by the variation of camera parameters, including both intrinsic and extrinsic parameters.
Naturally, the perspective geometry changes considerably when images are taken from a different viewpoint with a different sensor.
Due to the inherent geometric nature of the MDE task, this variation has a significant impact on accuracy, even when it induces a rather small difference in the final scene appearance on resulting images. 
Unfortunately, this problem is often ignored, primarily due to the lack of data with sufficiently diverse camera parameters. 
Since most datasets are acquired with a single vehicle-camera setup, testing is often performed on a different split of the same dataset, which does not contain the aforementioned perspective geometry perturbations.

\subsection{Overfitting to vertical image position}
In order to the problem of limited generalization capability, it is beneficial to investigate how models actually learn depth. 
Consider the illustration in Fig. \ref{fig:car}. 
One possible cue to estimate depth $d$ from monocular images is to use the relation between imaging size $s$ and real-world size $S$:
\begin{equation}
	d = f_v \frac{S}{s},
\end{equation}
where $f_v$ is the vertical focal length, with $(s, f_v)$ expressed in terms of pixels and $(S, d)$ in meters.
In such a manner, if the focal length is known, models are essentially optimized to recognize known objects and learn their real-world size $S$. 
This is the most common cue, which is heavily exploited in MDE for indoor scenes. 
Note that this assumes that the focal length is constant throughout the dataset. 
If the model is trained with data from multiple cameras, focal length is usually abstracted away, thus estimating depth up to an unknown scale. 
Such an approach is common in relative depth estimation methods \cite{chen2016single, yin2021learning, ranftl2020towards, xian2018monocular, yin2020diversedepth, wang2019web, xian2020structure, li2018megadepth}.

Another possible cue, which is specific to ground vehicles with a fixed camera system, is the vertical image position $y$, as visualized in Fig. \ref{fig:car}. 
Considering a calibrated pinhole camera model with negligible distortions, the depth at which optical rays at position $y$ intersect the ground plane can be calculated as
\begin{equation}
	d = \frac{f_vh}{(H - c_v - y)\cos(\alpha) - f_v\sin(\alpha)},
	\label{eqn:ground}
\end{equation}
where $H$ is the vertical sensor size, $c_v$ is the vertical principal point coordinate, $\alpha$ is the  camera pitch and $h$ is the camera height from the ground plane. 
Here we ignore the effects of camera yaw and roll, as these transformations are rare in autonomous driving.
If the vehicle-camera system is fixed throughout the dataset, the model has no incentive to directly estimate $(\alpha, h, f_v, c_v, H)$. 
Instead, it learns the mapping $y \to d$ which is induced by the constant values of parameters. 
It has been shown that this is a dominant cue in autonomous driving scenarios \cite{dijk2019neural}, where models almost completely ignore the object size cue.

With these insights, it is obvious how variations of camera parameters can affect depth estimation. 
If the parameters change compared to the training data, the learned mapping $y \to d$ no longer holds, which causes a significant drop in accuracy.
To ablate the effects of each parameter, we create bespoke datasets in the CARLA simulator \cite{dosovitskiy2017carla}. 
First, we train the model on a dataset with a fixed vehicle-camera system, and then we test it on data with parameter variations. 
We identify $(\alpha, h, f_v, c_v, H)$ as the primary causes of domain shift and visualize the effects of their changes in Fig. \ref{fig:params}. 
We argue that variations of other parameters are either not frequent enough (e.g. camera roll), or do not have a significant effect on performance (e.g., camera yaw, horizontal focal length). 

From the results in Fig. \ref{fig:params}, three main conclusions can be drawn: (i) although the model correctly estimates the structure and object edges, the parameter variation significantly degrades the accuracy of the depth estimation of the ground plane and the objects on it, (ii) while the degradation in accuracy can be mostly predicted by invalidation of the learned vertical image position cue, there are some inconsistencies, which means that the model uses additional unknown cues, (iii) parameter variations that affect the horizontal appearance of the image (e.g., horizontal focal length) do not degrade the accuracy, implying that the model primarily focuses on vertical cues.

Considering that the usage of the vertical image position cue is a primary cause of performance deterioration in the presence of parameter variation, one could argue that the model should be incentivized to focus on other possible cues. 
Some works explore this approach by carefully implementing augmentations that should focus the model on the object size cue \cite{peng2021excavating, he2022ra, wang2023planedepth}. 
However, we argue that ignoring the vertical position cue is both impossible and undesirable. 
A major drawback of a model that relies primarily on object size cue is the almost certain degradation in performance for image regions without recognizable objects.
This may be due to the complete absence of objects (e.g., cars), or even due to the presence of objects or structures without adequate representation in the training dataset. 
Furthermore, we believe that it is suboptimal to ignore the known geometric relationship between the camera system and the ground plane. 
To that end, our camera parameters embeddings presented in Section \ref{sec:embeddings} exploit this relationship to provide the model with useful geometric information that is entirely induced by the values of the intrinsic and extrinsic parameters.

\subsection{Diverse data generation}
While the inability to generalize for various camera parameters can be explained by the way models learn depth, the main cause of these issues is the lack of data with sufficiently diverse camera parameters. 
Let $\mathcal{D}_i$ represent the i'th domain composed of images $\bd{x} \in \mathbb{R}^{H_i \times W_i \times C_i}$ and depth labels $\bd{y} \in \mathbb{R}^{H_i \times W_i}$. 
Here, we consider that images and corresponding labels are sampled from a high-level distribution parameterized by domain variables $\psi_i$, i.e., $\mathcal{D}_i = \{(\bd{x}_j, \bd{y}_j)\}_{j=1}^{N_i} \sim P_{XY}(\psi_i)$, which can be further separated into environmental variables $\mathcal{E}_i$ and camera variables $\mathcal{C}_i$, i.e., $\psi_i = (\mathcal{E}_i, \mathcal{C}_i)$. 
Environmental variables may represent factors like snow, geographic location etc., while camera variables include both intrinsic and extrinsic parameters. 

For the scalable deployment of MDE in autonomous driving, models should be able to generalize over a wide distribution of environments and vehicle-camera systems. 
Formally, the model should perform accurately on the testing domain 
\begin{equation}
	\mathcal{D}_{Test} = \{\mathcal{D}_i \sim P_{XY}(\psi_i)\: | \: \psi_i \in \bm{
	\mathcal{E}} \times \bm{
		\mathcal{C}}\},
\end{equation}
where $\bm{\mathcal{E}}$ and $\bm{\mathcal{C}}$ represent possibly infinite sets of feasible environments and camera parameters.
For the sake of simplicity, we assume that satisfactory generalization performance is achieved by training on a finite number of environments $\{\mathcal{E}_i\}_{i=1}^P \subseteq \bm{\mathcal{E}}$ and vehicle-camera systems $\{\mathcal{C}_i\}_{i=1}^Q \subseteq \bm{\mathcal{C}}$. 
In order to achieve this, one may use different configurations of training data $\mathcal{D}_{train} = \{\mathcal{D}_i\}_{i=1}^M$. 
The optimal solution would sample data from all possible combinations of environments and vehicle-camera systems: 
\begin{equation}
	\mathcal{D}_{train} = \{\mathcal{D}_i \sim P_{XY}(\psi_i) \: | \: \psi_i \in
	\{\mathcal{E}_j\}_{i=j}^P \times \{\mathcal{C}_j\}_{i=j}^Q\}.
\end{equation}
Unfortunately, this is intractable, as the collection of such diverse data is operationally almost impossible. 
In each environment, data would need to be acquired with many different cameras capturing images from diverse viewpoints.

A more realistic approach would be to formulate the training data as 
\begin{equation}
	\mathcal{D}_{train} = \{\mathcal{D}_i \sim P_{XY}(\psi_i) \: | \: \psi_i \in
	\{(\mathcal{E}_i, \mathcal{C}_i)\}_{i=1}^M\}.
\end{equation}
Here, for each environment, only one vehicle-camera system is used to acquire the data. 
One could create such a dataset via mixing of publicly available datasets acquired with fixed vehicle-camera systems such as KITTI \cite{geiger2013vision}, Oxford RobotCar \cite{maddern20171} or Cityscapes \cite{cordts2016cityscapes}. 
However, we argue that this approach can fail under certain conditions, especially if the architecture and training procedure are not properly designed. 
As models have a high discriminatory capacity, they may learn to associate the environment with its corresponding camera parameters in the training data. 
For example, let's assume that environment $\mathcal{E}_1$ is distinctive enough from other environments in $\mathcal{D}_{train}$, e.g., the only environment with snow. The model could then learn to recognize snow, and if snow is identified in the image, estimate depth by overfitting to the camera parameters $\mathcal{C}_1$. 
Therefore, model would fail to perform well on a testing dataset collected in a snowy environment with different camera parameters (i.e., $(\mathcal{E}_1, \mathcal{C}_2)$), even if these camera parameters were present in the training data, albeit in a different environment. 
Nevertheless, this approach constitutes a promising research direction, as a careful implementation and data augmentation can resolve these issues. 

However, due to the lack of publicly available data with sufficiently diverse camera parameters, we choose a different approach in this work, which allows us to use easily collectable data from a simulation environment. 
Our proposed method is specifically designed to generalize to data acquired with any camera setting in the target environment, so that we are able to produce accurate metric depth predictions for multiple datasets with different camera intrinsics and extrinsics.
Remarkably, we achieve this by using only a single real-world dataset with a fixed vehicle camera system during training.

%% file: sections/method.tex
\begin{figure}[!t]
	\centering
	\includegraphics[width=\hsize]{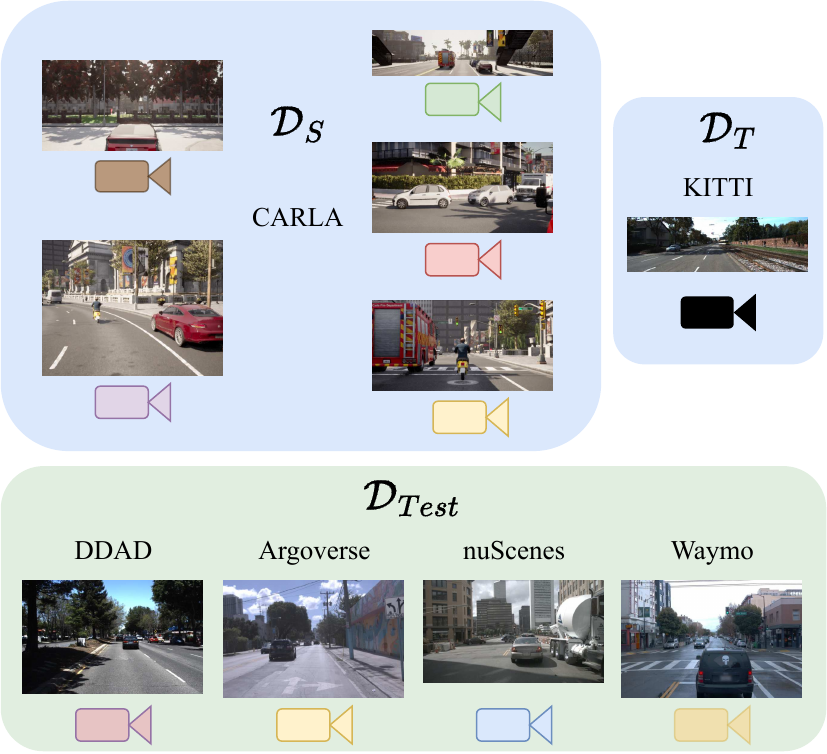}
	\caption{Our data configuration for sim2real adaptation. For the source dataset $\mathcal{D}_S$, we collect synthetic data from the CARLA simulator \cite{dosovitskiy2017carla}. The target dataset $\mathcal{D}_T$ is from the real-world environment with certain environmental characteristics (e.g., sunny daytime, such as KITTI \cite{geiger2013vision}. Our method should be able to generalize to arbitrary vehicle-camera systems in the target environment, e.g., DDAD \cite{guizilini20203d}, Argoverse \cite{Chang_2019_CVPR}, nuScenes \cite{nuscenes2019}, Waymo \cite{Sun_2020_CVPR}.)}
	\label{fig:config}
\end{figure}

In this section, we present our contributions that enable accurate monocular depth estimation in autonomous driving for an arbitrary vehicle-camera setup. 
We achieve this by the simultaneous optimization of two objectives: (i) equivariance to the camera parameters variations, facilitated by the usage of synthetic data collected with the diverse vehicle-camera systems, (ii) transfer of the learned equivariance to the environmental features of the real world, via the usage of a single real-world dataset with a fixed vehicle-camera system.
After a careful inspection of the problem and ablation of possible design ideas, we present our main contributions: a novel embedding of camera parameters based on the geometric relationship between the camera and the ground plane, and a novel architecture that carefully incorporates this parameterization to enable domain adaptation.

\subsection{Problem statement}
\begin{figure*}[!t]
	\includegraphics[width=\hsize]{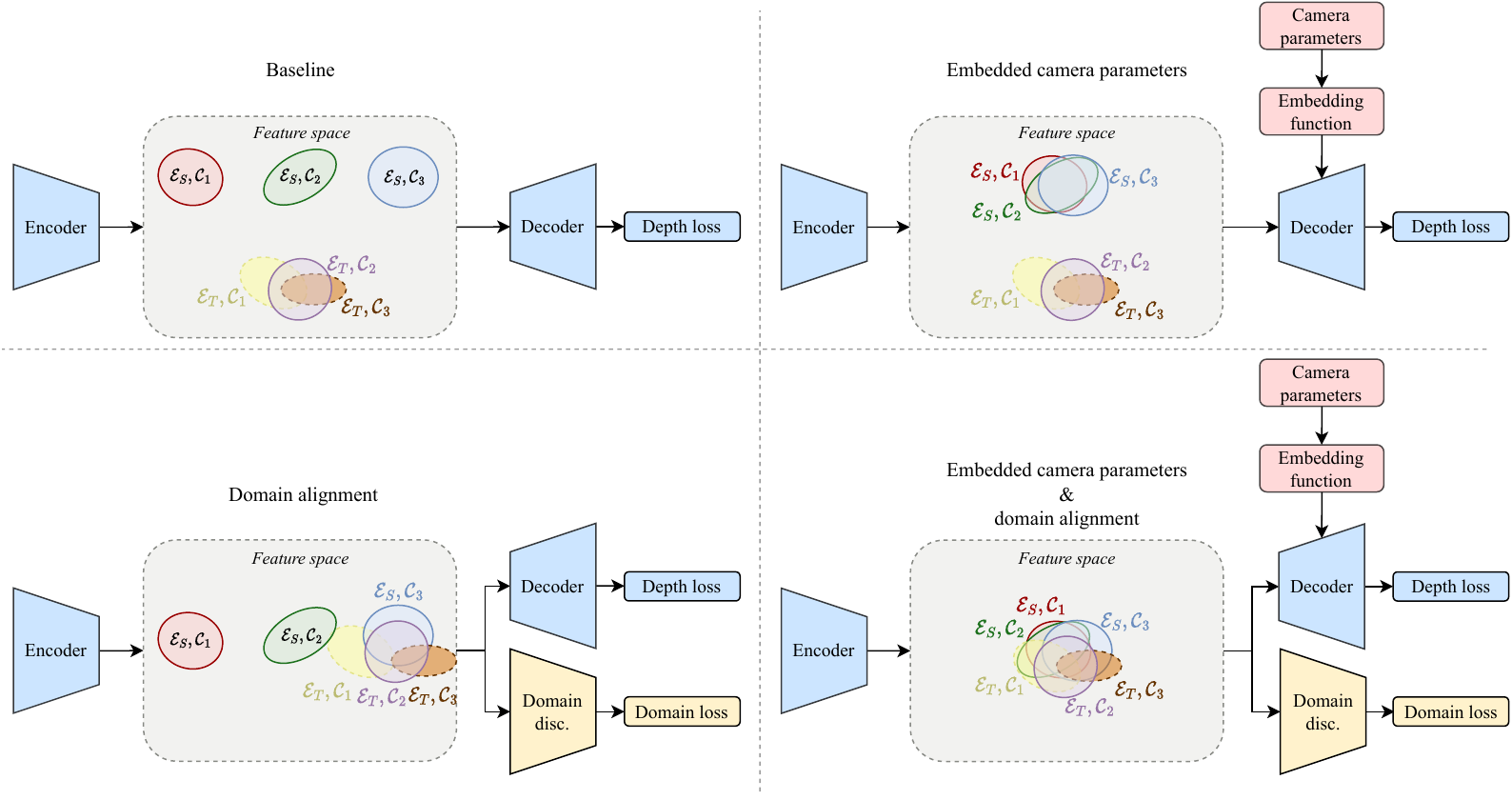}
	\caption{Possible feature space mapping with respect to the different training strategies in a simple data configuration. The training data includes images from the source domain acquired with three vehicle-camera systems $(\mathcal{E}_S, \mathcal{C}_1), (\mathcal{E}_S, \mathcal{C}_2), (\mathcal{E}_S, \mathcal{C}_3)$ and images from the target domain acquired with a single vehicle camera system $(\mathcal{E}_T, \mathcal{C}_2)$. The task is to generalize to the test data in the target domain with the configurations $(\mathcal{E}_T, \mathcal{C}_1), (\mathcal{E}_T, \mathcal{C}_3)$. We present hypothetical feature representations learned by the encoder for different training strategies: \textbf{top left} -- standard encoder-decoder architecture, \textbf{top right} -- known camera parameters are embedded and given to the decoder as an additional input, \textbf{bottom left} -- learned features from the source and target domains are adversarially aligned via domain discriminator, \textbf{bottom right} -- a combination of domain alignment and camera parameters embedding. }
	\label{fig:domains}
\end{figure*}
\label{sec:problem}
Let $\mathcal{D}_{Train}$ be the training data collected in both the source environment $\mathcal{E}_S$ and the target environment $\mathcal{E}_T$. 
Source data is collected with a diverse set of camera parameters $\{C_i\}_{i=1}^M$, while the target data includes images collected with a single vehicle-camera system $\mathcal{C}_T$.
Formally, we consider the following training data configuration:
\begin{equation}
	\begin{gathered}
		\mathcal{D}_{train} = \mathcal{D}_S \cup \mathcal{D}_T, \\
		\mathcal{D}_S = \{\mathcal{D}_i \sim P_{XY}(\psi_i) \: | \: \psi_i \in
		\{(\mathcal{E}_S, \mathcal{C}_i)\}_{i=1}^M\}, \\
		\mathcal{D}_T \sim P_{XY}(\psi_T), \psi_T = (\mathcal{E}_T, \mathcal{C}_T).
	\end{gathered}
\label{eqn:data}
\end{equation}
The task is to enable generalization to $\mathcal{D}_{Test}$, which contains images collected with varying vehicle-camera systems in the target environment:
\begin{equation}
	\mathcal{D}_{Test} = \{\mathcal{D}_i \sim P_{XY}(\psi_i)\ \: | \: \psi_i \in   \{\mathcal{E}_T\}\times \bm{\mathcal{C}}\},
\end{equation}
where $\bm{\mathcal{C}}$ represents a set of feasible camera parameters.

While we designate our method for sim2real adaptation scenario depicted in Fig. \ref{fig:config}, it is applicable to any problem that involves data from a source environmental domain with diverse camera parameters and a target environmental domain with images collected by the single vehicle-camera system.

\subsection{Design considerations}
\label{sec:design}
The described problem requires careful implementation of the architecture and optimization strategy, as it is prone to overfitting induced by both the environmental and camera parameter domain gaps. 
In Fig. \ref{fig:domains}, we explore different designs and visualize possible feature space representations, which are simplified for illustration purposes. 
We assume that domains $\mathcal{E}_S$ and $\mathcal{E}_T$ are sufficiently different, and that the encoder has enough discriminatory capacity to recognize the environmental domain to which an image belongs to.

First, we consider the usage of a simple encoder-decoder architecture termed "Baseline". 
With the assumption of sufficient discriminatory capacity, the encoder successfully maps images the from source and target environments to different regions of the feature space. 
Additionally, since the training data includes different vehicle-camera systems in the source domain, i.e., $(\mathcal{E}_S, \mathcal{C}_1), (\mathcal{E}_S, \mathcal{C}_2), (\mathcal{E}_S, \mathcal{C}_3)$, a properly optimized encoder learns features that are equivariant to camera parameter variations.
For example, one of such features could be the horizon level or homography of the ground plane. 
The decoder can then use this to accurately estimate depth for all three vehicle-camera systems. 
The main issue, however, is that these features are likely specific to the source domain. 
Training data from the target domain includes images collected with only one vehicle-camera system, i.e., $(\mathcal{E}_T, \mathcal{C}_2)$, which means that the model has no incentive to learn features that enable the estimation of camera parameters in the target domain. 
This translates to the inability of the system to successfully generalize to the test data, which includes $(\mathcal{E}_T, \mathcal{C}_1)$ and $(\mathcal{E}_T, \mathcal{C}_3)$. 
Basically, what happens is that the model successfully determines that images belong to the target environment, consequently inaccurately estimating all depths as if the images were collected by the single vehicle-camera system present in the training data, i.e., $(\mathcal{E}_T, \mathcal{C}_2)$.

The primary cause of this overfitting is that the features informing the decoder about the camera parameters do not translate well across environmental domain gaps. 
A possible solution to remedy this problem is to supplement the decoder directly with information about known camera parameters, providing them as an additional input. 
The encoder is therefore relieved from the requirement to discriminate between different vehicle-camera systems, hence enabling the learning of camera parameters invariant features. 
This invariance should theoretically transfer well to the target domain and enable generalization for $(\mathcal{E}_T, \mathcal{C}_1)$ and $(\mathcal{E}_T, \mathcal{C}_3)$. 
Unfortunately, as we will show in our ablation studies, this is usually not the case. 
When the model recognizes an image from the target domain, it learns to simply ignore embedded camera parameters, thereby causing the same generalization issues as in the simple encoder-decoder architecture.
This happens because the supplemented information is effectively useless if the training data includes only one vehicle-camera system, which is true for the target domain.

A possible solution is to use domain adaptation techniques  \cite{hoffman2016fcns, tzeng2017adversarial} to reduce the gap between the features of the source and target environmental domains. 
In MDE, this is frequently done via adversarial feature alignment \cite{kundu2018adadepth, zheng2018t2net, zhao2019geometry, gurram2021monocular}, where a domain discriminator tries to correctly classify the true domain, while the encoder is optimized to fool the discriminator. 
In Fig. \ref{fig:domains}, we show the possible ramifications of the direct usage of such an approach on our problem. 
The main issue is that the model does not explicitly know that the images from domains $(\mathcal{E}_S, \mathcal{C}_2)$ and $(\mathcal{E}_T, \mathcal{C}_2)$ are collected with the same vehicle-camera system. 
This can potentially obscure the domain alignment, leading once again to poor generalization. 

Both the inclusion of embedded camera parameters and the usage of adversarial alignment have potential ramifications when applied separately. 
However, we argue that they complement each other when carefully jointly implemented into the architecture. 
While the supplemented information by the embedded parameters enables learning of camera parameters invariant features, domain alignment allows seamless transfer of that invariance to the target environmental domain, as visualized in Fig. \ref{fig:domains}.
Due to the domain alignment, the model is no longer able to discriminate between the two environmental domains, which effectively means that it can no longer learn to ignore the embedded camera parameters when estimating depth for the images from the target domain.
To that end, we propose to use such design in our approach.

\subsection{Ground plane embedding}
\label{sec:embeddings}
\begin{figure}[!t]
	\includegraphics[width=0.8\hsize]{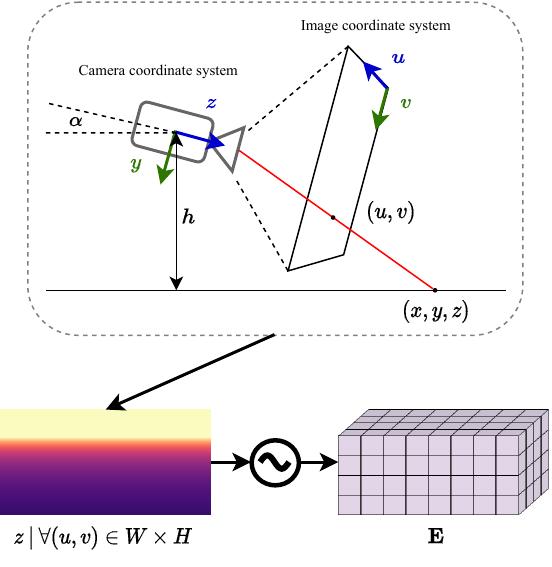}
	\centering
	\caption{Illustration of the proposed ground plane embedding. For each $(u, v)$ in the image, we calculate the depth ($z$ coordinate in the camera coordinate frame) of the intersection point between the optical ray and the ground plane. These depths are then assembled and Fourier encoded into the final embedding $\bd{E} \in [-1, 1]^{H \times W \times 2k + 1}$, where $k$ is the number of frequency bands.}
	\label{fig:embeddings}
\end{figure}

If the training dataset includes images with varying vehicle-camera systems, the model is required to learn features that, in some way, provide information about camera parameters. 
In order to alleviate the model from such a requirement, we propose to embed the parameters and provide the embeddings to the model as an additional input.

Here we consider the embedding of the same parameters whose variation was proved to affect the MDE accuracy in Section \ref{sec:params}. 
Namely, this includes camera pitch $\alpha$, camera height relative to the ground plane $h$, vertical focal length $f_v$, vertical principal point coordinate $c_v$ and image height $H$. 
Similarly to \cite{he2018learning}, a simple approach would be to fuse known values $(\alpha, h, f_v, c_v, H)$ via fully connected layers. 
On the other hand, we argue that the embedding function should perform the following mapping:
\begin{equation}
	(\alpha, h, f_v, c_v, H) \in \mathbb{R}^5 \mapsto \bd{E} \in \mathbb{R}^{H \times W \times C},
\end{equation}
where $\bd{E}$ represents resulting embeddings, with $H, W$ being the image height and width, respectively.
In addition to the encoding of camera parameters, well designed embeddings should provide per-pixel information related to the task in hand, i.e., MDE. 
For example, as the perspective geometry of the scene changes with the variation of camera parameters, each pixel location should be informed of how this variation affects the geometry of the corresponding pixel individually.

To achieve this, we draw inspiration from how models actually learn depth, as ablated in Section \ref{sec:params}. 
The primary cue for MDE in autonomous driving is usually vertical image position, which is strongly associated with the relation between the vehicle-camera system and the ground plane. 
This relation, specific to ground-based vehicles, transforms MDE from an ill-posed problem to a geometrically tractable one. 
Accordingly, we choose to exploit this relationship when designing our embeddings, as visualized in Fig. \ref{fig:embeddings}.

Given the pinhole camera model, the optical ray $\bd{o}$ passing through image coordinates $(u, v)$ can be expresses as a parametric line equation:
\begin{equation}
\bd{o}(z) = z\begin{bmatrix}
		(u - c_u)/f_u \\
		(v - c_v)/f_v \\
		1
	\end{bmatrix},
\end{equation}
where $(c_u, c_v)$ and $(f_u, f_v)$ are principal point coordinates and focal lengths in pixels, while the variable $z$ represents depth in the camera coordinate system. 
Additionally, given the camera extrinsic rotation matrix $\bd{R}(\alpha)$ and camera height from the ground plane $h$, the ground plane equation can be written as:
\begin{equation}
\bd{n}^{\mathrm{T}} \bd{R}^{\mathrm{T}}(\alpha) \bd{p}+h=0,
\end{equation}
where $\bd{n} = [0, -1, 0]^{\mathrm{T}}$ is an ideal ground plane normal.
By substituting $\bd{o}(z) \rightarrow \bd{p}$ we calculate $z$  as the depth of the intersection between the ground plane and the optical ray through pixel $(u, v)$. 
Note that we allow intersections behind the camera for rays above the horizon line, i.e., negative $z$ values. 
Depths are capped to a maximum value specific to the dataset and normalized to $[-1, 1]$.
Subsequently, inspired by \cite{mildenhall2021nerf}, we use Fourier encoding in order to map $z$ to a higher dimension:
\begin{equation}
	z \rightarrow [z, \sin(f_1z), \cos(f_1z), \cdots, \sin(f_kz), \cos(f_kz)]
\end{equation}
where $f_i$ are logarithmically spaced between $2^0\pi$ and $2^{k-1}\pi$.
This process is repeated for each $(u, v) \in W \times H$ and assembled into the final embedding $\bd{E} \in [-1, 1]^{H \times W \times 2k+1}$. 
We argue that this embedding possesses the following desirable properties: 
\begin{itemize}
\item as a function of parameters $(\alpha, h, f_v, c_v, H)$, it can accurately inform the model about parameter variations,
\item it provides the model with useful per-pixel geometric information about the relation between the camera and the ground plane (for example, $z$ is equal to the expected depth of the pixel if no occlusions are present on the ground plane),
\item it enhances generalization to the vehicle-camera systems that are not present in the training data.
\item it encourages the model to exploit the vertical image position cue, which is a cue that is more adaptable to environmental domain gaps (ground-plane contact point is a cue easily detectable across all domains, while priors involving sizes of known objects can vary significantly.)
\end{itemize}

\subsection{Model architecture}
\label{sec:arch}
\begin{figure*}[!t]
	\includegraphics[width=\hsize]{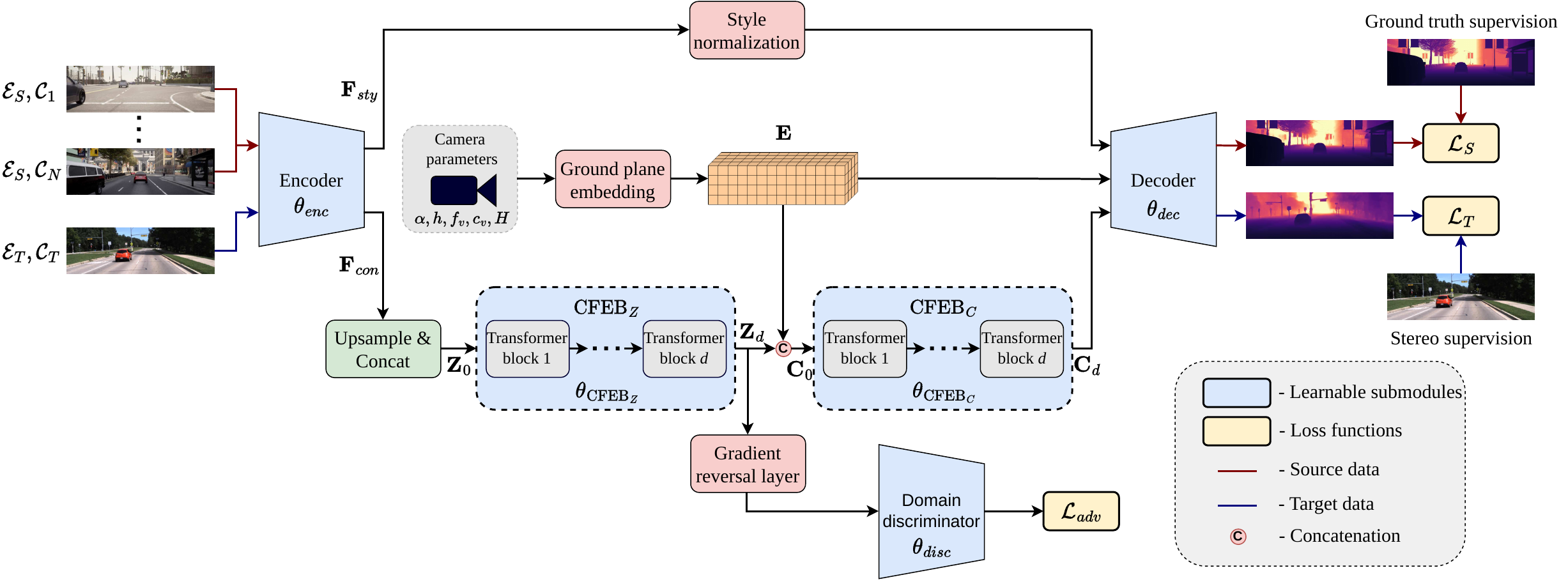}
	\caption{Training pipeline and system architecture. Our training data consists of images acquired with different camera parameters in the source environment $\{(\mathcal{E}_S, \mathcal{C}_1), ... ,  (\mathcal{E}_S, \mathcal{C}_N)\}$ and images acquired with a single vehicle-camera system in the target environment $(\mathcal{E}_T, \mathcal{C}_1)$. Based on the insights ablated in Section \ref{sec:design}, we design an architecture that carefully combines the adversarial domain adaptation and ground plane embedding presented in Section \ref{sec:embeddings}. Our system consists of five learnable submodules: encoder, domain discriminator, decoder, $\operatorname{CFEB}_Z$ and $\operatorname{CFEB}_C$. The first three are based on convolutional architectures, while the last two exploit self-attention. For the training losses, we use dense ground truth supervision in the source domain, and readily available and scalable stereo supervision in the target domain.}
	\label{fig:system}
\end{figure*}

As visualized in Fig. \ref{fig:system}, we design our architecture based on the concepts presented in Section \ref{sec:design}. 
As in standard practice, we consider the usage of an encoder model which produces feature maps at incrementally downscaled resolutions.
Such models are often based on the residual convolutional architectures \cite{he2016deep, liu2022convnet}. 
Recently, inspired by the success of the transformer architecture in language processing \cite{vaswani2017attention}, many works implement self-attention as a feature-learning backbone for general vision tasks \cite{dosovitskiy2015flownet, ranftl2021vision, liu2021swin, wang2021pyramid, li2022next}, or even in MDE \cite{bhat2021adabins, agarwal2023attention, li2022depthformer, piccinelli2023idisc}. 
The primary motivation is to exploit the global receptive field of the self-attention, thus enabling interaction between positionally distant image features, even in shallow layers. 

While this has obvious positive consequences in tasks such as feature matching, we argue that the usage of self-attention in the shallow layers is not beneficial for MDE, especially when accounting for a major increase in computational complexity. 
Feature maps in shallow layers most often contain information about the local structure, i.e., object edges and texture. 
In dense prediction tasks such as MDE, this local structure information is frequently fused in the final layers of the decoder via skip connections \cite{ronneberger2015u}, which is detrimental for accurate fine-grained depth estimation that is consistent with objects boundaries. 
To that end, we advocate towards usage of convolutions in shallow layers, due to the proven capability for seamless extraction of local structure. 

In Fig. \ref{fig:domains} we presented an approach that uses a combination of embedded camera parameters and domain alignment. 
In design of such an approach, we made a simplifying assumption that the feature space includes a singular feature map, which is then passed to the decoder. 
However, a standard practice in dense prediction is to obtain multiple feature maps at incrementally lower resolutions. 
Here, we consider the usage of encoder models that produce feature maps  $(\bd{F}_1, \bd{F}_2, \bd{F}_3, \bd{F}_4, \bd{F}_5)$, with downscaling ratios $(2, 4, 8, 16, 32)$ respectively. 
Unfortunately, it is infeasible to adversarially align all feature maps. 

Inspired by the style-transfer methods \cite{johnson2016perceptual, huang2017arbitrary}, we separate feature maps into $\bd{F}_{sty} = \{\bd{F}_1, \bd{F}_2\}$ and $\bd{F}_{con} = \{\bd{F}_3, \bd{F}_4, \bd{F}_5\}$. 
We presume that $\bd{F}_{sty}$ contains information about low-level local structure such as object edges and domain-specific style information such as texture and contrast. 
Adversarial domain alignment of these features is undesirable due to the presence of local structure information, which, when aligned, might significantly distort the final structure of the estimated depth image. 
On the other hand, $\bd{F}_{con}$ contains high-level content information, which is a good candidate for domain alignment.

Before domain alignment, we proceed with a joint refinement of all content features.
First of all, we concatenate content features as
\begin{equation}
	\bd{Z} = \mathrm{Concat}(\bd{F}_3, \mathrm{Up}(\bd{F}_4, 2), \mathrm{Up}(\bd{F}_5, 4)),
	\label{eqn:concat}
\end{equation}
where $\mathrm{Up}(\bd{x}, n)$ is an upscaling of spatial dimensions by a factor $n$ via nearest-neighbor interpolation. 
Contrary to $\bd{F}_{sty}$, $\bd{F}_{con}$ contains high-level structural information about objects and the underlying geometry, which is suitable for global feature interaction via self-attention. 
For example, it is beneficial to establish correspondence between distant objects that have the same vertical position of their ground plane contact points (this correspondence naturally complements our proposed camera parameters embeddings). 
Accordingly, we design CFEB (\textit{Content Feature Enhancement Block})  where we refine content features in a series of transformer blocks. 

\begin{figure}[!t]
	\centering
	\includegraphics[width=0.8\hsize]{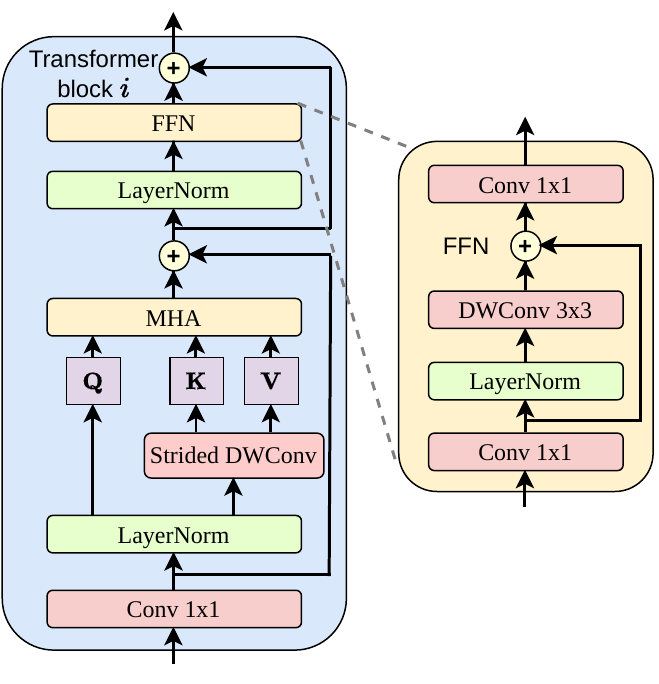}
	\caption{Architecture of the transformer blocks. We use strided depth-wise convolutions to reduce the spatial dimensionality of keys $\mathbf{K}$ and values $\mathbf{V}$. }
	\label{fig:transformer}
\end{figure}

We visualize the architecture of the proposed transformer blocks in Fig. \ref{fig:transformer}.
Due to the presence of varying image resolutions in our problem, we propose omitting the positional embedding \cite{dosovitskiy2020image} when implementing attention blocks. 
This is possible due to the presence of convolutional layers in the encoder, which inadvertently learn positional information \cite{islam2020much}. 
To that end, we use a simple 1x1 convolution as a projection layer that enables us to aggregate features across different channels and to control the channel dimension. 
After layer normalization \cite{ba2016layer}, we proceed with the calculation of multi-head attention (MHA). 
Inspired by \cite{wang2021pyramid}, we reduce the spatial dimension of features before the calculation of keys and values in order to minimize computational complexity. 
Instead of average pooling, we use strided depth-wise convolutions \cite{chollet2017xception}, which induce negligible computational cost. 
Therefore, we calculate queries, keys and values as:
\begin{gather}
	\bd{\tilde{Z}} = \mathrm{DWConv}(\bd{Z}), \\
	\bd{Q} = \bd{Z}\bd{W}_\bd{Q}, \bd{K} = \bd{\tilde{Z}}\bd{W}_\bd{K}, \bd{V} = \bd{\tilde{Z}}\bd{W}_\bd{V}.
\end{gather}
Here, $\mathrm{DWConv}$ transforms $\bd{Z} \in \mathbb{R}^{H_z \times W_z \times C_z}$ into $\bd{\tilde{Z}} \in \mathbb{R}^{\frac{H_z}{s} \times \frac{W_z}{s} \times C_z}$, where $s$ represents the stride of the convolutions.  
Afterwards, we rearrange $\bd{Z}$ and $\bd{\tilde{Z}}$ into two-dimensional tensors before projection with learnable parameters $\bd{W}_\bd{Q}, \bd{W}_\bd{K} , \bd{W}_\bd{V}  \in \mathbb{R}^{C_z \times d}$ to queries $\bd{Q} \in \mathbb{R}^{N_q \times d}$, keys $\bd{K} \in \mathbb{R}^{N_{kv} \times d}$, and values $\bd{V} \in \mathbb{R}^{N_{kv} \times d}$.
In a standard manner, we calculate the attended output $\bd{A}$ as:
\begin{equation}
	\bd{A}=\operatorname{softmax}\left(\frac{\bd{Q K}^{\mathrm{T}}}{\sqrt{d}}\right) \bd{V},
\end{equation}
which is calculated for multiple heads and then fused via the linear layer.
Due to the spatial reduction induced by the strided depth-wise convolutions, computational complexity of the attention calculation is reduced from $\mathcal{O}(2H_z^2W_z^2C_z)$ to $\mathcal{O}(\frac{2H_z^2W_z^2C_z}{s^2})$.

After attention calculation, we proceed with feature refinement in the feed-forward model (FFN). 
Similarly to \cite{li2021localvit, li2022next, wang2021pyramid}, we rearrange features back into three-dimensional tensors and perform spatial convolution operations to aggregate local features. As visualized in Fig. \ref{fig:transformer}, FFN is modeled as a 3x3 $\mathrm{DWConv}$ sandwiched between two 1x1 convolutions that expand and shrink the channel dimension by a factor of 4 in order to learn more complex feature representations.

To aggregate, $\operatorname{CFEB}_Z$ incorporates $d$ successive transformer blocks, with each performing a mapping $\bd{Z}_i \mapsto \bd{Z}_{i + 1}$:
\begin{gather}
	\bd{Z}_i = \operatorname{Conv}(\bd{Z}_i) \\
	\bd{Z}_i = \operatorname{MHA}(\operatorname{LayerNorm}(\bd{Z}_i)) + \bd{Z}_i \\
	\bd{Z}_{i + 1} = \operatorname{FFN}(\operatorname{LayerNorm}(\bd{Z}_i)) + \bd{Z}_i,
\end{gather}
with $\bd{Z}_0$ initialized via Eq. \eqref{eqn:concat}.
To that end, after performing the mapping $\operatorname{CFEB}_Z: \bd{Z}_0 \mapsto \bd{Z}_d$, we consider $\bd{Z}_d$ to be a good candidate for domain alignment. 
As visualized in Fig. \ref{fig:system}, we forward $\bd{Z}_d$ to the domain discriminator to adversarially align target and source domain features. This encourages the model to optimize $\bd{Z}_d$ as features that are invariant to the environmental domain gap.
Architecture of the domain discriminator is relatively simple. It performs the mapping $\bd{Z}_d \mapsto \{0, 1\}$ via a sequence of convolutional and pooling layers, finalized by global average pooling \cite{lin2013network} due to the variable image resolution in the training data.

We proceed with the architecture design as proposed in Fig. \ref{fig:domains}, where we provide embedded camera parameters to the model in conjunction with domain adaptation. 
It is paramount that the embeddings are provided to the model exclusively after the features were successfully adversarially aligned. 
If that was not the case, the discriminator would try to align the provided embeddings, essentially leading to the loss of the information about the known camera parameters.

To that end, we provide the proposed camera embeddings $\bd{E}$ after the model has learned domain-invariant features $\bd{Z}_d$. 
While we experimented with various means of fusing $\bd{E}$ and $\bd{Z}_d$, we determined that a simple concatenation followed by refinement via transformer blocks works best: 
\begin{gather}
	\bd{C}_0 = \operatorname{Concat}(\bd{Z}_d, \bd{E}), \\
	\operatorname{CFEB}_C: \bd{C}_0 \mapsto \bd{C}_d,
\end{gather}
Here, the $\operatorname{CFEB}_C$ block effectively fuses camera embeddings $\bd{E}$ with domain invariant features  $\bd{Z}_d$, producing features $\bd{C}_d \in \mathbb{R}^{H_c \times W_c \times C_c}$, which contain information about the known camera parameters. Since $(H_c, W_c) = \frac{1}{8}(H, W)$, similarly to \cite{godard2019digging}, we design a decoder that successively upsamples $\bd{C}_d$ into a final depth map $\bd{D}$. 

In order to accurately estimate fine details like object edges, a standard procedure is to fuse $\bd{F}_{sty} = \{\bd{F}_1, \bd{F}_2\}$, which contain information about local structure. 
However, these features also contain information about contrast and texture, which are specific to the particular environmental domain. 
This information may effectively produce features that are not domain-invariant, therefore hindering the desired effects of adversarial alignment. 

It has been shown \cite{johnson2016perceptual, huang2017arbitrary} that the style discrepancy can be primarily attributed to a difference in channel-wise mean $\mu$ and variance $\sigma$ of the corresponding feature maps. 
This discrepancy can be effectively removed via adaptive instance normalization \cite{huang2017arbitrary}. 
Specifically for our problem, we perform Style Normalization (SN), where we scale the feature maps of the source domain to align them with the statistics of the target domain. 
During training, each batch involves images sampled from the source domain $\mathcal{D}_S$ and the target domain $\mathcal{D}_t$, which induces features $\bd{F}_{sty}^S$ and  $\bd{F}_{sty}^T$. 
Accordingly, we normalize the source feature maps and scale 
\begin{equation}
	\bd{s_i} = \sigma(\bd{t_i})\left(\frac{\bd{s_i} - \mu(\bd{s_i})}{\sigma(\bd{s_i})}\right) + \mu(\bd{t_i}).
\end{equation}
Here, $(\bd{s}_i, \bd{t}_i) \in (\bd{F}_{sty}^S, \bd{F}_{sty}^T)$ are corresponding feature maps for each $i \in [1..C_{sty}]$, where $C_{sty}$ is the overall number of channels in $\bd{F}_{sty}$. 
Furthermore, we also fuse $\bd{E}$ in decoder convolutional layers to provide local information about the expected depth of the ground plane. 
Finally, at the output of the decoder, we estimate the depth map $\bd{D} \in [0, D_{max}]^{H \times W}$, with $D_{min}$ and $D_{max}$ being specific to the dataset.

\subsection{Optimization}
In order to optimize all network submodules, we use a combination of target, source and adversarial losses. 
We assume that the source domain data is collected with appropriate depth labels, i.e., $\mathcal{D}_S = \{(\bd{I}_i, \bd{\hat{D}}_i)\}_{i=1}^{N_S}$, where $\bd{I}_i \in \mathbb{R}^{H \times W \times C}$ refers to the image with a corresponding dense ground truth depth map $\bd{\hat{D}}_i \in \mathbb{R}^{H \times W}$. 
For the source training loss, we follow a standard approach for supervised depth estimation \cite{eigen2014depth}. Given the logarithmic distance $g_i = \log(\hat{d}_j) - \log(d_j)$ at a pixel location $j$, the source loss is:
\begin{equation}
	\mathcal{L}_S=\beta \sqrt{\frac{1}{|\bd{D}|} \sum_i g_i^2-\frac{\gamma}{|\bd{D}|^2}\left(\sum_i g_i\right)^2},
\end{equation}
where we use $\beta = 10$ and $\gamma = 0.85$ as in \cite{bhat2021adabins}.

To extend the applicability of our method, we consider a target domain without ground-truth labels, possibly due to the difficulty of data acquisition in the respective environment. 
To that end, target domain data $\mathcal{D}_S = \{(\bd{I}^L_i, \bd{I}^R_i)\}_{i=1}^{N_T}$ includes only left $\bd{I}^L_i \in \mathbb{R}^{H \times W \times C}$and right $\bd{I}^R_i \in \mathbb{R}^{H \times W \times C}$ stereo images, with known extrinsic calibration parameters $\mathcal{R}$ which enable rectification. 
In order to provide a supervision signal for the target domain images, we use a view reconstruction loss \cite{godard2017unsupervised}:
\begin{gather}
	\bd{\hat{I}}^L = \mathcal{W}(\bd{I}^R, \bd{D}, \mathcal{R}),\\
	\mathcal{L}_{recon} = \frac{\eta}{2}(1 - \operatorname{SSIM}(\bd{\hat{I}}^L, \bd{I}^L)) + (1 - \eta)\lVert\bd{\hat{I}}^L - \bd{I}^L \rVert,
\end{gather}
where $\bd{\hat{I}}^L \in \mathbb{R}^{H \times W \times C}$ is a generated left stereo image, synthesized by a warping function $\mathcal{W}$. $\operatorname{SSIM}$ refers to the Structural Similarity Index Measure \cite{wang2004image} with parameter $\eta = 0.85$.
Additionally, we incentivize smooth depth maps on planar regions via a smoothness loss \cite{godard2017unsupervised}:
\begin{equation}
\mathcal{L}_{smooth} = \lVert\partial_u\bd{D}^*\rVert e^{-\partial_u \bd{I}^L} + \lVert\partial_v\bd{D}^*\rVert e^{-\partial_v \bd{I}^L},
\end{equation}
where $\bd{D}^*$ is a mean-normalized depth. To aggregate, our target domain loss is
\begin{equation}
	\mathcal{L}_T = \mathcal{L}_{recon} +  \kappa \mathcal{L}_{smooth},
\end{equation}
with $\kappa = 10^{-3}$.

Furthermore, we encourage the model to learn domain invariant features via adversarial feature alignment.
We use the standard adversarial loss \cite{goodfellow2014generative}:
\begin{equation}
\begin{aligned}
	\mathcal{L}_{adv} = & -\mathbb{E}_{I\sim p(\psi_S)}[\log(\mathcal{Q}(\mathcal{F}(\bd{I})))] \\
	& -\mathbb{E}_ {I\sim p(\psi_T)}[\log(1 - \mathcal{Q}(\mathcal{F}(\bd{I})))],
	\label{eqn:adv}
\end{aligned}
\end{equation}
where $\mathcal{Q}$ represents the discriminator which tries to correctly classify whether features belong to either the target or the source domain, while the feature extractor $\mathcal{F}$ is optimized to fool the discriminator in a minimax optimization.
Similarly to \cite{gurram2021monocular}, we implement this via Gradient Reversal Layer (GRL) \cite{goodfellow2014generative}, which reverses the gradient vector during back-propagation.

Our final loss is a weighted sum of the aforementioned losses:
\begin{equation}
	\mathcal{L} = \mathcal{L}_S + \lambda_1 \mathcal{L}_T + \lambda_2 \mathcal{L}_{adv},
\end{equation}
with $\lambda_1 = 10$ and $\lambda_2 = 0.5$.

As visualized in Fig. \ref{fig:system}, our model consists of five submodules, each with the corresponding parameters: $(\theta_{enc}, \theta_{CFEB_Z}, \theta_{disc}, \theta_{CFEB_C}, \theta_{dec})$. 
Given the position of the adversarial discriminator, we can divide the parameters of the model submodules to enable optimization via Eq. \eqref{eqn:adv} : $\theta_\mathcal{F} = \{\theta_{enc}, \theta_{CFEB_Z}\}, \theta_\mathcal{Q} = \{\theta_{disc}\}, \theta_\mathcal{D} = \{\theta_{CFEB_C}, \theta_{dec}\}$.
Then, according to the proposed GRL optimization, parameters are updated as follows:
\begin{align}
\theta_\mathcal{F} &\leftarrow \theta_\mathcal{F} - \mu\left(\frac{\partial \mathcal{L}_S}{\partial \theta_\mathcal{F}} + \lambda_1 \frac{\partial \mathcal{L}_T}{\partial \theta_\mathcal{F}} - \lambda_2 \frac{\partial \mathcal{L}_{adv}}{\partial \theta_\mathcal{F}}\right), \\ 
\theta_\mathcal{Q} &\leftarrow \theta_\mathcal{Q} - \mu\frac{\partial \mathcal{L}_{adv}}{\partial \theta_\mathcal{Q}}, \\
\theta_\mathcal{D} &\leftarrow \theta_\mathcal{D} - \mu\left(\frac{\partial \mathcal{L}_S}{\partial \theta_\mathcal{D}} + \lambda_1 \frac{\partial \mathcal{L}_T}{\partial \theta_\mathcal{D}}\right),
\end{align}
where $\mu$ represents the current learning rate.

%% file: sections/experiments.tex
\subsection{Evaluation details}

\mypara{Datasets.}
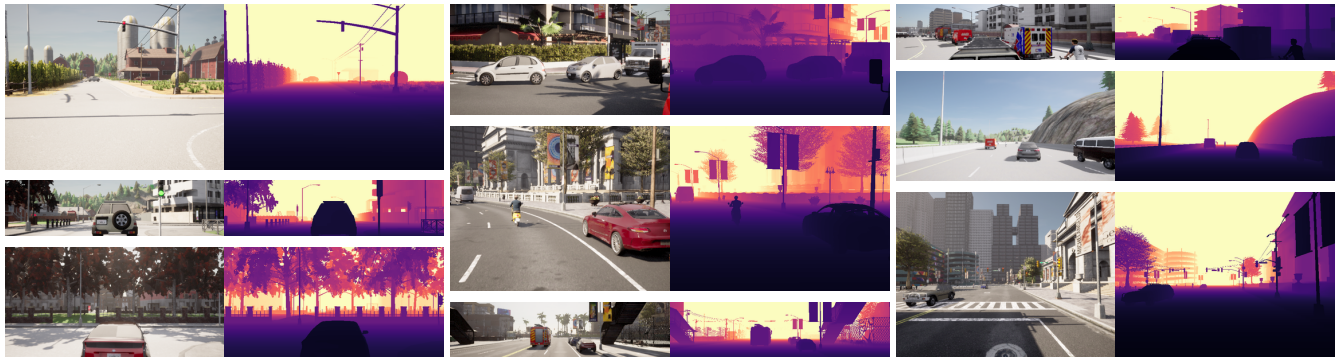
\begin{figure*}[!t]
	\centering
	\input{figures/carla/carla_fig.tex}
	\caption{Images and corresponding ground-truth depth maps from our publicly available dataset created in the CARLA simulator. Samples are acquired with a diverse set of vehicle-camera systems in urban, rural and highway environments.}
	\label{fig:carla_dataset}
\end{figure*}
To investigate the generalization ability of our method, we use several data sets with different camera configurations. 
Since our main focus is on generalization to different camera parameters, we use real-world datasets collected in similar environments, i.e., urban areas with predominantly sunny and clear weather. 
During training and testing, all datasets are downscaled to $W$ = 640, where $H$ retains the original aspect ratio. 
Table 1 lists the camera parameters for the relevant proprietary and open-source datasets used in this work. 
For datasets with different calibrations in different segments, we give mean values. The datasets used in this evaluation study are briefly presented below..

\textit{CARLA dataset}. This is an open-source dataset that we created in the CARLA simulator \cite{dosovitskiy2017carla}. We simulated autonomous driving scenarios in urban, rural and highway environments on 8 different maps Town01 - Town07, Town10HD. The maps were populated with a variety of traffic actors that were autonomously controlled in compliance with traffic rules. We simulated forward-looking RGB and depth cameras and adjusted the distortion parameters and post-processing effects to mimic the KITTI dataset \cite{geiger2013vision} as closely as possible. As shown in Fig. \ref{fig:carla_dataset}, the data was collected using different vehicle camera systems, with camera parameters randomly initialized at each data instance. This dataset serves as the primary source of data with different camera parameters, i.e., it corresponds to the $\mathcal{D}_S$ in Eq. \eqref{eqn:data}.

\textit{KITTI dataset} \cite{geiger2013vision}. This is an open-source dataset on autonomous driving in the urban area of Karlsruhe and surroundings. We use forward-facing RGB stereo cameras for training and LiDaR-generated ground-truth depths for validation. As usual in practice, we use the Eigen split \cite{eigen2015predicting}, which comprises 45200 training images.
This dataset servers as the primary source of training data from the target environmental domain, i.e., it corresponds to the $\mathcal{D}_T$ in Eq.~\eqref{eqn:data}.

\textit{DDAD dataset} \cite{guizilini20203d}. This is an open-source dataset collected in various cities in the US and Japan under challenging and different conditions. The vehicles were equipped with 6 monocular surround-view cameras and a LiDaR, which enabled the generation of corresponding ground-truth depth maps. In both training and testing, we only use frontal cameras to maintain consistency with other datasets. We use the official validation split, which amounts to 3850 samples in 50 different scenes. 

\textit{Argoverse dataset} \cite{Chang_2019_CVPR}. We use the Argoverse 1 Stereo Dataset, which includes rectified stereo images and ground-truth disparity maps collected in the Miami and Pittsburgh metropolitan areas.
Ground-truth disparity is generated using the LiDaR scene flow method \cite{pontes2020scene}, accumulating points from 11 nearby frames.
We use the official validation split with 1522 images across 17 scenes.

\textit{Waymo dataset} \cite{Sun_2020_CVPR}. We use the Waymo Perception Dataset, which consists of highly diverse autonomous driving data in various urban and suburban environments in the US.
The sensor suite includes hardware-synchronized 5 surround-view RGB cameras, 1 mid-range LiDaR and 4 short-range LiDaRs, with a diverse set of human-labeled annotations that support various tasks in semantic scene understanding and geometric perception.
For the ground-truth depth data, we project mid-range LiDaR point clouds onto the frontal RGB camera coordinates.
Since the dataset contains both nighttime and diverse weather sequences, we select a set of sunny day segments from the validation set.
This results in 8117 validation images across 100 scenes. 

\textit{nuScenes dataset} \cite{caesar2020nuscenes}. This is an open-source dataset for autonomous driving collected in urban environments in Boston and Singapore. To obtain ground-truth depth data, we project LiDaR scans onto the image coordinates using the officially provided calibration parameters. We use a small subset of the total dataset, consisting of 3000 test images taken during daytime in sunny environments.

\textit{KITTI-360 dataset} \cite{liao2022kitti}. This is a dataset collected in similar environments as KITTI, with different sensor configurations and additional annotations focusing on semantic scene understanding. We use this dataset to test the usability of GenDepth on downstream tasks. In particular, we test the performance of our monocular visual odometry (VO) algorithm MOFT \cite{koledic2023moft} on 9 provided sequences with ground-truth poses obtained by precise OxTS INS measurements.

\begin{table}[!t]
	\centering
	\caption{Camera parameters for: (i) our dataset collected in the CARLA simulator, (ii) open-source datasets collected in the real-world. To simplify the comparison, the values of $f_v$ and $c_v$ are expressed as a function of $H$.}
	\label{tbl:datasets}
	\scalebox{0.95}{
		\renewcommand{\arraystretch}{1.1}
		\begin{tabular}{c@{\hspace{1mm}}| c@{\hspace{2mm}}|c@{\hspace{1.5mm}}c@{\hspace{1.5mm}}c@{\hspace{1.5mm}}c@{\hspace{1.5mm}}c@{\hspace{1.5mm}}c}
			
			\toprule[0.7mm]
			
			& Dataset & $\alpha$[\textdegree] & $h$[m]  & $H$ [px] & $f_v$[$H$]  & $c_v$[$H$] \\
			
			\midrule
			
		 $\mathcal{E}_S$&CARLA& $[-10, 5]$ & $[1.0, 2.2]$ & $[160, 480]$ & $[0.85, 2.75]$ & $[0.4, 0.6]$ \\	
			
			\midrule
			
			&KITTI & $0.750$ & $1.650$ & $192$ & $1.925$ & $0.462$ \\
			&DDAD & $-0.511$ & $1.562$ & $384$ & $1.794$ & $0.507$ \\
			&Argoverse & $-1.085$ & $1.680$ & $448$ & $1.831$ & $0.509$ \\
			$\mathcal{E}_T$&Waymo & $-0.010$ & $2.115$ & $416$ & $1.614$ & $0.506$ \\
			&nuScenes & $-0.326$ & $1.511$ & $352$ & $1.406$ & $0.546$ \\
			&KITTI-360 & $-5.000$ & $1.550$ & $192$ & $1.469$ & $0.635$ \\
			
			\bottomrule[0.7mm]
		\end{tabular}}
\end{table}

\mypara{Evaluation metrics.}
For quantitative analysis and objective comparison, we use the established MDE metrics that have been widely used in previous works \cite{godard2019digging, guizilini20203d, wang2023planedepth}: absolute relative error (Abs Rel), relative square error (Sq Rel), root mean squared error (RMSE), root mean squared log error (RMSE log), and accuracy $\delta$ under a threshold $\tau \in \{1.25, 1.25^2, 1.25^3\}$. Detailed equations can be found in the Appendix.

\mypara{Architecture and training details.}
\begin{table*}[!t]
	\renewcommand{\arraystretch}{1.2}
	\centering
	\caption{Depth estimation results on the DDAD dataset \cite{guizilini20203d}. Dataset -- real-world data used in training. Supervision -- training loss used on the real-world data (M -- monocular supervision, S -- stereo supervision, v -- velocity supervision, D -- ground-truth depth supervision). GT align -- least squares alignment of the unknown scale and shift. The best results of the zero-shot transfer are shown in \textbf{bold}.}
	\label{tbl:ddad_results}
	\begin{tabular}{cccccccccccc}
		\toprule[0.6mm]
		
		\multirow{2}{*}[-0.4em]{Method}& \multirow{2}{*}[-0.4em]{Dataset}& \multirow{2}{*}[-0.4em]{Supervision} & \multirow{2}{*}[-0.4em]{GT align} &  \multicolumn{4}{c}{Error $\downarrow$} &  \multicolumn{3}{c}{Accuracy $\uparrow$}\\
		\cmidrule(lr){5-8} \cmidrule(lr){9-11} 	
		& & & & Abs Rel & Sq Rel & RMSE  & RMSE log & $\delta < 1.25 $ & $\delta < 1.25^{2}$ & $\delta < 1.25^{3}$\\
		\midrule[0.6mm]
		Monodepth2\cite{godard2019digging} & KITTI & M + S & \xmark & 0.378 & 5.268 & 14.606 & 0.599 & 0.201 & 0.457 & 0.740 \\
		DIFFNet \cite{zhou_diffnet} & KITTI & M + S & \xmark & 0.373 & 5.128 & 14.406 & 0.563 & 0.161 & 0.426 & 0.791 \\
		PackNet-SfM \cite{guizilini20203d} & KITTI & M + v & \xmark &  0.369 & 5.500 & 15.127 & 0.576 & 0.195 & 0.452 & 0.764 \\
		MonoViT \cite{zhao2022monovit} & KITTI & M + S & \xmark & 0.218 & 2.912 & 11.469 & 0.357 & 0.556 & 0.831 & 0.928\\
		
		\midrule
		PackNet-SfM \cite{guizilini20203d} & KITTI & D & \xmark &  0.322 & 4.834 & 14.478 & 0.536 & 0.305 & 0.617 & 0.821\\
		iDisc \cite{piccinelli2023idisc} & KITTI & D & \xmark &  0.274 & 3.993 & 13.304 & 0.461 & 0.384 & 0.743 & 0.884\\
		NeWCRFs \cite{yuan2022newcrfs} & KITTI & D & \xmark & 0.337 & 5.450 & 15.572 & 0.571 & 0.302 & 0.629 & 0.792\\
		
		\midrule
		
		RA-Depth \cite{he2022ra} & KITTI & M & \cmark & 0.165 & 1.992 & 8.979 & 0.234 & 0.775 & 0.910 & 0.960\\	
		EPCDepth \cite{peng2021excavating} & KITTI & M + S & \xmark & 0.301 & 3.956 & 12.820 & 0.469 & 0.293 & 0.686 & 0.889\\			
		PlaneDepth \cite{wang2023planedepth} & KITTI & M + S & \xmark & 0.327 & 4.760 & 13.955 & 0.529 & 0.379 & 0.467 & 0.792 \\
		
		\midrule
		
		MiDaS \cite{ranftl2020towards} & Mix & S & \cmark  & 0.150 & 1.923 & 8.428 & 0.337 & 0.803 & 0.941 & 0.979\\
		LeReS \cite{yin2021learning} & Mix & S & \cmark & 0.170 & 1.511 & 7.054 & 0.333 & 0.761 & 0.939 & 0.976 \\
		
		\midrule
		\textbf{GenDepth} & KITTI & S & \xmark & \textbf{0.121} & \textbf{1.421} & \textbf{6.992} & \textbf{0.200} & \textbf{0.840} & \textbf{0.954} & \textbf{0.983}\\
		
		\midrule		
		
		PackNet-SfM \cite{guizilini20203d} & DDAD & M & \cmark & 0.148 & 3.001 & 7.229 & 0.210 & 0.865 & 0.947 & 0.972\\
		PackNet-SfM \cite{guizilini20203d} & DDAD & D & \xmark & 0.080 & 1.202 & 6.105 & 0.170 & 0.915 & 0.966 & 0.990\\
		
		\bottomrule[0.6mm]
	\end{tabular}
\end{table*}
As advocated in Section \ref{sec:arch}, we consider the use of convolutional feature encoders, with the addition of self-attention mechanism in the deeper layers to learn the relationships between spatially distant features. 
Although our architecture is adaptable to any encoder model, in this work we use a simple ResNet 50 \cite{he2016deep} encoder to limit the computational cost and allow a fair comparison with other methods. 
Due to the architectural limitations of the encoder, we enforce $H$ and $W$ to be multiples of 32. 
Following the encoder, our attention modules $\operatorname{CFEB}_Z$ and $\operatorname{CFEB}_C$ consist of two transformer blocks, with channel dimensions $\{1024, 512\}$ and $\{512, 256\}$, respectively. 
The stride $s$ of depth-wise convolutions before the key and value projections is set to 4. 
For our camera embeddings $\bd{E}$, we use $k = 8$ frequency bins.
The decoder architecture is similar to \cite{godard2019digging}, with channel dimensions $\{128, 64, 16\}$ for the respective upsampling layers.
We use GELU \cite{hendrycks2016gaussian} activation functions in all modules after the encoder.
Finally, we estimate the depth map $\mathbf{D} \in \left[0, D_{max}\right]^{H \times W}$, where we set $D_{max} = 80$ for all datasets.

As is common in MDE practice, the encoder is initialized with pretrained ImageNet \cite{russakovsky2015imagenet} weights, while the other layers are initialized randomly. 
The model is trained with the Adam optimizer \cite{kingma2014adam}, with exponential decay rates $\beta_1 = 0.9$ and $\beta_2 = 0.999$. 
For the main experiments, each batch consists of 4 samples of $\mathcal{D}_S$ (CARLA) and 4 samples of $\mathcal{D}_T$ (KITTI). 
Since our training data contains images with different resolutions that cannot be concatenated into a single tensor, we implement a sampling strategy that generates batches of images with the same resolutions. 
To distribute the different resolutions to a single parameter update, we perform gradient accumulation and update the weights after 4 successive batches with different resolutions. 
The learning rate $\mu$ is linearly decreased from $4 \times 10^{-5}$ to $4 \times 10^{-6}$. 
The training data is randomly augmented: horizontal mirroring with 50\% probability, and random color changes in brightness, contrast, saturation and hue. 
We define an epoch as a single processing of all images from $\mathcal{D}_S$. 
With such a definition, we train all models for 50 epochs. 
All models are trained and tested on a single Nvidia RTX A5000 GPU, resulting in a training time of approximately 38 hours for the main experiments, which include both $\mathcal{D}_S$ and $\mathcal{D}_T$ training data. 
We argue that this represents a relatively simple setup in terms of data and hardware capacity, further demonstrating the efficiency of our method while achieving the desired effectiveness.

\subsection{Generalization ability experiments}
In this section, we conduct a series of experiments that demonstrate the superior generalization ability of our method. 
First, we focus on the problem described in Section  \ref{sec:problem}.
From the datasets listed in Table \ref{tbl:datasets}, we use the CARLA dataset as the source dataset $\mathcal{D}_S$ and the KITTI dataset as the target dataset $\mathcal{D}_T$. 
We then tested how well our method generalizes to all vehicle camera systems in the target environmental domain $\mathcal{E}_T$, i.e., $\mathcal{D}_{Test}$ = \{DDAD, Argoverse, nuScenes, Waymo\}.

\mypara{Quantitative results on the DDAD dataset.}
Table \ref{tbl:ddad_results} shows the results of the quantitative depth estimation. 
For comparison, we select representative state-of-the-art MDE methods, which include both supervised and self-supervised approaches, and test their ability of zero-shot transfer to the DDAD dataset. 
Considering that the training datasets are collected in an environmental domain similar to DDAD, we mainly test the generalization ability for different vehicle-camera systems.
Note that some of the methods trained on the KITTI dataset provide official model weights, even though they were trained on images with different resolutions than those in  Table \ref{tbl:datasets}. 
To ensure consistency in the  domain gap of the camera parameters, we retrain all methods on 640x192 images. 
Further results with median scaling and a detailed description of the model architectures can be found in the Appendix.

As shown in Table \ref{tbl:ddad_results}, all self-supervised methods trained on KITTI perform poorly when tested on the DDAD dataset. 
This is particularly evident when compared to the KITTI test results in the corresponding publications.
As originally suggested in Section \ref{sec:params}, the main cause of such a performance drop is the overfitting to the perspective geometry of the training dataset. 
Interestingly, Monovit \cite{zhao2022monovit}, which utilizes self-attention in the encoding stage, exhibits relatively better robustness compared to purely convolutional methods \cite{godard2019digging, zhou_diffnet, guizilini20203d}. 
Moreover, the same performance degradation is also observed for methods that use ground-truth depth supervision during training. 
Although such methods perform much better on the training dataset than the self-supervised approaches, they experience similar difficulties when confronted with a domain gap caused by varying camera parameters. 

We also tested methods that supposedly achieve robustness for a specific subset of camera parameter variations by incorporating training data augmentations \cite{he2022ra, peng2021excavating, wang2023planedepth}, which should comparatively improve the performance. 
First, we tested RA-Depth \cite{he2022ra}, which augments the data to stimulate training with varying intrinsics. 
However, they only train on monocular images, resulting in depth maps of unknown scale. 
Therefore, we argue that this method is not truly adaptable to different image resolutions, as it cannot estimate metrically accurate depth maps. 
Even if the estimated depth maps are aligned with the ground-truth data, the error of RA-Depth is significantly higher than that of GenDepth. 
PlaneDepth \cite{wang2023planedepth} also uses a similar augmentation strategy that enables metrically accurate estimation for different resolutions. 
However, this does not translate well when tested on the DDAD dataset, as exhibited by the performance drop similar to \cite{godard2019digging, zhou_diffnet, zhao2022monovit}. 
We believe that there are two main resons for this: (i) training data augmentation has a limited capacity, e.g., it can not increase FoV or simulate extrinsic variations, (ii) the generalization ability that can be achieved with such augmented data is not easily transferable to other domains, as we show in Section \ref{sec:real_ablations}. 
Furthermore, we compare the performance of EPCDepth \cite{peng2021excavating}, which uses data grafting in order to encourage the model to focus on the relative object size cue. 
Again, the results show that this is not a viable strategy for zero-shot transfer.

Finally, we compare our method with the state-of-the-art relative depth estimation methods MiDaS \cite{ranftl2020towards} and LeReS \cite{yin2021learning}. 
These methods estimate depth up to unknown scale and shift due to the use of diverse datasets during training, thereby requiring alignment to the ground-truth data during testing. 
In contrast, GenDepth achieves much more accurate depth estimation without any test-time alignment. 
We argue that this is primarily enabled by the additional information added by our ground plane embeddings. 
Finally, to contextualize our results, we report the results for PackNet-SfM \cite{guizilini20203d} trained on the DDAD dataset.
Although we do not use any DDAD data during training, GenDepth outperforms the self-supervised PackNet-SfM, which is conceptually similar due to the absence of real-world GT data during training.

\begin{table}[!t]
	\renewcommand{\arraystretch}{1.2}
	\centering
	\caption{Depth estimation results on the Argoverse, Waymo and nuScenes datasets. All method configurations are the same as in Table \ref{tbl:ddad_results}.}
	\label{tbl:waymo_results}
	\setlength{\tabcolsep}{4pt}
	\begin{tabular}{ccccccc}
		\toprule[0.6mm]
		
		& Method & Abs Rel $\downarrow$ & RMSE $\downarrow$ & RMSE log $\downarrow$ & $\delta < 1.25 \uparrow$&\\
		
		\midrule[0.6mm]
		
		\parbox[t]{3mm}{\multirow{5}{*}{\rotatebox[origin=c]{90}{Argoverse \hspace{0.1cm}}}} & Monodepth2 & 0.501 & 17.210 & 0.800 & 0.090 \\
		& iDisc & 0.489 & 16.722 & 0.774 & 0.042 \\
		& PlaneDepth & 0.483 & 17.059 & 0.838 & 0.153 \\
		& LeReS & 0.138 & \textbf{5.294} & 0.202 & 0.828 \\
		\cmidrule{2-7}
		& \textbf{GenDepth} & \textbf{0.135} & 6.520 & \textbf{0.190} & \textbf{0.840} \\
		
		\midrule[0.6mm]
		
		\parbox[t]{3mm}{\multirow{5}{*}{\rotatebox[origin=c]{90}{Waymo \hspace{0.1cm}}}} & Monodepth2 & 0.399 & 13.626 & 0.618 & 0.167 \\
		& iDisc & 0.372 & 13.913 & 0.551 & 0.131 \\
		& PlaneDepth & 0.335 & 13.932 & 0.578 & 0.427 \\
		& LeReS & 0.192 & \textbf{6.277} & 0.329 & 0.720 \\
		\cmidrule{2-7}
		& \textbf{GenDepth} & \textbf{0.154} & 7.680 & \textbf{0.213} & \textbf{0.828} \\
		
		\midrule[0.6mm]
		
		\parbox[t]{3mm}{\multirow{5}{*}{\rotatebox[origin=c]{90}{nuScenes \hspace{0.1cm}}}} & Monodepth2 & 0.281 & 11.109 & 0.456 & 0.478 \\
		& iDisc & 0.262 & 11.952 & 0.492 & 0.572 \\
		& PlaneDepth & 0.295 & 12.562 & 0.508 & 0.455 \\
		& LeReS  & 0.202 & \textbf{6.458} & 0.300 & 0.688 \\
		\cmidrule{2-7}
		& \textbf{GenDepth} & \textbf{0.175} & 7.980 & \textbf{0.256} & \textbf{0.752} \\
		\bottomrule[0.6mm]
	\end{tabular}
\end{table}

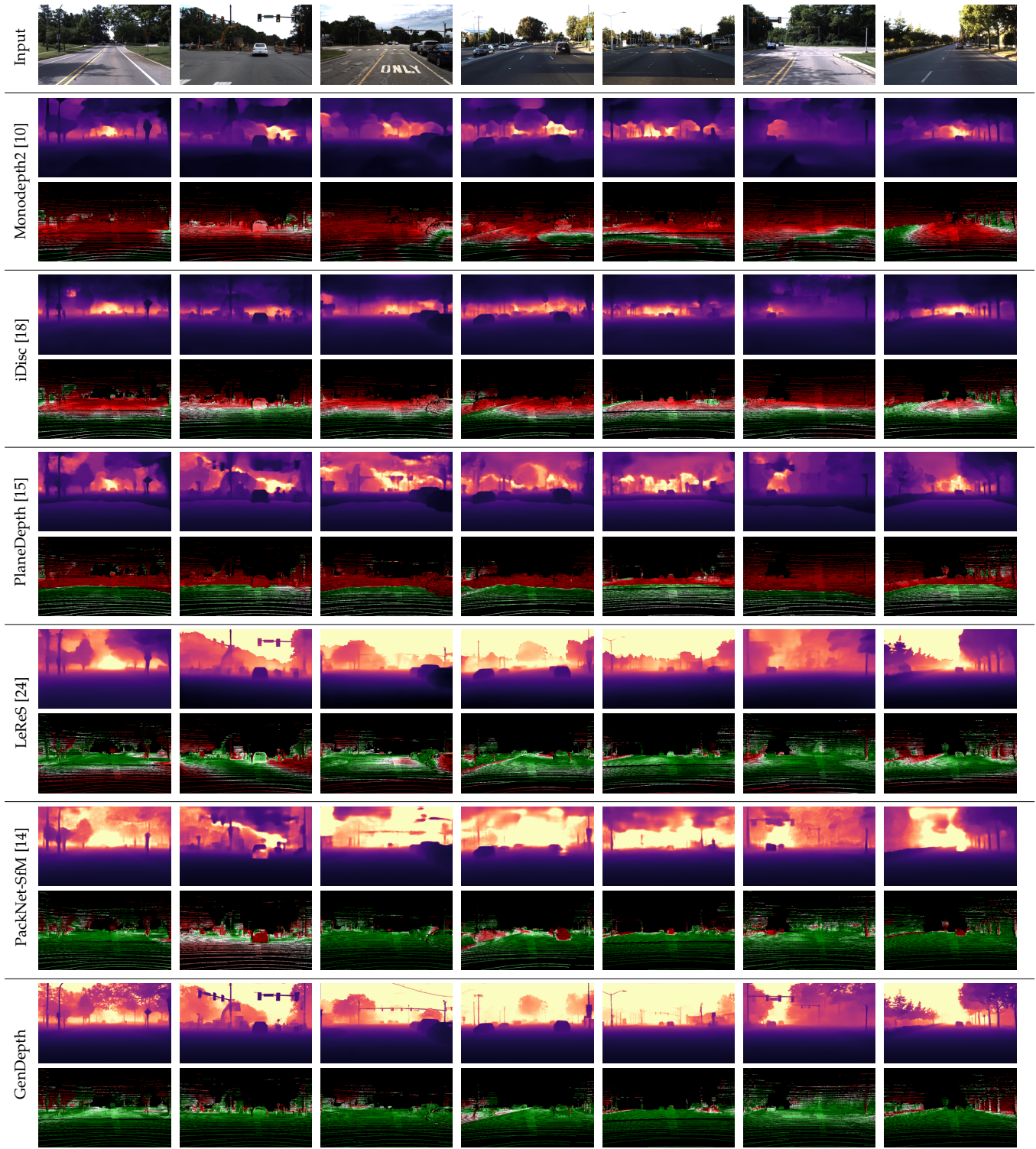
\begin{figure*}[!t]
	\centering
	\resizebox{\linewidth}{!}{\input{figures/ddad/ddad_fig.tex}}
	\caption{Qualitative results on the DDAD dataset, including both estimated depths (upper row) and Abs Rel error maps (lower row).    \textcolor{mygreen}{Green} -- low error, \textcolor{red}{red} -- high error. The only method that uses DDAD data during training is self-supervised PackNet-SfM.}
	\label{fig:ddad_results}
\end{figure*}

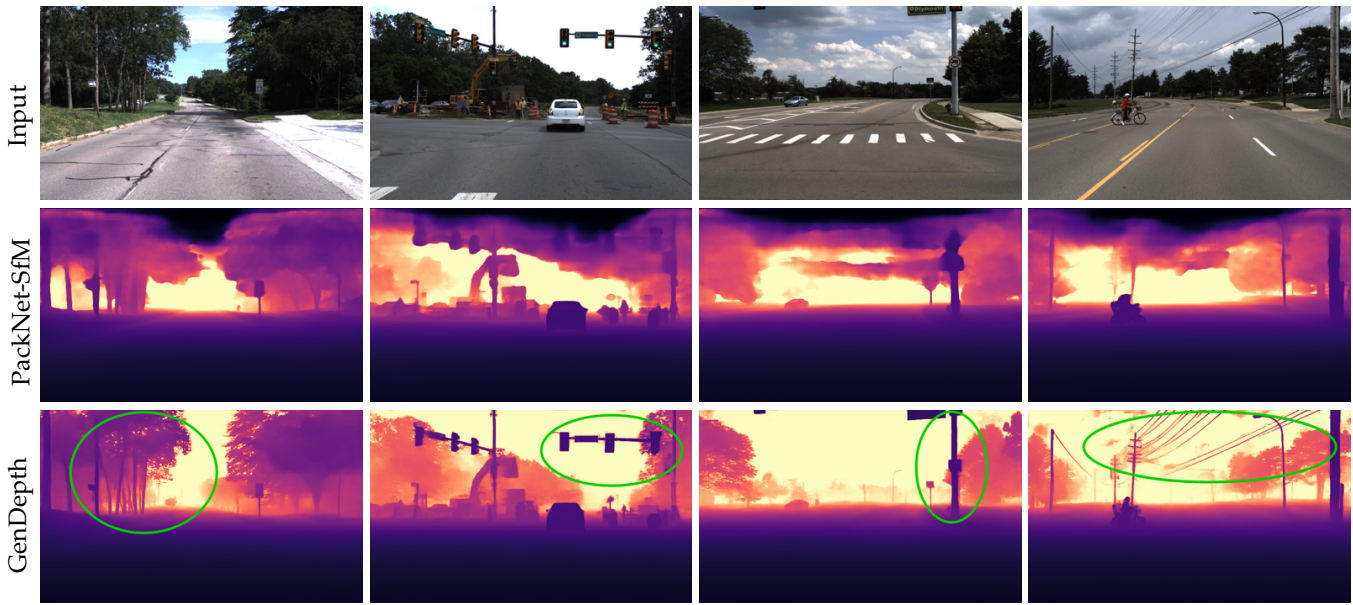
\begin{figure*}[!t]
	\centering
	{\input{figures/ddad_details/ddad_details.tex}}
	\caption{Examples of highly-detailed fine depth predictions on the DDAD dataset. We compare our method with the ground-truth supervised PackNet-SfM trained on the DDAD data.}
	\label{fig:ddad_details}
\end{figure*}

\begin{figure}[!t]
	\centering
	\resizebox{0.9\columnwidth}{!}{
		\begin{tabular}{@{\hskip 0mm}c@{\hskip 1mm}c@{\hskip 0mm}}
			\includegraphics[width=0.48\columnwidth]{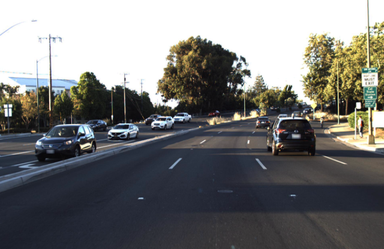}&
			\includegraphics[width=0.48\columnwidth]{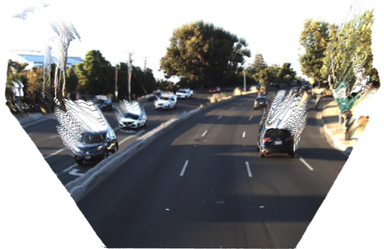} \\ 
			\includegraphics[width=0.48\columnwidth]{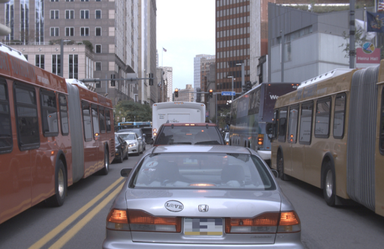}&
			\includegraphics[width=0.48\columnwidth]{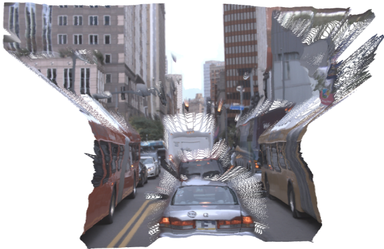} \\
			\includegraphics[width=0.48\columnwidth]{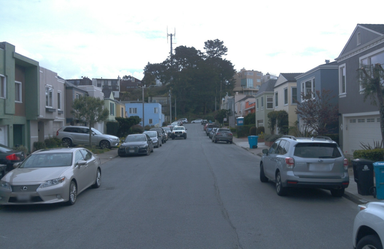}&
			\includegraphics[width=0.48\columnwidth]{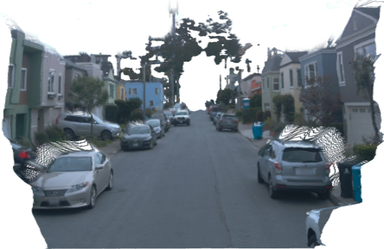}
	\end{tabular}}
	\caption{Point cloud reconstructions on the DDAD, Argoverse and Waymo datasets. GenDepth method estimates consistent metric depths, enabling authentic 3D reconstruction.}
	\label{fig:pointclouds}
\end{figure}
\mypara{Quantitative results on other datasets.} We continue with our quantitative evaluation on the Argoverse, Waymo and nuScenes datasets. We selected representative methods and evaluated their capability of zero-shot estimation on the respective datasets in Table \ref{tbl:waymo_results}.
Note that all the training data and post-processing configurations were the same as in Table \ref{tbl:ddad_results}.
Again, we found that the traditional methods exhibit strong performance degradation due to variations in perspective geometry.
The only method that is reasonably competitive is LeReS, which does not predict the metric depth and therefore requires ground-truth alignment during testing.

Overall, GenDepth demonstrates an ability to generalize across domains with varying camera parameters, especially when compared to traditional MDE methods trained on a single dataset.
We achieve this with a relatively simple training and data configuration without any fine-tuning, post-processing, or retraining, while using only a single real-world dataset without ground-truth labels. 

\mypara{DDAD qualitative results.}
To test whether our superior quantitative results culminate in visually more accurate depth maps, we provide qualitative comparisons here for the DDAD dataset, while qualitative results for other datasets can be found in the Appendix. 
Fig. \ref{fig:ddad_results} shows the estimated depth maps and the corresponding error maps, where the training configurations of the methods are the same as in Table \ref{tbl:ddad_results}. 
As can be seen, all traditional methods trained on a single dataset are strongly affected by the induced domain gap. 
In contrast, as shown in the error maps, the estimated depth maps of GenDepth are highly consistent, both for the ground plane and for the objects on it.

Fig. 11 shows the ability of the proposed GenDepth method to predict depths of extremely detailed fine structures. 
We compare GenDepth with ground-truth-trained PackNet-SfM, which uses training data from the same domain, i.e., from the DDAD dataset. 
Although the GenDepth method has never seen DDAD data during training and does not use any real-world ground-truth supervision, it can estimate depth maps of higher structural quality (notice the tree leaves, traffic signs, traffic poles, and even the wires of power lines). 
This is primarily due to the precision of the dense ground-truth data from the synthetic environment.

Fig. \ref{fig:ddad_details} also shows that PackNet-SfM has considerable difficulty estimating the sky and other objects in the upper part of the image (such as traffic lights). 
This is a common problem with all methods trained with LiDaR-based ground-truth data. 
Since this data is sparse, the model cannot distinguish between pixels outside the LiDaR range and pixels for which no ground truth data is available. 
In contrast, GenDepth uses easily generated, dense ground-truth data from the simulation environment, which significantly alleviates these problems. 
The problems with depth estimation of objects in the upper half of the image are well known, but performance degradations in these regions are often neglected (e.g., the commonly used Garg crop \cite{garg2016unsupervised} masks out the upper part of the image in KITTI evaluation) as they are considered irrelevant for safety-critical operation of autonomous vehicles. 
However, we argue that the accurate overall structure of the depth map is crucial for downstream applications such as SLAM \cite{teed2021droid}, or novel view-synthesis via NeRF \cite{roessle2022dense}

In Fig. \ref{fig:pointclouds}, we present point cloud reconstructions from the higher view showing accurate 3D structure estimation for three real-world datasets. 
In general, the qualitative results show that GenDepth performs very accurate structure estimation for many objects that are not present in either the source or target datasets. 
This means that GenDepth learns meaningful structural priors and does not overfit to the structure in the training data.

\subsection{Ablations on real-world data}
\label{sec:real_ablations}

\begin{table*}[!t]
	\renewcommand{\arraystretch}{1.2}
	\centering
	\caption{Quantitative ablation results on the DDAD dataset. CARLA, KITTI, GPE, DA, SN refer to the use of CARLA data, KITTI data, ground plane embeddings, domain alignment via an adversarial discriminator and style normalization  during training.}
	\begin{tabular}{cccccccccccc}
		\toprule[0.6mm]
		
		\multirow{2}{*}[-0.4em]{CARLA}& \multirow{2}{*}[-0.4em]{KITTI}& \multirow{2}{*}[-0.4em]{GPE} & \multirow{2}{*}[-0.4em]{DA} &
		\multirow{2}{*}[-0.4em]{SN} &  \multicolumn{4}{c}{Error $\downarrow$} &  \multicolumn{3}{c}{Accuracy $\uparrow$}\\
		\cmidrule(lr){6-9} \cmidrule(lr){10-12} 	
		& & & & & Abs Rel & Sq Rel & RMSE  & RMSE log & $\delta < 1.25 $ & $\delta < 1.25^{2}$ & $\delta < 1.25^{3}$\\
		\midrule[0.6mm]
		\cmark & & & & & 0.400 & 5.846 & 15.189 & 0.590 & 0.168 & 0.410 & 0.682\\
		& \cmark & & & &0.310 & 4.820 & 14.345 & 0.482 & 0.321 & 0.703 & 0.871\\
		\cmark & \cmark & & & &0.276 & 3.290 & 11.473 & 0.391 & 0.363 & 0.711 & 0.920 \\
		\cmark &  & \cmark& & &0.182 & 2.576 & 9.987 & 0.299 & 0.701 & 0.868 & 0.938\\
		\cmark &\cmark & \cmark & & & 0.266 & 3.195 & 11.107 & 0.390 & 0.371 & 0.730 & 0.934\\
		\cmark &\cmark & & \cmark& & 0.249 & 2.689 & 10.132 & 0.345 & 0.425 & 0.811 & 0.958\\
		\cmark & \cmark& \cmark& \cmark& & 0.125 & 1.499 & 7.050 & 0.205 & 0.833 & 0.952 & 0.982\\
		\cmark & \cmark& \cmark& \cmark& \cmark & \textbf{0.121} & \textbf{1.421} & \textbf{6.992} & \textbf{0.200} & \textbf{0.840} & \textbf{0.954} & \textbf{0.983}\\
		
		\bottomrule[0.6mm]
	\end{tabular}
	
	\label{tbl:ablations}
\end{table*}

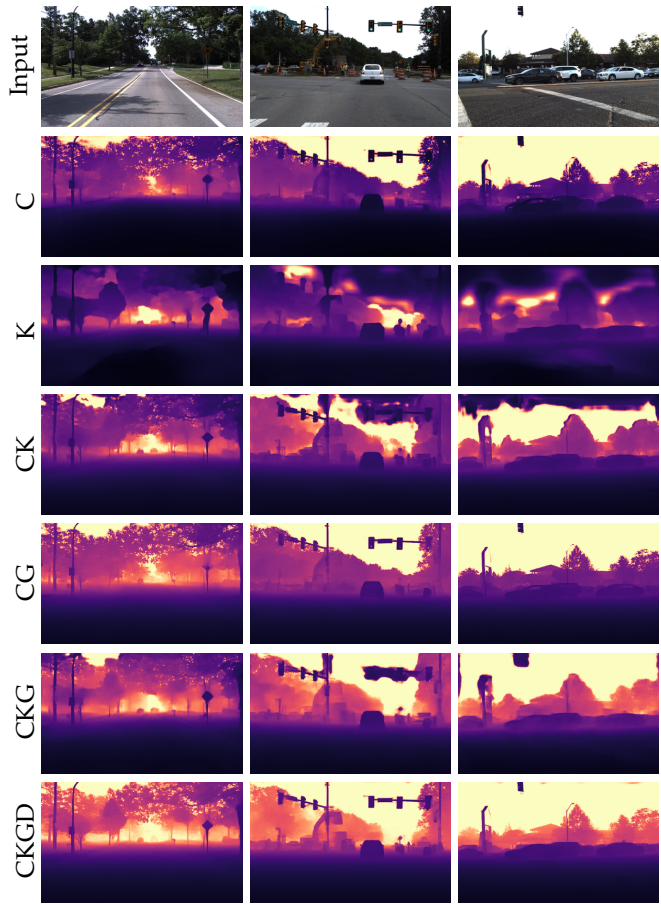
\begin{figure}[!t]
	\centering
	{\input{figures/ablations/ablations_fig.tex}}
	\caption{Qualitative ablation results. Abbreviations refer to the columns in Table \ref{tbl:ablations}, i.e., CKG = (CARLA, KITTI, GPE).}
	\label{fig:ablations}
\end{figure}

In Section \ref{sec:design} we described the possible ramifications of the different design options. 
Due to the inherent nature of our problem, generalization issues could arise due to both camera parameters and environmental domain gaps. 
These potential generalization issues are particularly evident in our training setup where we exclusively use CARLA data as the source domain data $\mathcal{D}_S$. 
Existing domain adaptation approaches in MDE \cite{zheng2018t2net, lopez2023desc} almost exclusively use the VKITTI dataset \cite{gaidon2016virtual} as the source domain, which is closer to the KITTI dataset in both style and content.

Table \ref{tbl:ablations} shows the ablation results for the DDAD dataset. 
We examine the effects of different training configurations to demonstrate the effectiveness of our contributions. 
In addition, we compare the qualitative results for the same configurations in Fig. \ref{fig:ablations}. We refer to configurations via abbreviations related to columns in Table \ref{tbl:ablations}, i.e., CKG = (CARLA, KITTI, GPE)

First, we train our model using only CARLA data, which contains images with different camera parameters. 
Surprisingly, the qualitative results show that model can estimate very fine and accurate 2D structures (C configuration), even though it has never seen real-world data during training. 
This is primarily enabled by the density of the CARLA ground-truth depth, which facilitates extremely fine detection of object boundaries. 
However, the quantitative results show that the estimated depths are largely inconsistent. 
As already suggested (Fig. \ref{fig:domains} top-left), the learned features that inform the decoder about the camera parameters are specific to the CARLA environmental domain, i.e., they do not transfer well from $\mathcal{E}_S$ to $\mathcal{E}_T$, which hampers the final estimation. 

Then, we train our model exclusively on KITTI data with stereo self-supervision loss (K configuration).
The large camera parameters domain gap between KITTI and DDAD precipitates our model to experience similar generalization issues as the other state-of-the-art methods in Table \ref{tbl:ddad_results}.
Moreover, the qualitative results in show similar patterns caused by the low precision of the self-supervised training (poor long-distance estimation, lack of fine structural details). 
As expected, adding CARLA data to the training (CK configuration) does not significantly improve the quantitative results, again as a result of issues shown in Fig. \ref{fig:domains}, top-left.
On the other hand, a relatively simple combination of CARLA data with supplementation of ground plane embedding (GPE) demonstrates impressive generalization results (CG configuration). 
The addition of GPE allows a seamless transition of camera parameters generalization capability from the source environmental domain $\mathcal{E}_S$ to the target environmental domain $\mathcal{E}_T$. 
The only issue is that the estimated depths tend to be inconsistent for some objects due to the large sim2real gap, as shown by the performance gap compared to our results with the full configuration.

Somewhat confusingly, when trying to alleviate the sim2real gap by adding the the KITTI data during training (CKG configuration), both quantitative and qualitative performance drops significantly. 
We believe that the model experiences issues as shown in Fig. \ref{fig:domains}, top-right. 
It acts discriminatively, i.e., it recognizes that the DDAD images belong to the target environmental domain $\mathcal{E}_T$ and estimates the depths as if the images were collected with the KITTI vehicle-camera system, which is a vehicle-camera system associated with the $\mathcal{E}_T$ in the training data, completely ignoring the embedded camera parameters.

We then employ domain adversarial alignment to reduce the sim2real gap. 
When the domain alignment is used without GPE (CKD configuration), the discriminator can not properly align the features due to varying camera parameters in the source dataset (see Fig. \ref{fig:domains} bottom-left). 
In contrast, careful integration with GPE (CKGD configuration) results in excellent generalization capability, both quantitatively and qualitatively. 

Finally, we examine the effects of style normalization (CKGDS configuration).
Although this simple normalization technique leads to relatively small performance improvements, we include it in our final solution since it requires only an insignificant increase in computation time during training and no modifications during inference.

\subsection{Ablations on synthetic data}
\begin{table}[!t]
	\centering
	\caption{Camera parameter configurations for several ablative datasets collected in CARLA. For ease of comparison, the values of $f_v$ and $c_v$ are expressed as a function of $H$.}
	\resizebox{\columnwidth}{!}{
		\renewcommand{\arraystretch}{1.1}
		\begin{tabular}{c@{\hspace{1.5mm}} c@{\hspace{1.5mm}}c@{\hspace{1.5mm}}c@{\hspace{1.5mm}}c@{\hspace{1.5mm}}c@{\hspace{1.5mm}}c}
			
			\toprule[0.6mm]
			
			Dataset  & $\alpha$[\textdegree] & $h$[m]  & $H$ [px] & $f_v$[$H$]  & $c_v$[$H$] \\
			
			\midrule[0.6mm]
			
			$\mathcal{B}$ & 0 & 1.5 & 320 & 1.5 & 0.5 \\ 
			
			$\mathcal{U}$ & [-10, 5] & [1.0, 2.2] & [160, 480] & [0.85, 2.75] & [0.4, 0.6] \\
			
			$\mathcal{U}_\alpha$ & [-10, 5] & 1.5& 320 & 1.5 & 0.5 \\ 
			
			$\mathcal{U}_h$ & 0 & [1.0, 2.2] & 320 & 1.5 & 0.5 \\ 
			
			$\mathcal{U}_H$ & 0 & 1.5 & [160, 480] & 1.5 & 0.5 \\ 
			
			$\mathcal{U}_{f_v}$ & 0 & 1.5 & 320 & [0.85, 2.75] & $0.5$ \\ 
			
			$\mathcal{U}_{c_v}$ & 0 & 1.5 & 320 & 1.5 & [0.4, 0.6] \\
			
			$\mathcal{D}$ & \{-10, 2.5, 5\} &  \{1, 1.6, 2.2\} & \{160, 320, 480\} & \{0.85, 1.8, 2.75\} &  \{0.4, 0.5, 0.6\}\\
			\bottomrule[0.6mm]
	\end{tabular}}
	\label{tbl:carla_dataset}
\end{table}

\begin{table*}[!t]
	\renewcommand{\arraystretch}{1.1}
	\centering
	\caption{Results of the different model and dataset configurations on synthetic data. Baseline architecture refers to the standard encoder-decoder architecture from \cite{godard2019digging} with a ResNet 50 encoder. CFEB and GPE refer to use of our Content Feature Enhancement Block and ground-plane embedding.}
	\begin{tabular}{ccccccccccc}
		\toprule[0.6mm]
		
		\multirow{2}{*}[-0.4em]{Architecture}& \multirow{2}{*}[-0.4em]{Training}& \multirow{2}{*}[-0.4em]{Testing} &  \multicolumn{4}{c}{Error $\downarrow$} &  \multicolumn{3}{c}{Accuracy $\uparrow$}\\
		\cmidrule(lr){4-7} \cmidrule(lr){8-10} 	
		& & & Abs Rel & Sq Rel & RMSE  & RMSE log & $\delta < 1.25 $ & $\delta < 1.25^{2}$ & $\delta < 1.25^{3}$\\
		\midrule[0.6mm]
		
		 Baseline & $\mathcal{B}$ & $\mathcal{U}$ & 0.301 & 2.156 & 7.500 & 0.310 & 0.475 & 0.831 & 0.942 \\
		
		Baseline & $\mathcal{U}$ & $\mathcal{U}$ & 0.071 & 0.466 & 3.990 & 0.111 & 0.956 & 0.990 & 0.995 \\
		
		 CFEB & $\mathcal{U}$ & $\mathcal{U}$ & 0.066 & 0.416 & 3.573 & 0.105 &0.959& 0.991 &0.996 \\
		
		CFEB + GPE & $\mathcal{U}$ & $\mathcal{U}$ & \textbf{0.036} & \textbf{0.356} & \textbf{3.478} & \textbf{0.085} &\textbf{ 0.974}& \textbf{0.992} & \textbf{0.996} \\
		\midrule

		Baseline & $\mathcal{B}$ & $\mathcal{U}_{\alpha}$ & 0.237 & 1.210 & 5.634 & 0.190 & 0.711 & 0.935 & 0.989 \\
		
		Baseline & $\mathcal{U}_{\alpha}$ & $\mathcal{U}_{\alpha}$ & 0.041 & 0.350 & 3.748 & 0.021 & 0.973 & 0.991 & 0.997 \\
		
		CFEB + GPE& $\mathcal{U}_{\alpha}$ & $\mathcal{U}_{\alpha}$ & \textbf{0.030} & \textbf{0.331} & \textbf{3.608} & \textbf{0.078} & \textbf{0.979} & \textbf{0.992} & \textbf{0.997} \\
		\midrule
		
		Baseline & $\mathcal{B}$ & $\mathcal{U}_{h}$ & 0.224 & 1.256 & 6.041 & 0.240 & 0.650 & 0.919 & 0.987 \\
		
		Baseline& $\mathcal{U}_{h}$ & $\mathcal{U}_{h}$ & 0.042 & 0.326 & 3.599 & 0.089 & 0.961 & 0.990 & 0.996 \\
		
		CFEB + GPE& $\mathcal{U}_{h}$ & $\mathcal{U}_{h}$ & \textbf{0.031} & \textbf{0.302} & \textbf{3.511} & \textbf{0.087} & \textbf{0.968} & \textbf{0.992} & \textbf{0.997} \\
		\midrule
		
		Baseline & $\mathcal{B}$ & $\mathcal{U}_{f_v}$ & 0.260 & 1.731 & 7.107 & 0.257 & 0.559 & 0.866 & 0.986 \\
		
		Baseline & $\mathcal{U}_{f_v}$ & $\mathcal{U}_{f_v}$ & 0.036 & 0.347 & 3.721 & 0.083 & 0.976 & 0.991 & 0.997 \\
		
		CFEB + GPE & $\mathcal{U}_{f_v}$ & $\mathcal{U}_{f_v}$ & \textbf{0.032} & \textbf{0.336} & \textbf{3.690} & \textbf{0.082} & \textbf{0.978} & \textbf{0.991} & \textbf{0.997} \\
		\midrule
		
		Baseline & $\mathcal{B}$ & $\mathcal{U}_{c_v}$ & 0.245 & 1.223 & 5.926 & 0.210 & 0.563 & 0.848 & 0.986 \\
		
		Baseline & $\mathcal{U}_{c_v}$ & $\mathcal{U}_{c_v}$ & 0.039 & 0.406 & 3.737 & 0.081 & 0.974 & 0.986 & 0.996 \\
		
		CFEB + GPE & $\mathcal{U}_{c_v}$ & $\mathcal{U}_{c_v}$ & \textbf{0.035} & \textbf{0.376} & \textbf{3.555} & \textbf{0.074} & \textbf{0.979} & \textbf{0.989} & \textbf{0.997} \\
		\midrule
		
		Baseline & $\mathcal{B}$ & $\mathcal{U}_{H}$ & 0.290 & 2.230 & 8.072 & 0.285 & 0.503 & 0.831 & 0.966 \\
		
		Baseline & $\mathcal{U}_{H}$ & $\mathcal{U}_{H}$ & 0.048 & 0.399 & 3.782 & 0.090 & 0.973 & 0.986 & 0.996 \\
		
		CFEB + GPE & $\mathcal{U}_{H}$ & $\mathcal{U}_{H}$ & \textbf{0.037} & \textbf{0.386} & \textbf{3.690} & \textbf{0.080} & \textbf{0.978} & \textbf{0.987} & \textbf{0.997} \\
		\midrule
		
		Baseline & $\mathcal{D}$ & $\mathcal{U}$ & 0.143 & 0.728 & 5.120 & 0.191 & 0.802 & 0.956 & 0.985 \\
		
		CFEB + GPE & $\mathcal{D}$ & $\mathcal{U}$ & \textbf{0.078} & \textbf{0.425} & \textbf{4.112} & \textbf{0.121} & \textbf{0.966} & \textbf{0.989} & \textbf{0.995} \\
		\bottomrule[0.6mm]
	\end{tabular}
	\label{tbl:carla_ablation}
\end{table*}

Similarly as in \cite{koledic2023towards}, we perform a series of ablative experiments on the synthetic data. 
We are mainly concerned with the following analysis: (i) To what extend do the variations of camera parameter impact the network accuracy? (ii) Can the network learn features that enable generalization for different camera parameters when the training data contains a diverse set of vehicle-camera systems? (iii) Do our proposed ground plane embeddings (GPE) and CFEB layers increase accuracy and generalization ability?

To address these concerns, we generate additional ablative datasets in the CARLA simulator with parameter configurations listed in Table \ref{tbl:carla_dataset}. 
For example, dataset $\mathcal{U}$ is acquired with random camera parameters within a certain range, sampled from the corresponding uniform distribution. 
For each dataset, we create a 90\%/10\% training and testing split. 
In particular, we perform the following experiments without the domain discriminator and the consequent domain alignment, since we only use the source domain data $\mathcal{D}_S$.

First of all, in Table \ref{tbl:carla_ablation} we show the results of the experiments with dataset $\mathcal{U}$, which contains images with different camera parameters. 
It can be seen that the baseline architecture trained on a dataset with a fixed vehicle-camera system $\mathcal{B}$ shows poor performance, which is due to the perspective geometry bias in the training data. 
Conversely, the baseline method trained on U performs surprisingly well, suggesting that the model can leverage semantic cues such as horizon level or field of view to infer informative features indicative of the geometry of the current vehicle-camera system. 
The addition of our attention-based CFEB blocks further improves the performance, especially on the outlier-sensitive metrics Sq Rel and RMSE. 
Furthermore, our proposed ground-plane embedding significantly improves the performance, proving the usefulness of the provided information.

To isolate the effects of the individual parameters, we perform a similar experiment, selectively varying only one camera parameter in the entire dataset. 
In general, the results are comparable, with a smaller difference in performance gap between the Baseline architecture and CFEB + GPE. 
Since only one parameter is varied, a model with the same capacity can more easily learn informative features specific to that camera parameter.

We would also like to discuss the results for the dataset $\mathcal{U}_{f_v}$, which contains different focal lengths. Previous works such as \cite{he2018learning, facil2019cam} often emphasize the well-known focal length depth ambiguity as a limitation that leads to models being unable to infer the metric depth when trained on datasets with varying intrinsics.
However, our results suggest otherwise, as even the Baseline architecture, which does not include embeddings that inform the network about focal length, performed well. 
We believe that the model effectively learns to estimate the sensor's field of view, which serves as an effective cue for estimating the focal length and provides a consistent solution to the ambiguity mentioned above.

Finally, we investigate whether our embedding leads to an increased generalization ability for unseen camera parameters. 
To this end, we train on the dataset $\mathcal{D}$, which includes three discrete samples of each parameter. 
Then, we evaluate the generalization performance on $\mathcal{U}$, i.e., how the model estimates the depth for parameters between these discrete samples.
Our contributions lead to significantly better quantitative results (last row in Table \ref{tbl:carla_ablation}), confirming that the model effectively learns the relationship between the provided ground-plane depth and the depth of the scene.

\subsection{Additional experiments}
\mypara{Semantic accuracy.}
\begin{figure*}
	\centering
	\includegraphics[width=\linewidth]{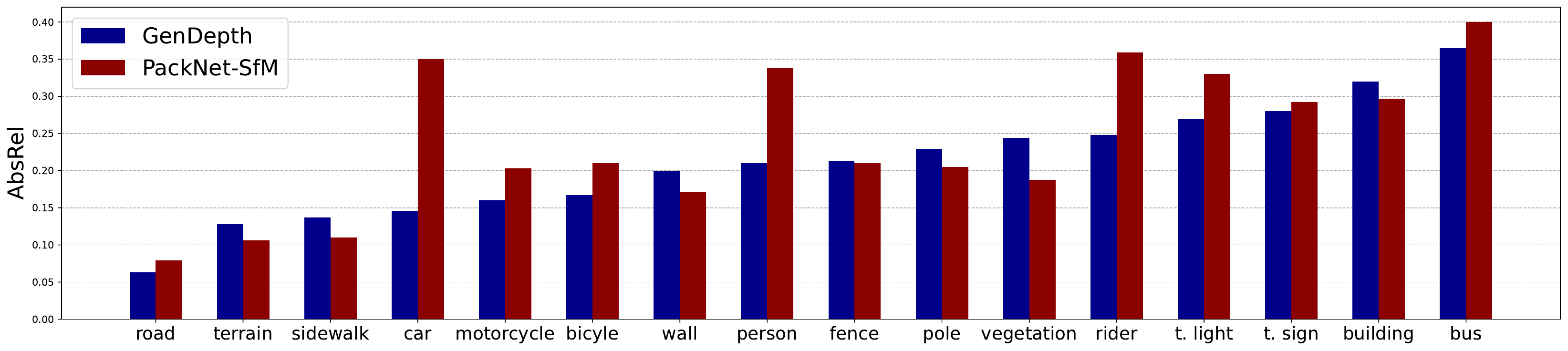}
	\caption{AbsRel errors for different semantic classes on the DDAD dataset, compared to self-supervised PackNet-SfM trained on DDAD.}
	\label{fig:semantics}
\end{figure*}
The embeddings provided to our model are in fact the expected depths of the road surface. 
A possible side effect of such a design is that the model might learn to use the embeddings only for road surface depth estimation. 
Since the road constitutes a fairly large part of the image, this would lead to acceptable quantitative performance, and hide the fact that the model does not estimate accurate depths for the important objects in the image.

Even though the error maps in Fig. \ref{fig:ddad_results} show an accurate estimation for various object classes, we decided to verify these results quantitatively.
For each image from the DDAD validation split, we determine the object class of each pixel via the EfficientPS model \cite{mohan2021efficientps} trained on the CityScapes dataset \cite{cordts2016cityscapes}.  
Accordingly, we show the depth estimation error per class in Fig. \ref{fig:semantics}.

When compared with PackNet-SfM, it is obvious that our model does not exhibit the aforementioned issues, with particular outliers worthy of detailed analysis.
First, as expected, our method performs better on the road surface. 
However, even though terrain and sidewalk are similar planar regions, GenDepth experiences a relative drop in performance. 
This suggests that it may have trouble generalizing our ground-plane embeddings to slightly elevated regions compared to the road surface. 
Most importantly, GenDepth performs exceptionally well on safety-critical objects such as vehicles and pedestrians. 
PackNet-SfM struggles to estimate such dynamic objects due to the nature of self-supervised loss that comes from static scenes. 
Interestingly, GenDepth estimates the depth of pedestrians quite well. 
This proves the effectiveness of our domain alignment strategy since our synthetic CARLA dataset does not contain models of pedestrians.

\begin{figure}
	\newcommand{\figwidth}{0.48\columnwidth}
	\centering
	\begin{tabular}{@{\hskip 0mm}c@{\hskip 1mm}c}
		
		\includegraphics[width=\figwidth]{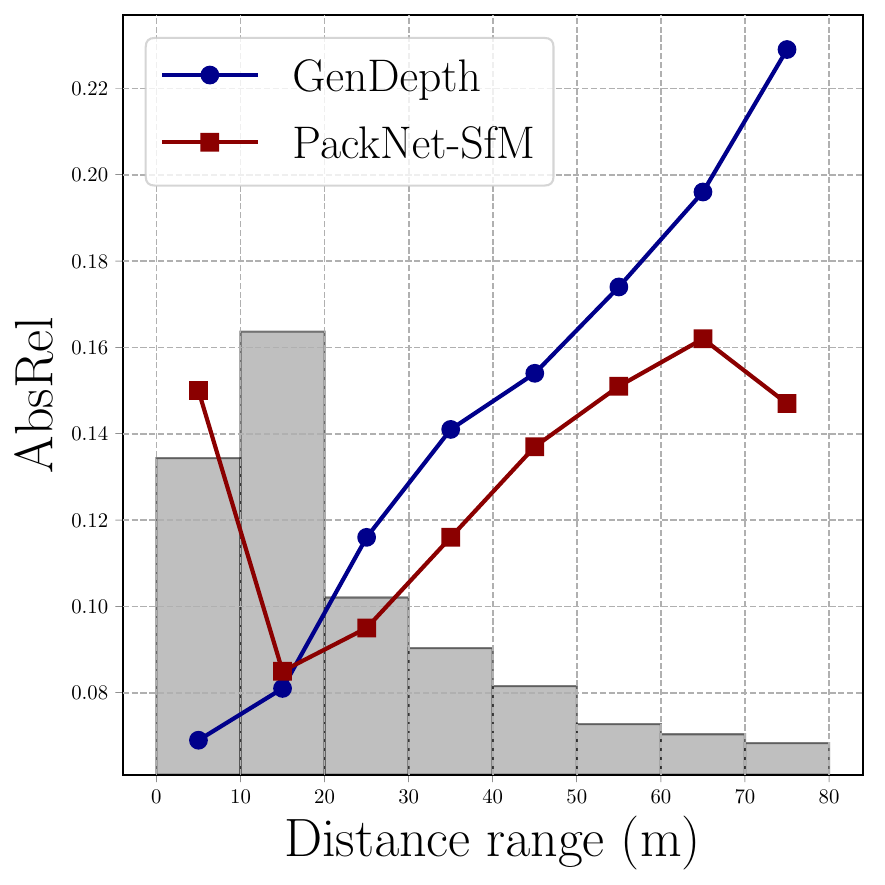} & \includegraphics[width=\figwidth]{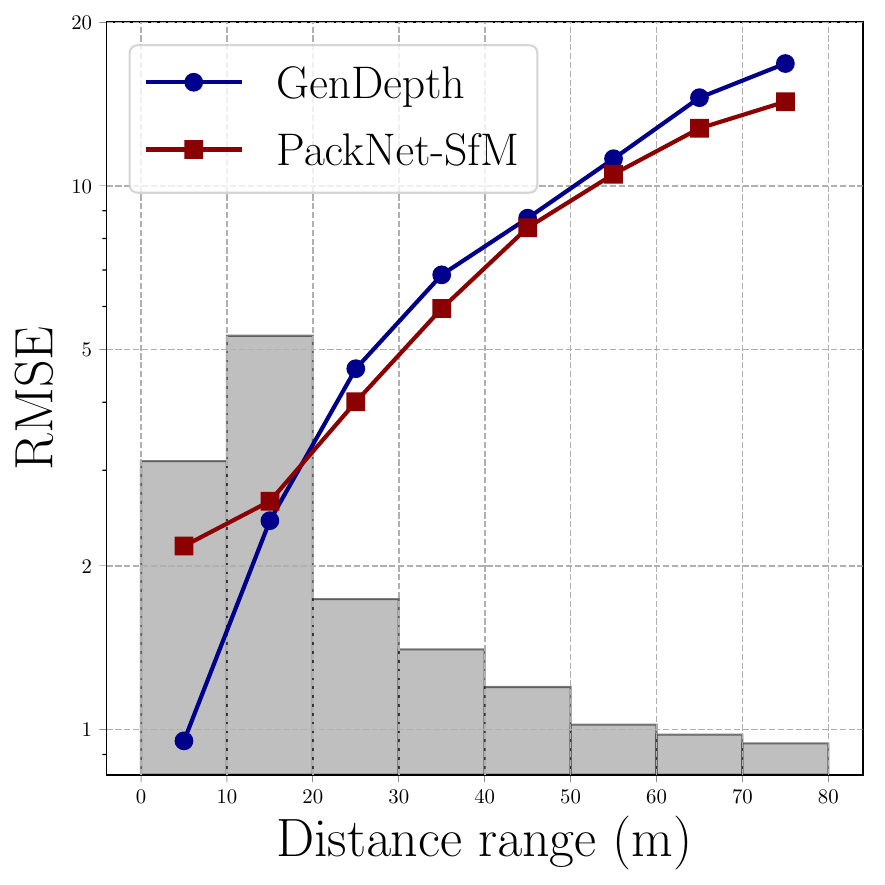}
		
	\end{tabular}
	\caption{AbsRel and RMSE for different distance ranges in the DDAD dataset. We compare with the self-supervised PackNet-SfM trained on DDAD. The histograms show the distribution of ground-truth points.}
	\label{fig:distance}
\end{figure}

\begin{figure}
	\newcommand{\figwidth}{0.48\columnwidth}
	\centering
	\begin{tabular}{@{\hskip 0mm}c@{\hskip 1mm}c}
		
		\includegraphics[width=\figwidth]{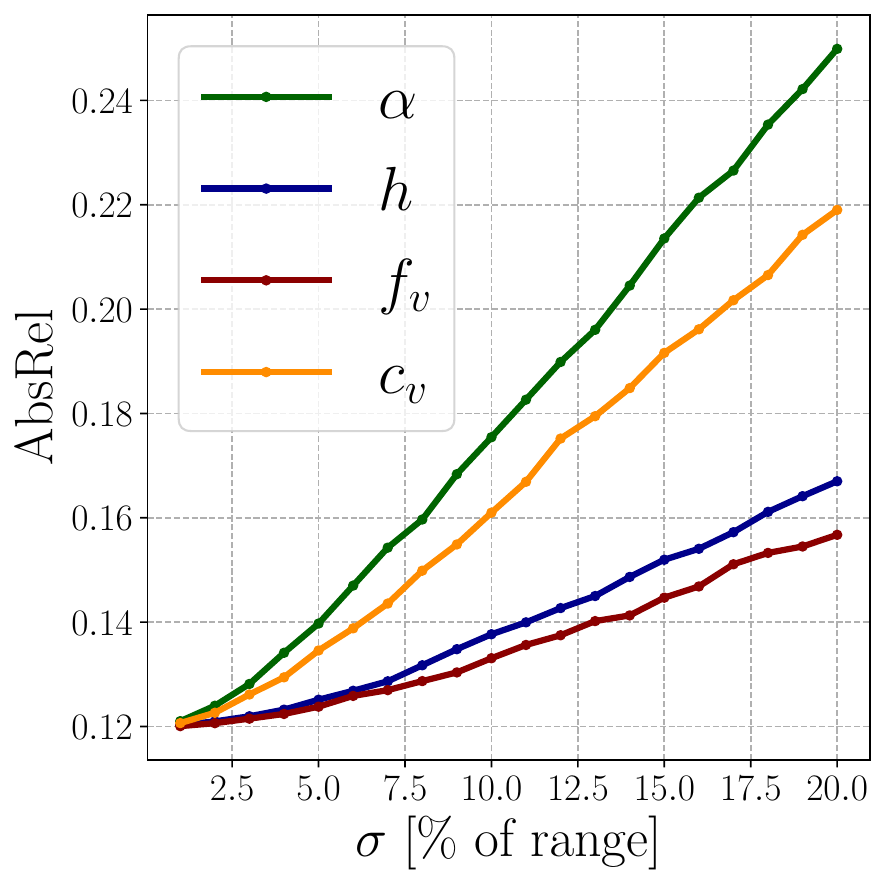} & \includegraphics[width=\figwidth]{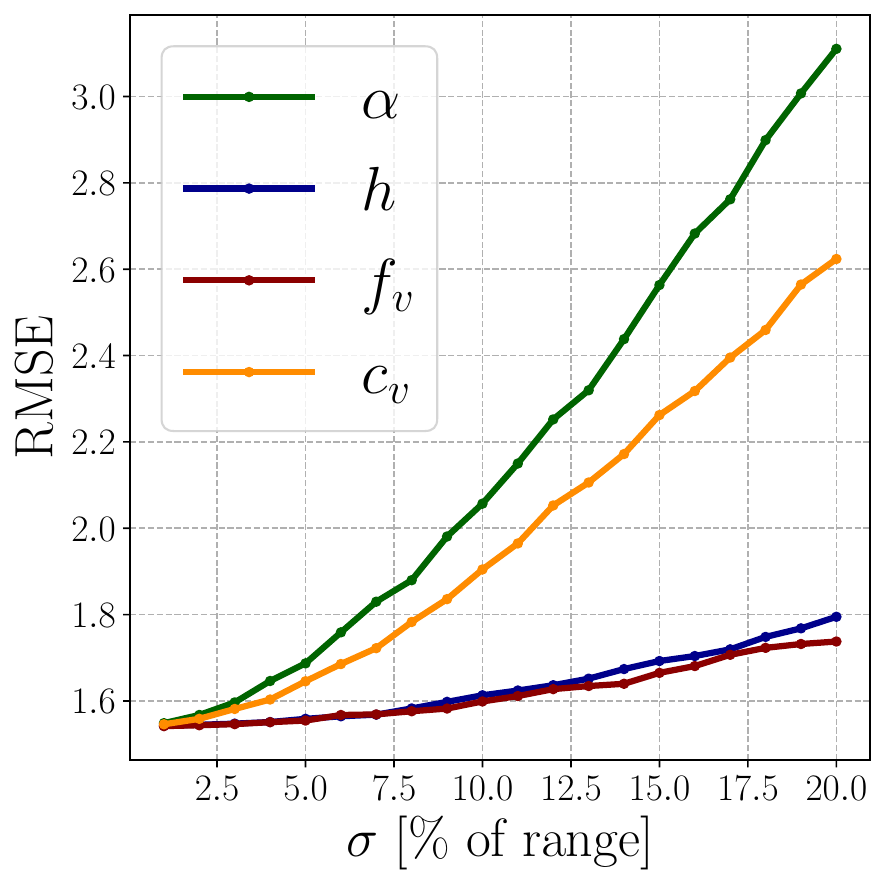}
		
	\end{tabular}
	\caption{AbsRel and RMSE for different noise levels on the DDAD dataset. Noise levels are calculated as a percentage of the admissible parameter range for our CARLA dataset, as defined in Table \ref{tbl:datasets}.}
	\label{fig:calib}
\end{figure}
\mypara{Sensitivity analysis.} Environmental perception tasks are often sensitive to the ever-changing perturbations that are not present in the training data. 
Due to its inherently geometric nature, MDE is particularly sensitive to calibration errors that violate the perspective geometry assumptions. 
Our ground plane embeddings explicitly require the accurate camera calibration during test time, with the additional assumptions of perpendicularity and planarity of the road surface. 
Any violation of these assumptions could potentially lead to metrically less accurate depth maps. 
This also applies to other models trained on the single vehicle-camera setup, albeit implicitly, as they overfit to the configuration of the training data.

The road surface perturbation has a particularly detrimental effect on the accuracy of our embeddings for the distant objects. 
Fig. \ref{fig:distance} shows the performance of GenDepth and PackNet-SfM for different distance ranges.
Note that unlike GenDepth, PackNet-SfM is trained in-domain, i.e., it has seen the DDAD data during training.
As shown, our method has a sharper error inclination as depth increases. 
However, we compensate for this with significantly better performance in the nearby regions, which are more important for safety-critical operations and constitute a higher proportion of the overall image.

Since the camera parameters can change in real time due to physical shocks or temperature fluctuations, we also analyze the calibration sensitivity.
In Fig. \ref{fig:calib}, known calibration parameters are perturbed with an additive Gaussian noise with increasing standard deviation on the x-axis. 
As expected, the errors increase in a fairly linear manner with stronger perturbations.
However, this is a standard behavior of all camera-based geometric perception algorithms and can be remedied by careful and accurate online calibration.

\mypara{Visual odometry evaluation.}
Dense monocular depth can be used as a strong source of information for many downstream tasks such as NeRFs \cite{deng2022depth, roessle2022dense}, geometric VO, and SLAM \cite{yang2018deep, yang2020d3vo,  teed2021droid}. 
It is particularly important for improving monocular VO and SLAM algorithms that suffer from excessive scale-drift and metric ambiguity. 
To test the applicability of GenDepth, we use it within our monocular VO MOFT \cite{koledic2023moft} to improve both front-end tracking and back-end optimization.
Once again, GenDepth is used in zero-shot mode, i.e., it has never seen KITTI-360 data during training.

\begin{figure}
	\newcommand{\figwidth}{0.48\columnwidth}
	\centering
	\begin{tabular}{@{\hskip 0mm}c@{\hskip 1mm}c@{\hskip 0mm}}
		\includegraphics[width=\figwidth]{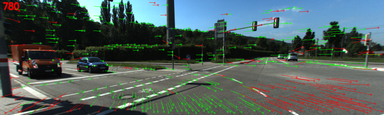} & 
		\includegraphics[width=\figwidth]{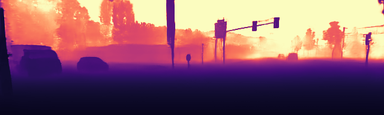}\\ 
		\includegraphics[width=\figwidth]{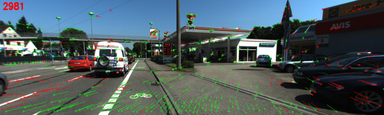} & 
		\includegraphics[width=\figwidth]{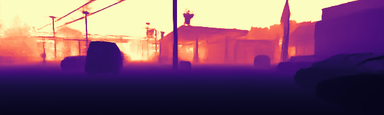} 
	\end{tabular}
	\caption{MOFT\cite{koledic2023moft} feature matches and corresponding depths on the KITTI-360 sequences. GenDepth depth estimation enhances feature tracking and enables robust matching with a high percentage of inliers (\textcolor{mygreen}{green} matches) after RANSAC optimization.}
	\label{fig:moft_features}
\end{figure}

\begin{figure}
	\newcommand{\figwidth}{0.48\columnwidth}
	\centering
	\begin{tabular}{@{\hskip 0mm}c@{\hskip 1mm}c}
		
		\includegraphics[width=\figwidth]{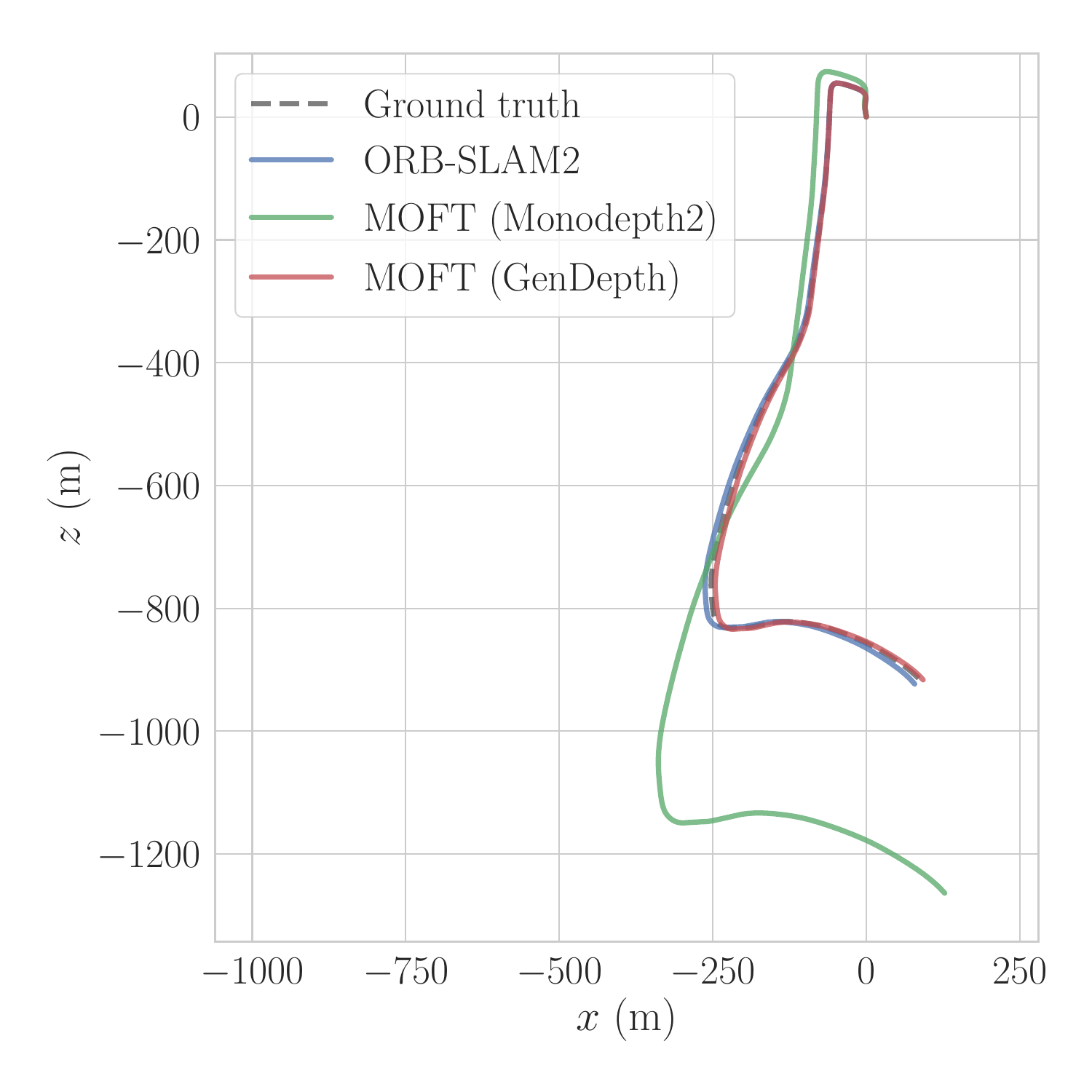} & 
		\includegraphics[width=\figwidth]{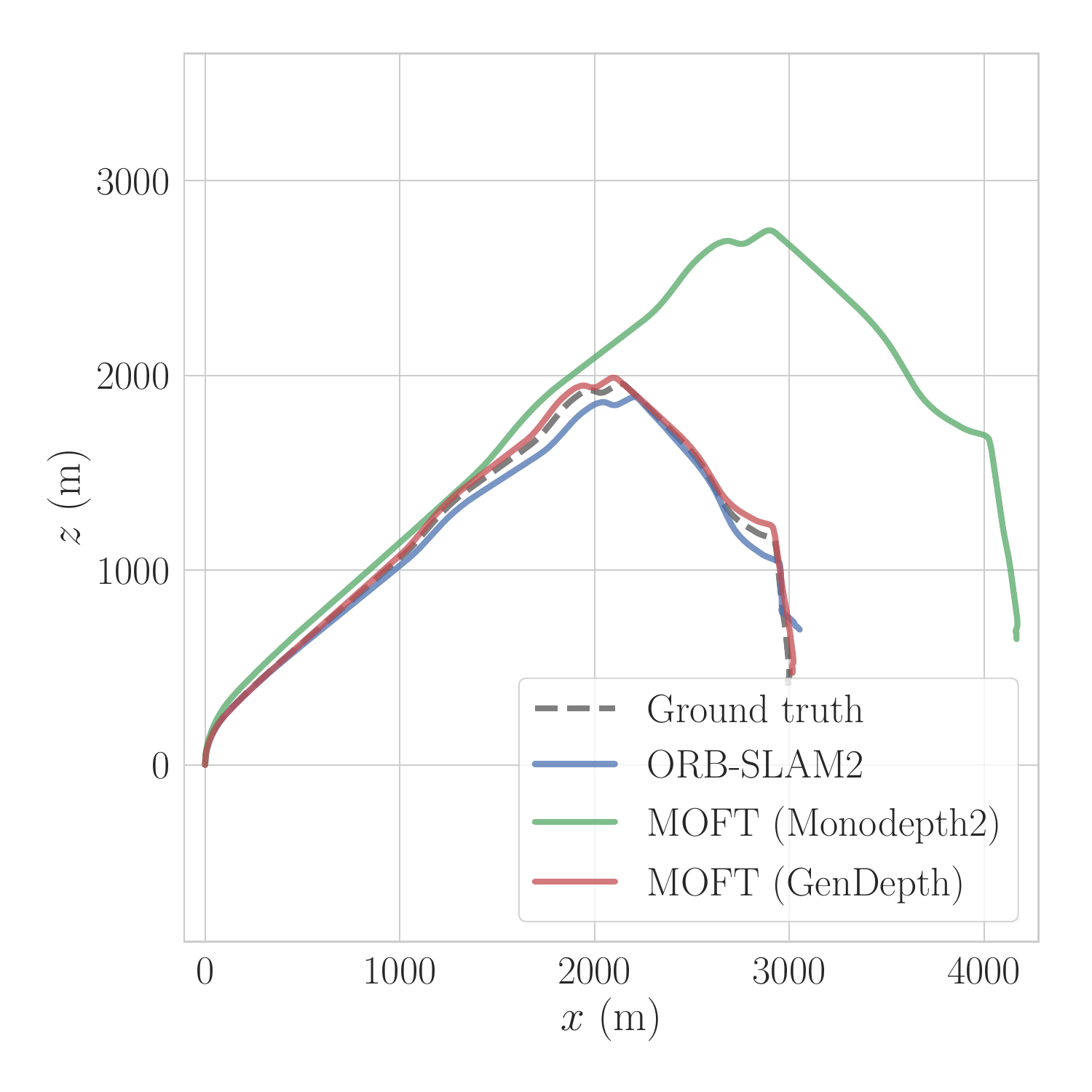} \\
		Sequence 03 & Sequence 10 \\
	\end{tabular}
	\caption{Visual odometry trajectories on the KITTI-360 dataset. The fusion of estimated depths during MOFT back-end optimization results in metrically accurate trajectories on par with stereo ORB-SLAM2.}
	\label{fig:moft_traj}
\end{figure}

\begin{table}[!t]
	\setlength{\tabcolsep}{2.8pt} 
	\renewcommand{\arraystretch}{1.3}
	\centering
	\caption{Visual odometry translation error $t_{rel}$ and rotation error $r_{rel}$ for 9 sequences of KITTI-360. GenDepth enhances geometric monocular odometry MOFT and enables results on par with stereo ORB-SLAM2.}
	\resizebox{\columnwidth}{!}{
		\begin{tabular}{cccccccccccc}
			\toprule[0.5mm]
			& & 00 & 02 & 03 & 04 & 05 & 06 & 07 & 09 & 10 \\
			\midrule
			\multirow{2}{*}{MOFT} & $t_{rel}$ & 0.41 & \textbf{0.54} & 0.68 & 0.53 & 0.55 & \textbf{0.48} & \textbf{0.66} & \textbf{0.78} & \textbf{1.31} \\
			& $r_{rel}$ & \textbf{0.14} & \textbf{0.21} & \textbf{0.15} & 0.23 & 0.25 & \textbf{0.17 }&\textbf{0.15} & \textbf{0.18} & \textbf{0.23}\\
			\multirow{2}{*}{ORB-SLAM2} & $t_{rel}$ & \textbf{0.33} & 0.58 & \textbf{0.49} & \textbf{0.52} & \textbf{0.46} & 0.52 & 5.08 & 1.07 & 1.73 \\
			& $r_{rel}$ & 0.15 & 0.23 & 0.17 &\textbf{0.22} & \textbf{0.25} & 0.18 & 0.97 & 0.18 & 0.43\\
			\bottomrule[0.5mm]
	\end{tabular}}
	
	\label{tbl:kitti360}
\end{table}

Fig. \ref{fig:moft_features} shows the detected feature matches with inliers and outliers after the RANSAC back-end optimization. 
It is evident that the use of estimated depth during feature tracking facilitates highly accurate and robust matching.
This enables MOFT to function as a simple frame-to-frame monocular VO, without the complex monocular multi-frame matching that often induces severe scale-drift in the resulting trajectory.

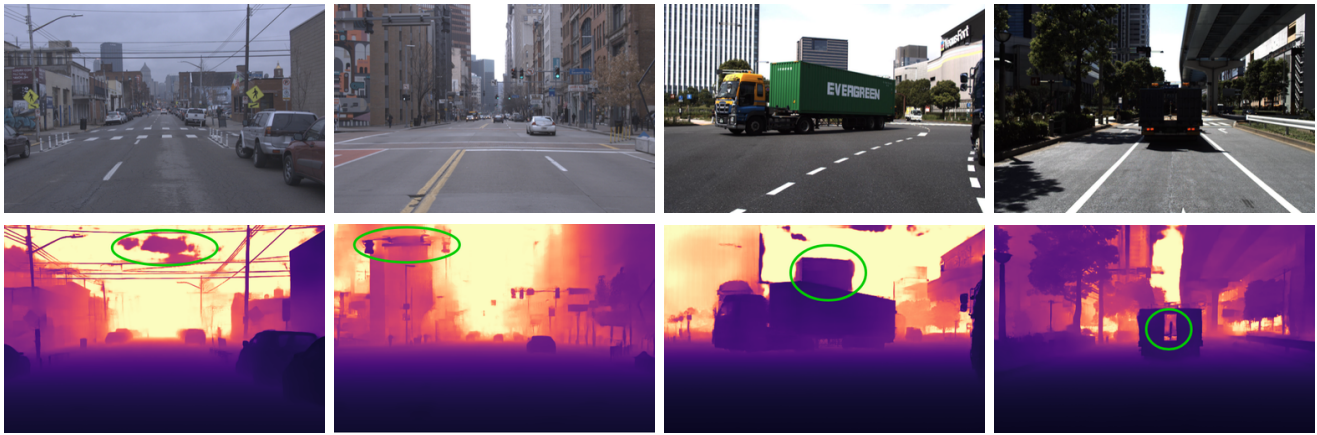
\begin{figure*}[!t]
	\centering
	\resizebox{\linewidth}{!}{
		{\input{figures/failures/failures_fig.tex}}}
	\caption{Examples of error cases of our model.}
	\label{fig:failures}
\end{figure*}

\begin{figure}[!t]
	\centering
	\resizebox{\columnwidth}{!}{
		\begin{tabular}{@{\hskip 0mm}c@{\hskip 1mm}c@{\hskip 1mm}c@{\hskip 0mm}}
			\includegraphics[width=0.3\columnwidth]{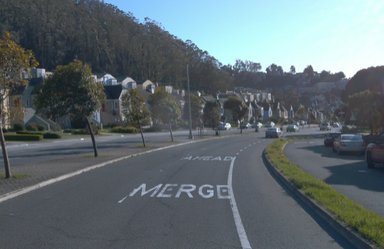}&
			\includegraphics[width=0.3\columnwidth]{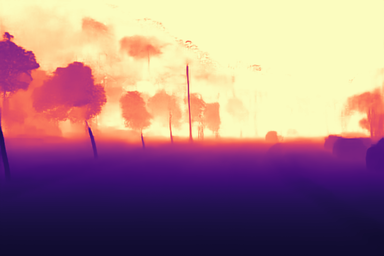}&
			\includegraphics[width=0.3\linewidth]{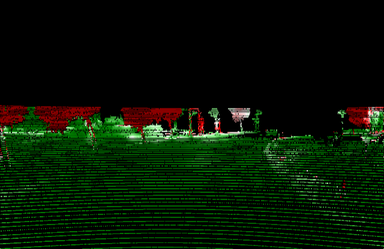} \\
			
			\includegraphics[width=0.3\columnwidth]{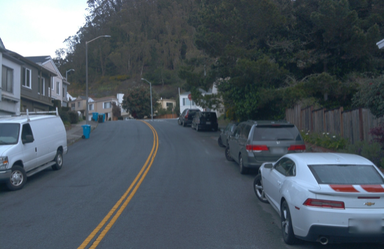}&
			\includegraphics[width=0.3\columnwidth]{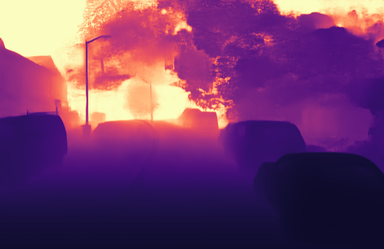}&
			\includegraphics[width=0.3\columnwidth]{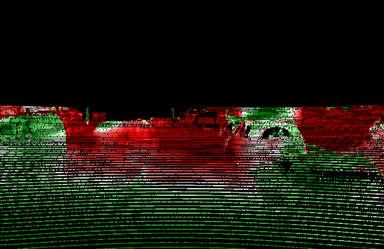}
	\end{tabular}}
	\caption{Adaptability to scenes that violate the assumption of an ideal orthogonal ground plane. Our model can handle moderate slopes in the training data, but performs poorly on images with severe violations.}
	\label{fig:plane}
\end{figure}
We also use the estimated depth during back-end optimization, which gives us metrically accurate trajectories, as shown in Fig. \ref{fig:moft_traj}. 
MOFT (GenDepth) performs significantly better than MOFT (Monodepth2), which uses Monodepth2 \cite{godard2019digging} trained on KITTI. 
The latter experiences a high degree of drift, as Monodepth2 is not able to overcome the camera parameters domain gap between KITTI and KITTI-360.
Finally, in Table \ref{tbl:kitti360}, we evaluate the GenDepth-enhanced MOFT with standard odometry metrics \cite{geiger2013vision}.
MOFT shows comparable results with stereo ORB-SLAM2, while it is a simple frame-to-frame monocular VO.

\subsection{Limitations and discussion}
MDE methods often experience intricate failures due to the natural complexity of the task. 
In our case, we are additionally dealing with domain gaps induced by both the changes in perspective geometry due to the varying camera parameters and environmental changes due to sim2real adaptation. 
Although the latter has already been thoroughly explored in the literature \cite{atapour2018real, zhao2019geometry, gurram2021monocular}, it is especially prominent in our setup due to the relative photo-realistic simplicity of the CARLA simulator.

In Fig. \ref{fig:failures} we present several error cases of our method. 
For example, the first image shows a very cloudy sky, which is in contrast to our CARLA dataset that includes mostly clear weather. 
This large domain gap leads to an incorrect depth estimation for the clouds. 
Another interesting example is presented in the third image. 
The model is unable to recognize the background building as a separate object, resulting in a depth estimate as if the building was part of the truck. 
The model essentially overfits to the vertical image position cue, which may have been caused by the design of our embeddings.
However, our embeddings do not force the model to use the vertical position cue exclusively, but merely encode the camera parameters so that they can be used for different vehicle-camera systems. 
In this way, the model should make an informed decision depending on the object class and position.
Instead, what happens is probably a domain gap issue; the model has not seen similar buildings in the training data, which would enable their recognition and delineation from foreground objects.
Nonetheless, our model generally does not encounter artifacts that are not omnipresent in other MDE methods. 
Although some of these artifacts are induced by the large sim2real domain gap, we argue that the potential benefits of the synthetic data far outweigh these problems. 

Another possible limitation is that we rely on the ideal flatness of the road surface. 
Fig. \ref{fig:plane} shows our results for images with high slopes of the ground plane. 
GenDepth can easily adapt to the moderate slopes that are present in the training data. 
However, severe slopes lead to a rather inconsistent estimation of the depth map. 
This can be alleviated by a more diverse data configuration or a simple online estimation of the ground plane.

We would also like to address the problem of erroneous calibration. 
As already mentioned, all MDE methods are prone to these errors. 
Other works unfortunately often do not take them into account, as they train and test the models on the same vehicle-camera systems, unknowingly overfitting to erroneous calibrations without any impact on the quantitative performance. 
This is particularly the case with methods that use LiDaR ground-truth data for training. 
In addition to the erroneous camera calibrations, these methods overfit to errors in multi-sensor calibrations, used to project the LiDaR scans to camera coordinates.
Although this problem is already known \cite{cvivsic2021recalibrating}, we have observed a particularly severe case of this problem in the KITTI dataset, which is usually a primary training and testing dataset for many state-of-the-art methods \cite{piccinelli2023idisc, wang2023planedepth}. 
On the other hand, our model uses synthetic data that naturally has proper camera calibration and ground-truth data, which avoids overfitting to the erroneous calibrations of the specific camera model.
This leads to rather confusing negative ramifications; when evaluated on popular real-world benchmarks, the results of our model may appear worse than they actually are.

Overall, we have shown impressive generalization results despite the large domain gap, the relatively simple data configuration, and the domain alignment procedure.
Furthermore, we have demonstrated the advantages of a dense ground-truth depth that allows accurate estimation of complicated structures compared to methods trained on real-world data, especially for image regions without LiDaR scans.
This leads us to believe that the synthetic datasets, together with the proper use of domain adaptation and domain generalization techniques, are far underutilized in the current MDE literature dealing with autonomous driving scenarios.
Existing approaches to domain adaptation \cite{atapour2018real, kundu2018adadepth, zheng2018t2net, zhao2019geometry, akada2022self, lo2022learning, gurram2021monocular, pnvr2020sharingan, lopez2023desc} are almost exclusively limited to the VKITTI $\rightarrow$ KITTI configuration. 
On the other hand, domain generalization approaches \cite{yin2021learning, ranftl2020towards} merely train on a large-scale real-world data without ground-truth annotations, which means that the depths can only be estimated up to an unknown scale and shift.
This is in contrast to works in other fields, such as semantic segmentation, which have an established and proven method for robust domain generalization \cite{peng2022semantic, zhao2023style}. 
Therefore, we believe that future MDE research should follow similar avenues.

%% file: figures/carla/carla_fig.tex
\newcommand{\newwidth}{0.15\textwidth}
\centering

\begin{minipage}{0.32\textwidth}
	\begin{subfigure}{.5\textwidth}
		\centering
		\includegraphics[width=\linewidth]{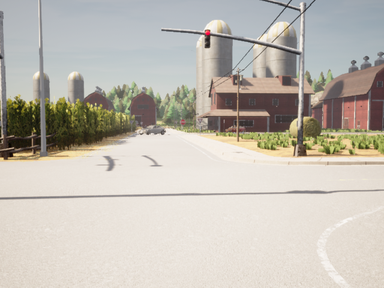}
	\end{subfigure}%
	\begin{subfigure}{.5\textwidth}
		\centering
		\includegraphics[width=\linewidth]{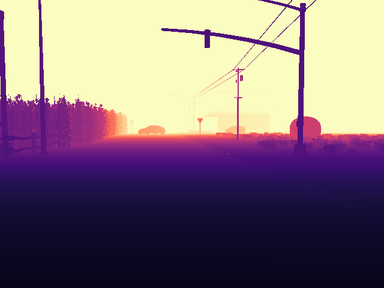}
	\end{subfigure}
	\\[1ex]
	\begin{subfigure}{.5\textwidth}
	\centering
	\includegraphics[width=\linewidth]{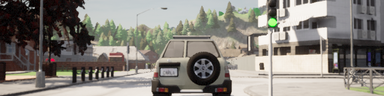}
	\end{subfigure}%
	\begin{subfigure}{.5\textwidth}
		\centering
		\includegraphics[width=\linewidth]{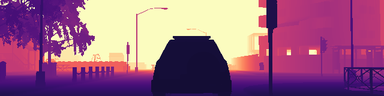}
	\end{subfigure}
		\\[1ex]
	\begin{subfigure}{.5\textwidth}
		\centering
		\includegraphics[width=\linewidth]{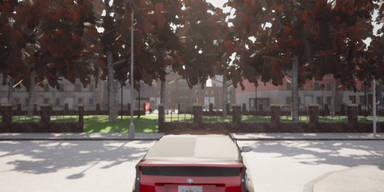}
	\end{subfigure}%
	\begin{subfigure}{.5\textwidth}
		\centering
		\includegraphics[width=\linewidth]{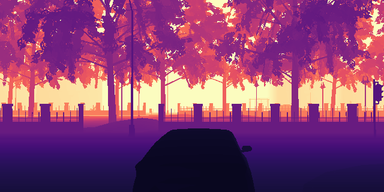}
	\end{subfigure}
\end{minipage}
\begin{minipage}{0.32\textwidth}
	\begin{subfigure}{.5\textwidth}
		\centering
		\includegraphics[width=\linewidth]{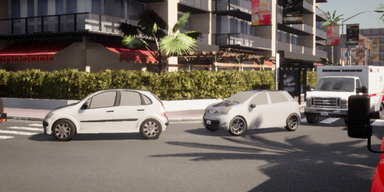}
	\end{subfigure}%
	\begin{subfigure}{.5\textwidth}
		\centering
		\includegraphics[width=\linewidth]{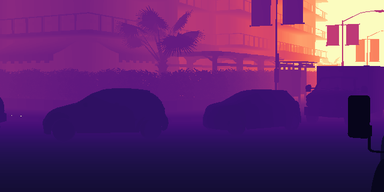}
	\end{subfigure}
	\\[1ex]
	\begin{subfigure}{.5\textwidth}
		\centering
		\includegraphics[width=\linewidth]{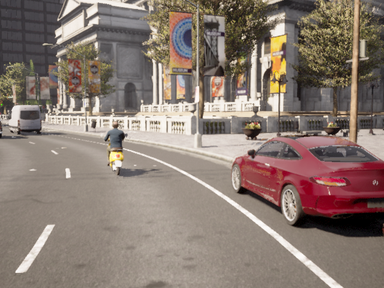}
	\end{subfigure}%
	\begin{subfigure}{.5\textwidth}
		\centering
		\includegraphics[width=\linewidth]{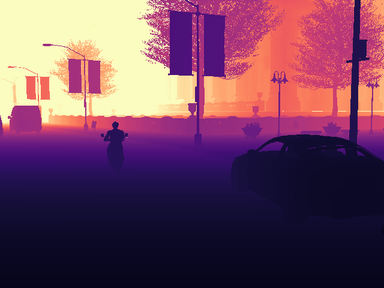}
	\end{subfigure}
	\\[1ex]
	\begin{subfigure}{.5\textwidth}
		\centering
		\includegraphics[width=\linewidth]{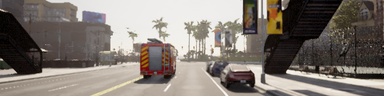}
	\end{subfigure}%
	\begin{subfigure}{.5\textwidth}
		\centering
		\includegraphics[width=\linewidth]{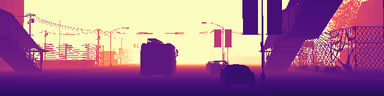}
	\end{subfigure}
\end{minipage}
\begin{minipage}{0.32\textwidth}
	\begin{subfigure}{.5\textwidth}
		\centering
		\includegraphics[width=\linewidth]{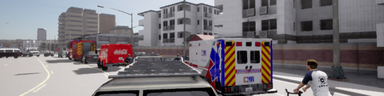}
	\end{subfigure}%
	\begin{subfigure}{.5\textwidth}
		\centering
		\includegraphics[width=\linewidth]{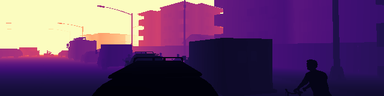}
	\end{subfigure}
	\\[1ex]
	\begin{subfigure}{.5\textwidth}
		\centering
		\includegraphics[width=\linewidth]{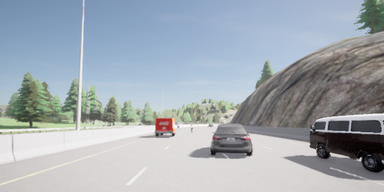}
	\end{subfigure}%
	\begin{subfigure}{.5\textwidth}
		\centering
		\includegraphics[width=\linewidth]{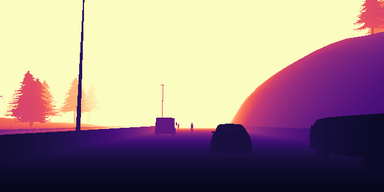}
	\end{subfigure}
	\\[1ex]
	\begin{subfigure}{.5\textwidth}
		\centering
		\includegraphics[width=\linewidth]{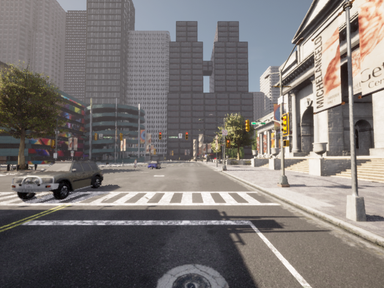}
	\end{subfigure}%
	\begin{subfigure}{.5\textwidth}
		\centering
		\includegraphics[width=\linewidth]{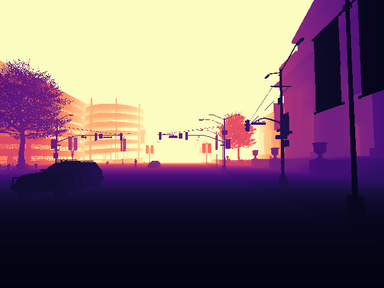}
	\end{subfigure}
\end{minipage}

%% file: figures/ddad/ddad_fig.tex
\newcommand{\turnheightnew}{0.195\columnwidth}
\newcommand{\len}{2mm}
\newcommand{\STAB}[1]{\begin{tabular}{@{}c@{}}#1\end{tabular}}

\centering

\begin{tabular}{c@{\hskip 2mm}c@{\hskip 2mm}c@{\hskip 2mm}c@{\hskip 2mm}c@{\hskip 2mm}c@{\hskip 2mm}c@{\hskip 2mm}c@{\hskip 2mm}c@{\hskip 2mm}}
	
	{\rotatebox{90}{\hspace{4mm}Input}}& 
	\includegraphics[height=\turnheightnew]{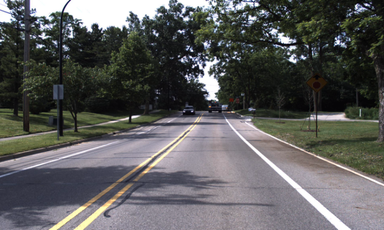}&
	\includegraphics[height=\turnheightnew]{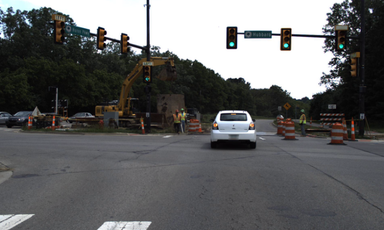}&
	\includegraphics[height=\turnheightnew]{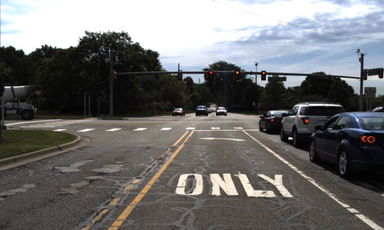}&
	\includegraphics[height=\turnheightnew]{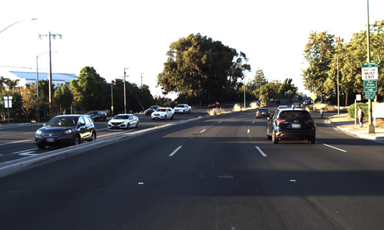}&
	\includegraphics[height=\turnheightnew]{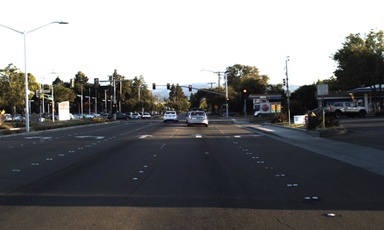}&
	\includegraphics[height=\turnheightnew]{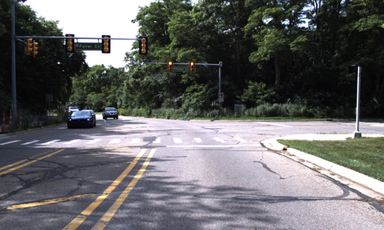}&
	\includegraphics[height=\turnheightnew]{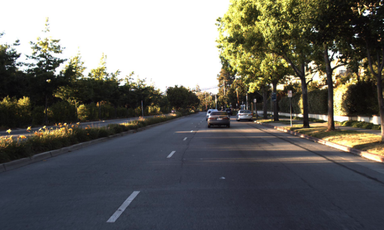}&

\\
\midrule

	\multirow{2}{*}{\rotatebox[origin=c]{90}{Monodepth2 \cite{godard2019digging} \hspace{-1cm}}}& 
\includegraphics[height=\turnheightnew]{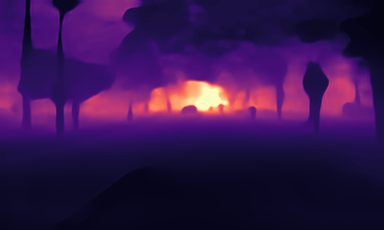}&
\includegraphics[height=\turnheightnew]{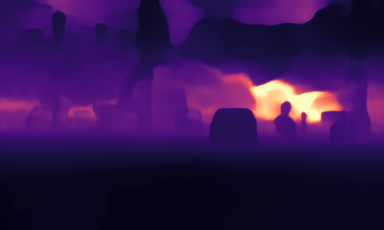}&
\includegraphics[height=\turnheightnew]{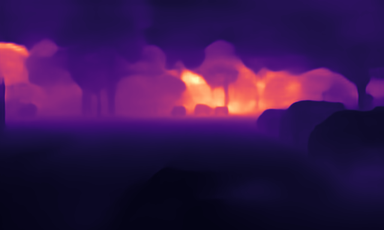}&
\includegraphics[height=\turnheightnew]{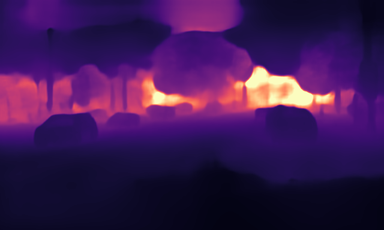}&
\includegraphics[height=\turnheightnew]{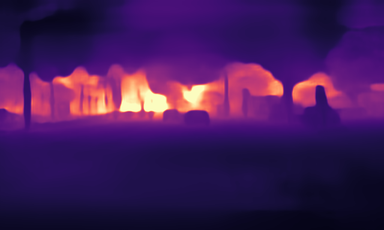}&
\includegraphics[height=\turnheightnew]{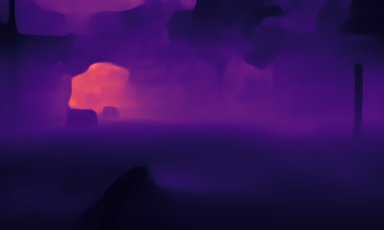}&
\includegraphics[height=\turnheightnew]{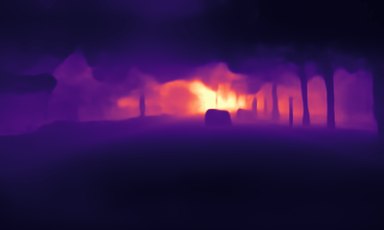}&

\\

&
\includegraphics[height=\turnheightnew]{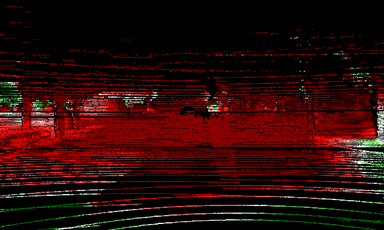}&
\includegraphics[height=\turnheightnew]{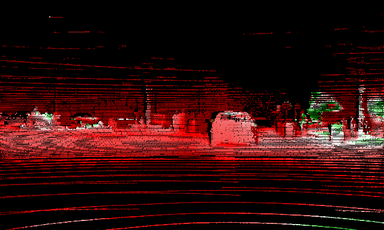}&
\includegraphics[height=\turnheightnew]{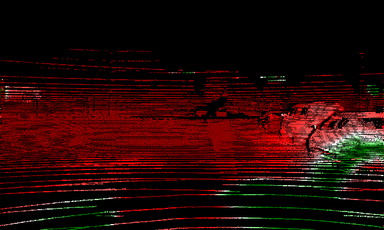}&
\includegraphics[height=\turnheightnew]{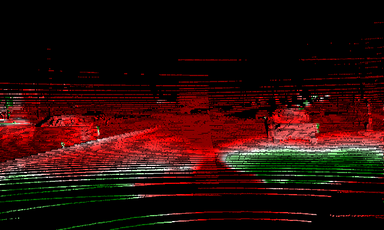}&
\includegraphics[height=\turnheightnew]{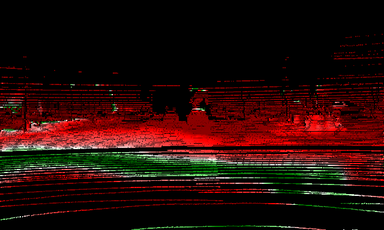}&
\includegraphics[height=\turnheightnew]{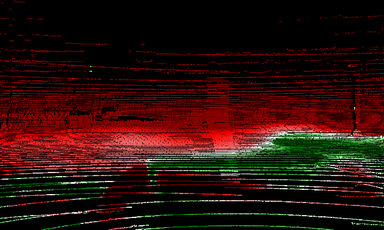}&
\includegraphics[height=\turnheightnew]{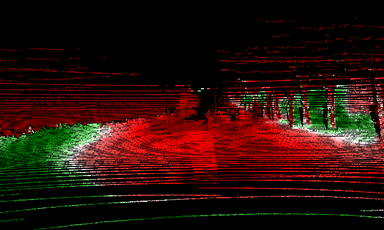}&

\\
\midrule

\multirow{2}{*}{\rotatebox[origin=c]{90}{iDisc\cite{piccinelli2023idisc} \hspace{-0.6cm}}}& 
\includegraphics[height=\turnheightnew]{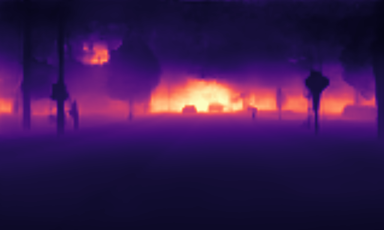}&
\includegraphics[height=\turnheightnew]{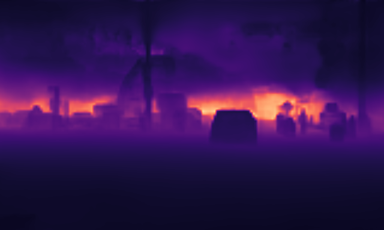}&
\includegraphics[height=\turnheightnew]{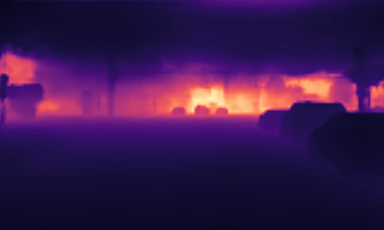}&
\includegraphics[height=\turnheightnew]{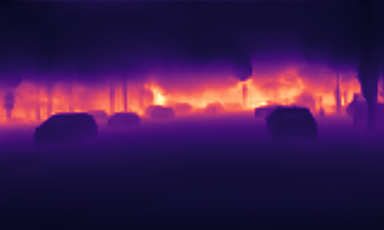}&
\includegraphics[height=\turnheightnew]{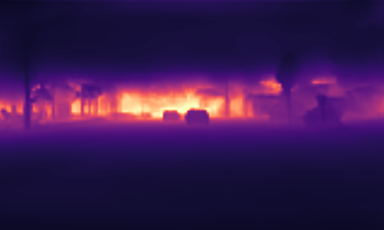}&
\includegraphics[height=\turnheightnew]{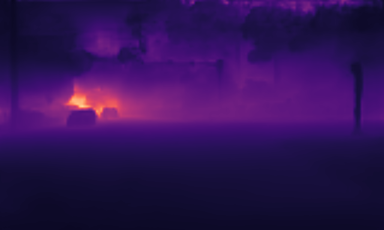}&
\includegraphics[height=\turnheightnew]{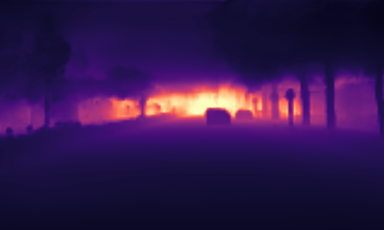}&

\\

& 
\includegraphics[height=\turnheightnew]{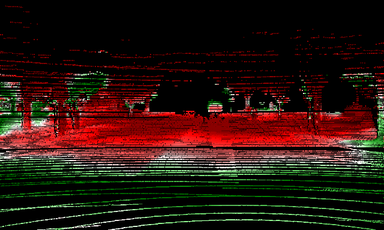}&
\includegraphics[height=\turnheightnew]{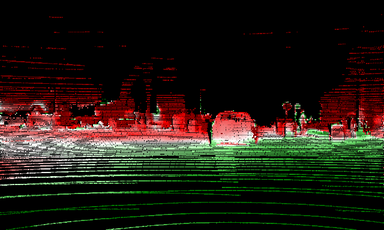}&
\includegraphics[height=\turnheightnew]{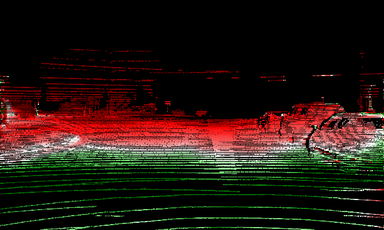}&
\includegraphics[height=\turnheightnew]{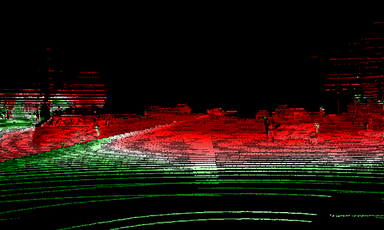}&
\includegraphics[height=\turnheightnew]{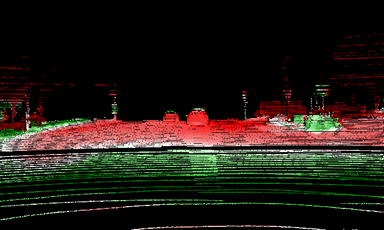}&
\includegraphics[height=\turnheightnew]{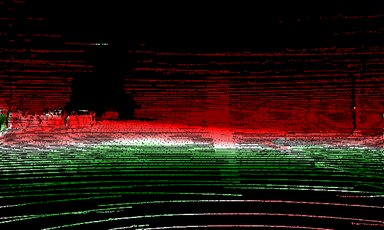}&
\includegraphics[height=\turnheightnew]{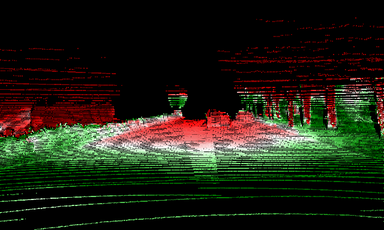}&

\\
\midrule

\multirow{2}{*}{\rotatebox[origin=c]{90}{PlaneDepth\cite{wang2023planedepth} \hspace{-1cm}}}& 
\includegraphics[height=\turnheightnew]{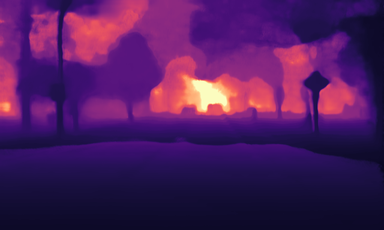}&
\includegraphics[height=\turnheightnew]{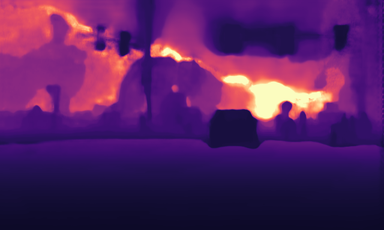}&
\includegraphics[height=\turnheightnew]{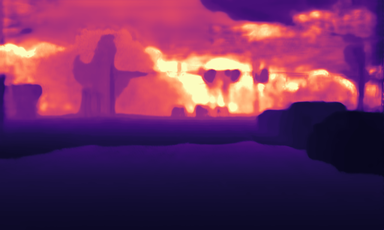}&
\includegraphics[height=\turnheightnew]{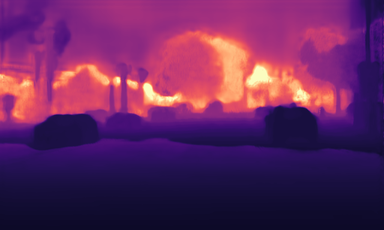}&
\includegraphics[height=\turnheightnew]{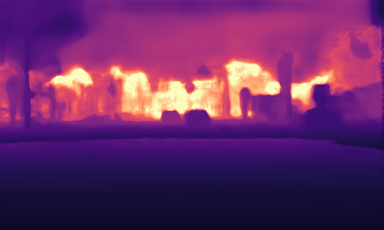}&
\includegraphics[height=\turnheightnew]{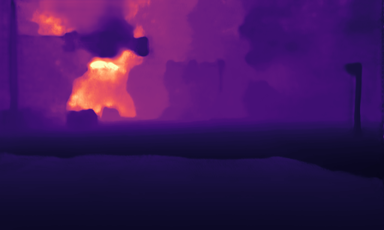}&
\includegraphics[height=\turnheightnew]{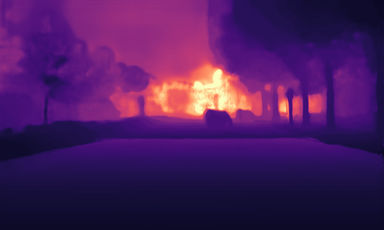}&

\\

& 
\includegraphics[height=\turnheightnew]{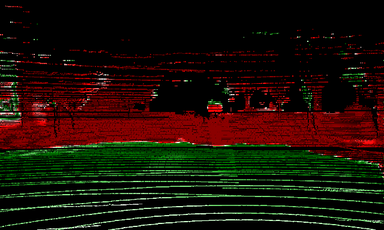}&
\includegraphics[height=\turnheightnew]{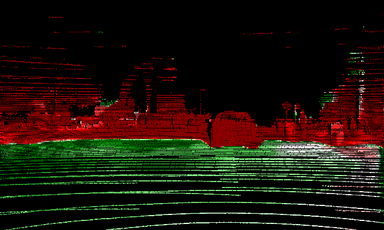}&
\includegraphics[height=\turnheightnew]{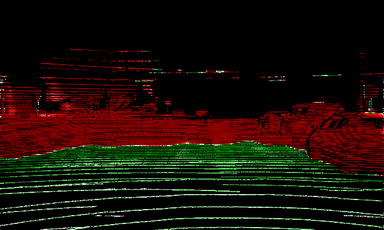}&
\includegraphics[height=\turnheightnew]{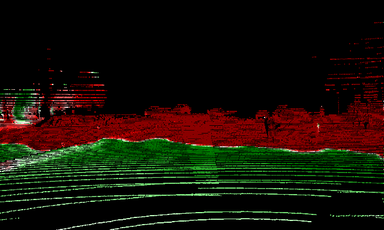}&
\includegraphics[height=\turnheightnew]{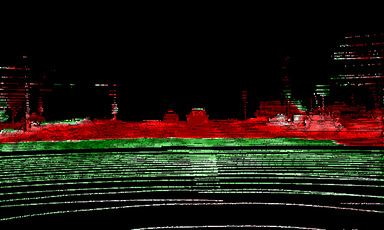}&
\includegraphics[height=\turnheightnew]{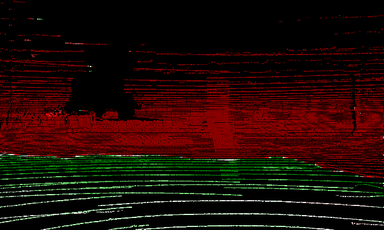}&
\includegraphics[height=\turnheightnew]{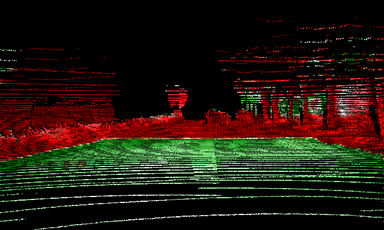}&

\\
\midrule

\multirow{2}{*}{\rotatebox[origin=c]{90}{LeReS\cite{yin2021learning} \hspace{-0.5cm}}}& 
\includegraphics[height=\turnheightnew]{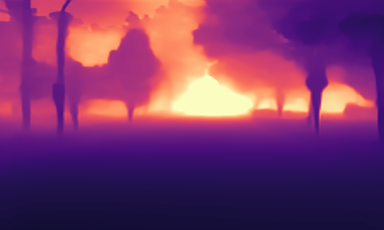}&
\includegraphics[height=\turnheightnew]{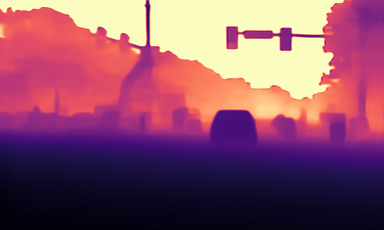}&
\includegraphics[height=\turnheightnew]{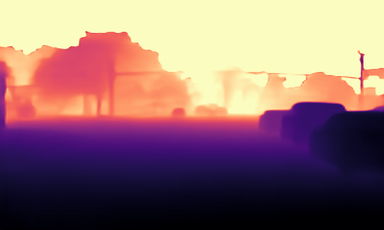}&
\includegraphics[height=\turnheightnew]{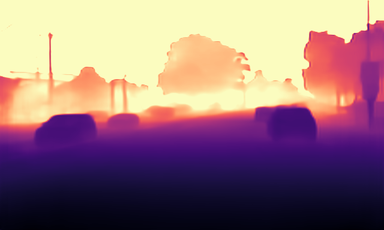}&
\includegraphics[height=\turnheightnew]{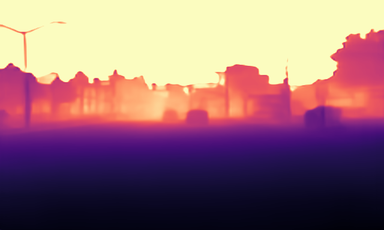}&
\includegraphics[height=\turnheightnew]{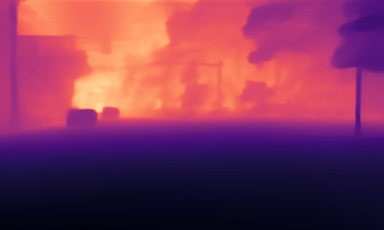}&
\includegraphics[height=\turnheightnew]{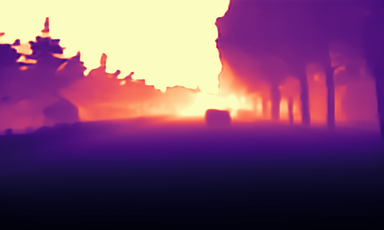}&

\\

& 
\includegraphics[height=\turnheightnew]{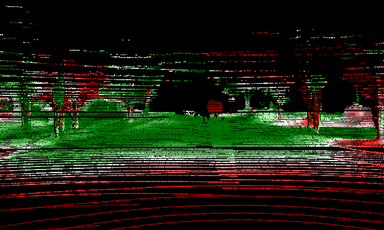}&
\includegraphics[height=\turnheightnew]{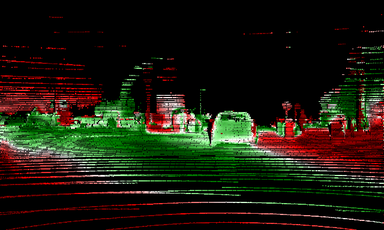}&
\includegraphics[height=\turnheightnew]{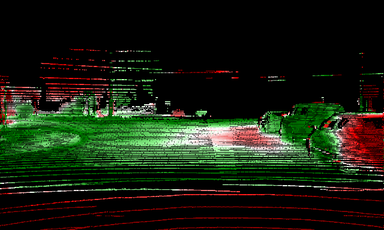}&
\includegraphics[height=\turnheightnew]{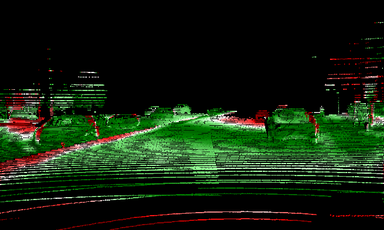}&
\includegraphics[height=\turnheightnew]{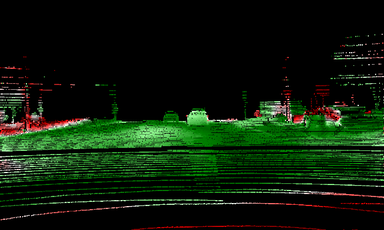}&
\includegraphics[height=\turnheightnew]{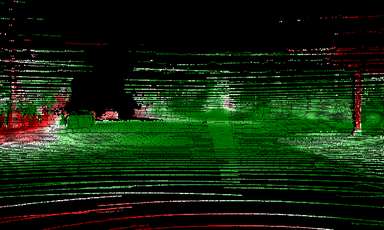}&
\includegraphics[height=\turnheightnew]{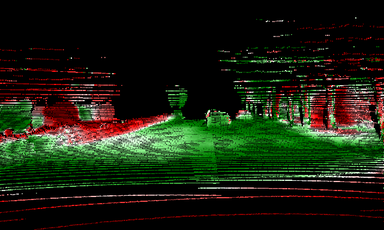}&

\\
\midrule

\multirow{2}{*}{\rotatebox[origin=c]{90}{PackNet-SfM\cite{guizilini20203d}\hspace{-1cm}}}& 
\includegraphics[height=\turnheightnew]{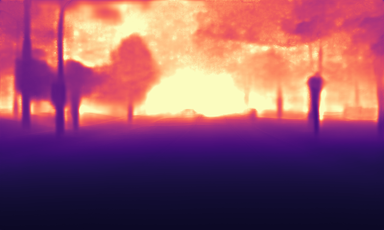}&
\includegraphics[height=\turnheightnew]{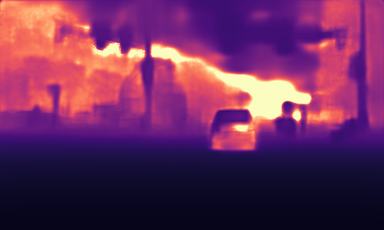}&
\includegraphics[height=\turnheightnew]{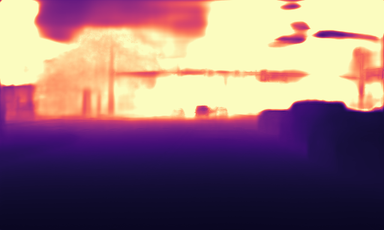}&
\includegraphics[height=\turnheightnew]{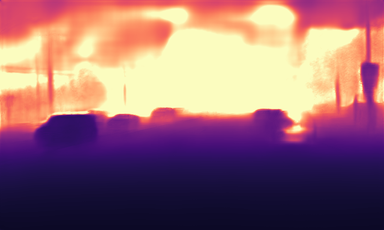}&
\includegraphics[height=\turnheightnew]{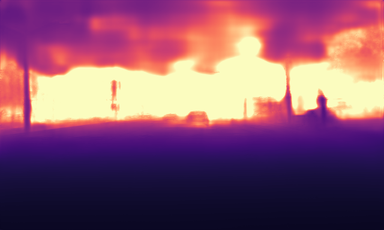}&
\includegraphics[height=\turnheightnew]{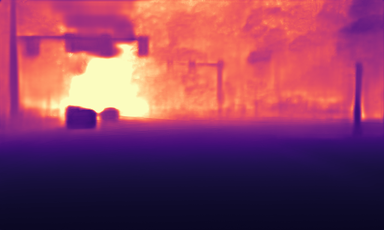}&
\includegraphics[height=\turnheightnew]{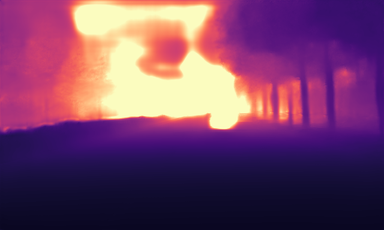}&

\\

& 
\includegraphics[height=\turnheightnew]{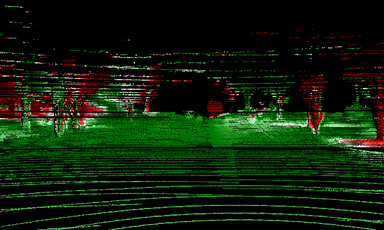}&
\includegraphics[height=\turnheightnew]{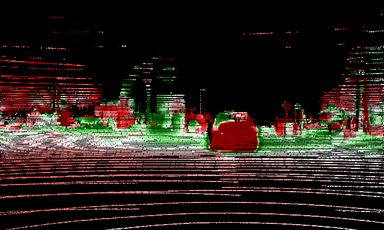}&
\includegraphics[height=\turnheightnew]{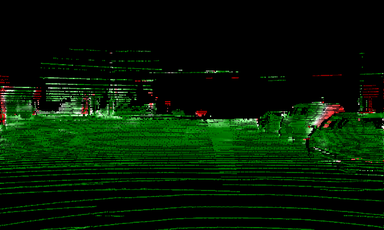}&
\includegraphics[height=\turnheightnew]{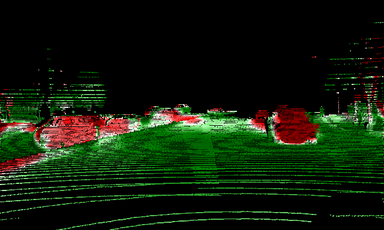}&
\includegraphics[height=\turnheightnew]{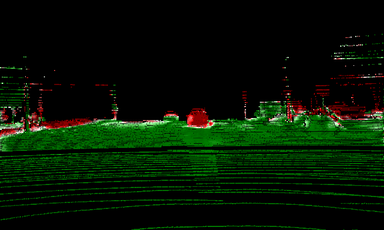}&
\includegraphics[height=\turnheightnew]{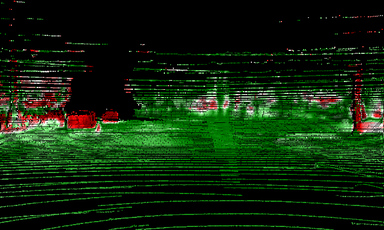}&
\includegraphics[height=\turnheightnew]{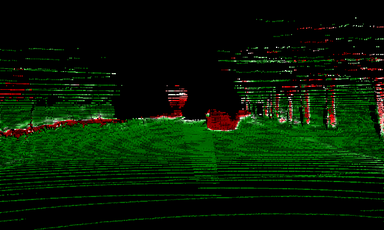}&
\\
\midrule

\multirow{2}{*}{\rotatebox[origin=c]{90}{GenDepth\hspace{-0.45cm}}}& 
\includegraphics[height=\turnheightnew]{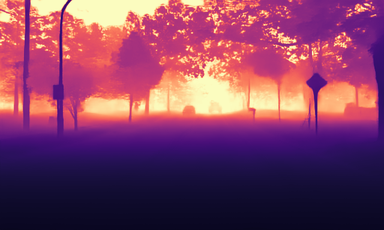}&
\includegraphics[height=\turnheightnew]{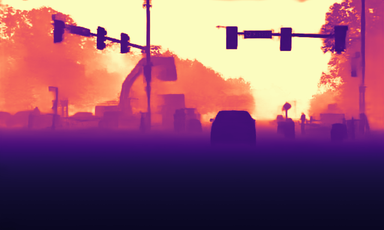}&
\includegraphics[height=\turnheightnew]{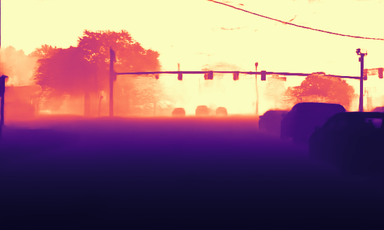}&
\includegraphics[height=\turnheightnew]{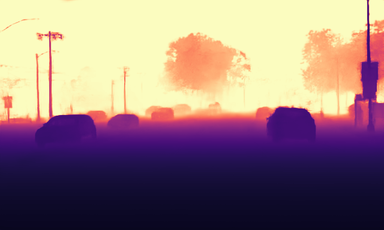}&
\includegraphics[height=\turnheightnew]{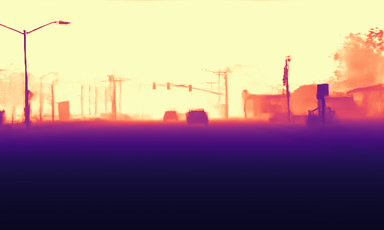}&
\includegraphics[height=\turnheightnew]{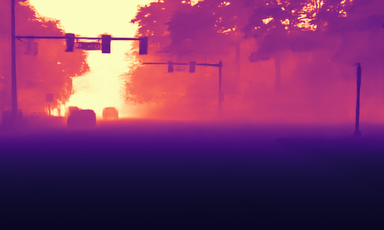}&
\includegraphics[height=\turnheightnew]{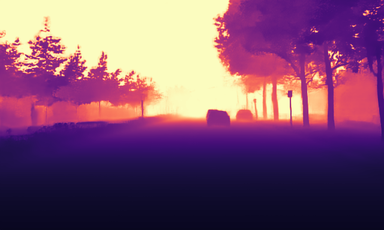}&
\\

& 
\includegraphics[height=\turnheightnew]{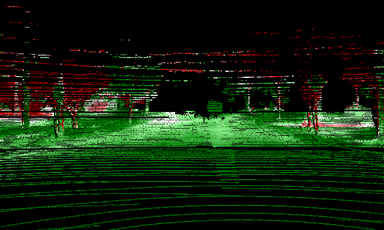}&
\includegraphics[height=\turnheightnew]{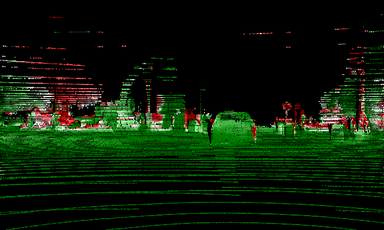}&
\includegraphics[height=\turnheightnew]{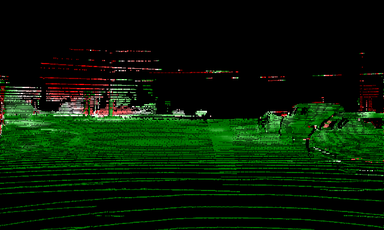}&
\includegraphics[height=\turnheightnew]{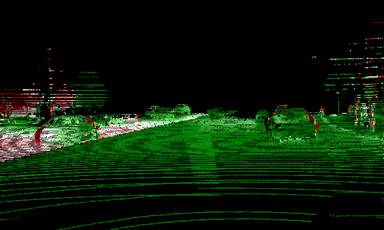}&
\includegraphics[height=\turnheightnew]{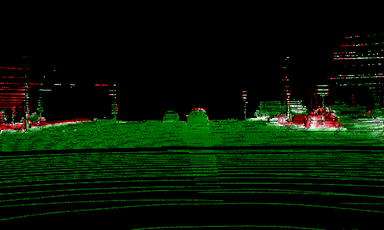}&
\includegraphics[height=\turnheightnew]{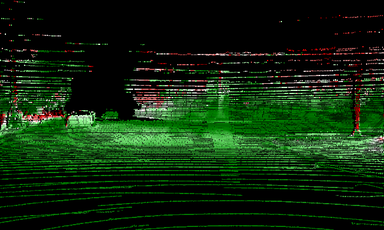}&
\includegraphics[height=\turnheightnew]{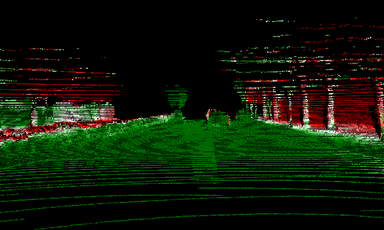}&
\\
\end{tabular}

	

%% file: figures/ddad_details/ddad_details.tex
\newcommand{\turnwidthnew}{0.235\linewidth}
\newcommand{\len}{2mm}

\centering

\begin{tabular}{@{\hskip 0mm}c@{\hskip 1mm}c@{\hskip 1mm}c@{\hskip 1mm}c@{\hskip 1mm}c}
	
	{\rotatebox{90}{\hspace{7mm}Input}}& 
	\includegraphics[width=\turnwidthnew]{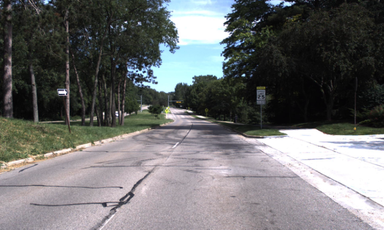}&
	\includegraphics[width=\turnwidthnew]{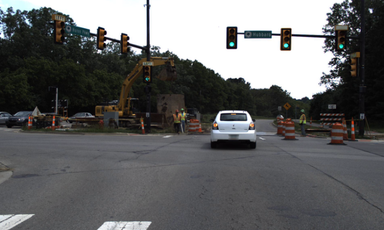}&
	\includegraphics[width=\turnwidthnew]{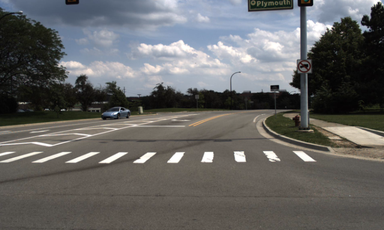}&
	\includegraphics[width=\turnwidthnew]{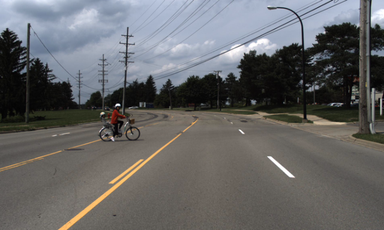}
	\\
	
	\rotatebox{90}{\hspace{2mm}PackNet-SfM}& 
	\includegraphics[width=\turnwidthnew]{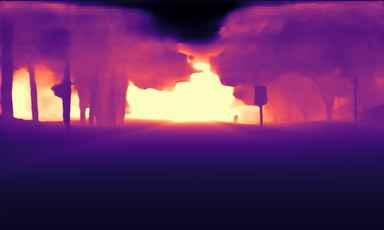}&
	\includegraphics[width=\turnwidthnew]{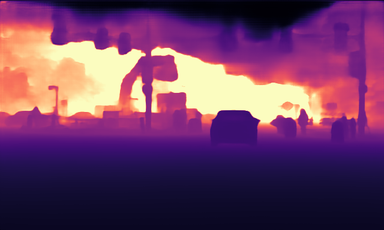}&
	\includegraphics[width=\turnwidthnew]{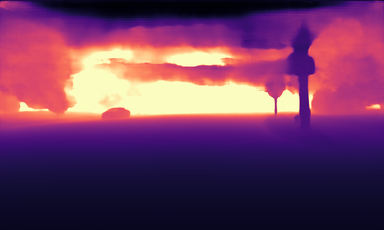}&
	\includegraphics[width=\turnwidthnew]{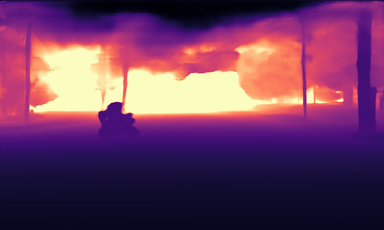} \\
	
	\rotatebox{90}{\hspace{4mm}GenDepth}& 
	\includegraphics[width=\turnwidthnew]{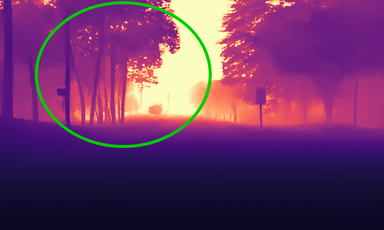}&
	\includegraphics[width=\turnwidthnew]{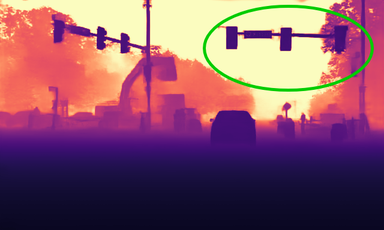}&
	\includegraphics[width=\turnwidthnew]{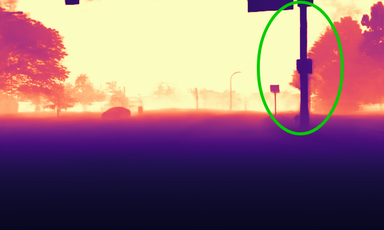}&
	\includegraphics[width=\turnwidthnew]{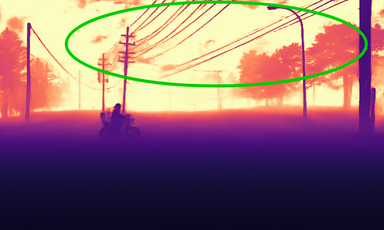}
	
	\\

\end{tabular}

%% file: figures/ablations/ablations_fig.tex
\newcommand{\turnwidthnew}{0.3\columnwidth}
\newcommand{\len}{2mm}
\newcommand{\STAB}[1]{\begin{tabular}{@{}c@{}}#1\end{tabular}}

\centering

\begin{tabular}{@{\hskip 0mm}r@{\hskip 1mm}c@{\hskip 1mm}c@{\hskip 1mm}c@{\hskip 1mm}c@{\hskip 1mm}}
	
	\raisebox{1\normalbaselineskip}[0pt][0pt]{\rotatebox{90}{Input}}& 
	\includegraphics[width=\turnwidthnew]{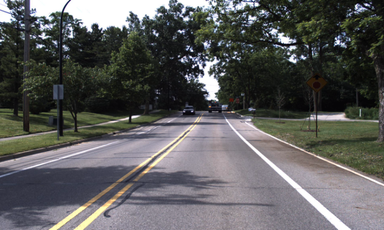}&
	\includegraphics[width=\turnwidthnew]{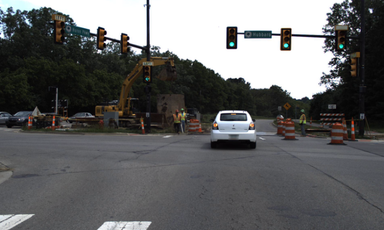}&
	\includegraphics[width=\turnwidthnew]{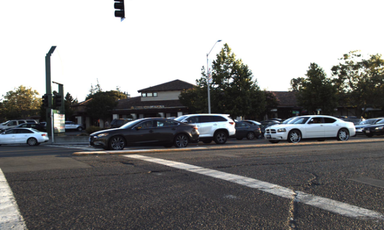}&

\\
	\raisebox{1.5\normalbaselineskip}[0pt][0pt]{\rotatebox{90}{C}}& 
\includegraphics[width=\turnwidthnew]{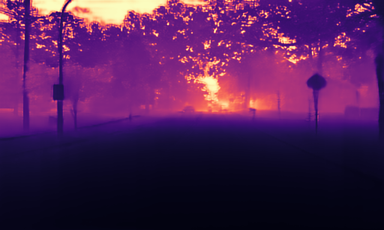}&
\includegraphics[width=\turnwidthnew]{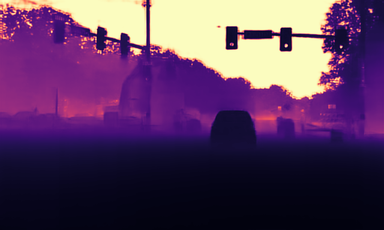}&
\includegraphics[width=\turnwidthnew]{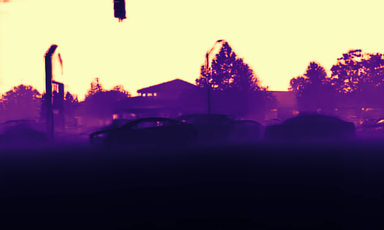}&

\\

	\raisebox{1.5\normalbaselineskip}[0pt][0pt]{\rotatebox{90}{K}}& 
\includegraphics[width=\turnwidthnew]{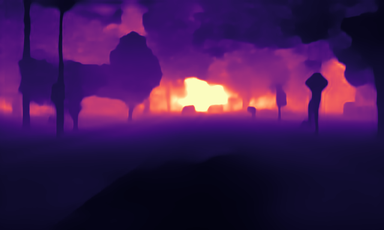}&
\includegraphics[width=\turnwidthnew]{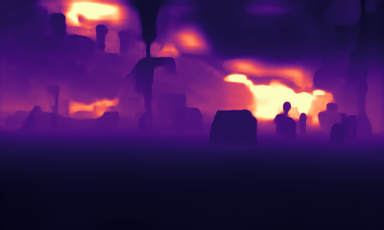}&
\includegraphics[width=\turnwidthnew]{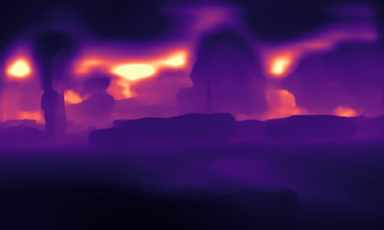}&

\\

	\raisebox{1.4\normalbaselineskip}[0pt][0pt]{\rotatebox{90}{CK}}& 
\includegraphics[width=\turnwidthnew]{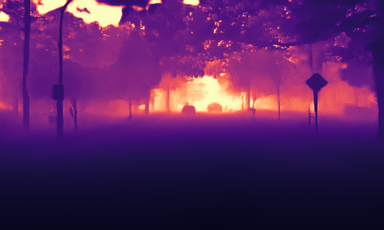}&
\includegraphics[width=\turnwidthnew]{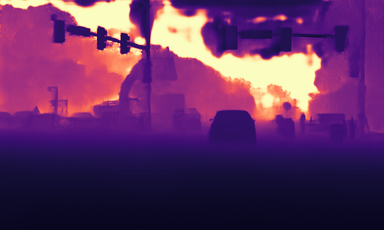}&
\includegraphics[width=\turnwidthnew]{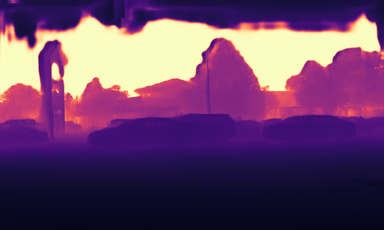}&

\\

	\raisebox{1.4\normalbaselineskip}[0pt][0pt]{\rotatebox{90}{CG}}& 
\includegraphics[width=\turnwidthnew]{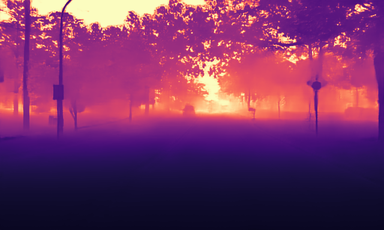}&
\includegraphics[width=\turnwidthnew]{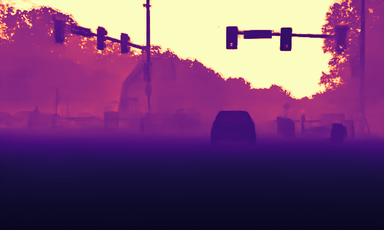}&
\includegraphics[width=\turnwidthnew]{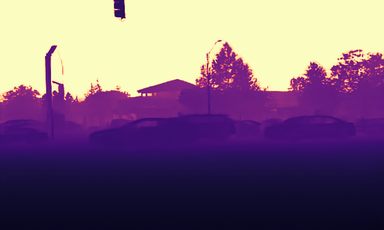}&

\\

	\raisebox{1.1\normalbaselineskip}[0pt][0pt]{\rotatebox{90}{CKG}}& 
\includegraphics[width=\turnwidthnew]{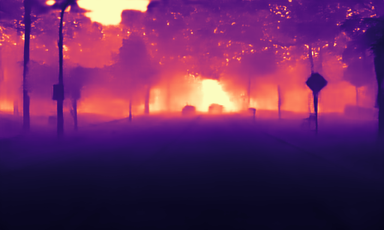}&
\includegraphics[width=\turnwidthnew]{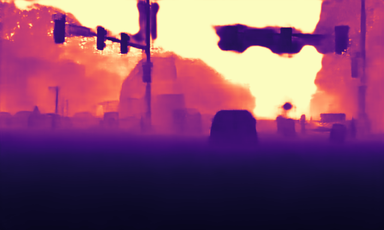}&
\includegraphics[width=\turnwidthnew]{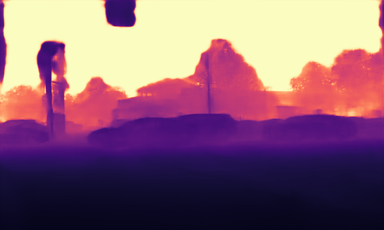}&

\\

	\raisebox{0.8\normalbaselineskip}[0pt][0pt]{\rotatebox{90}{CKGD}}& 
\includegraphics[width=\turnwidthnew]{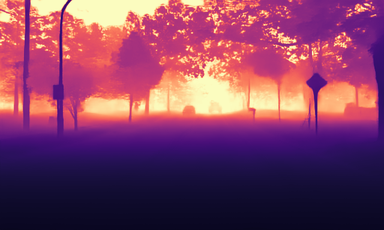}&
\includegraphics[width=\turnwidthnew]{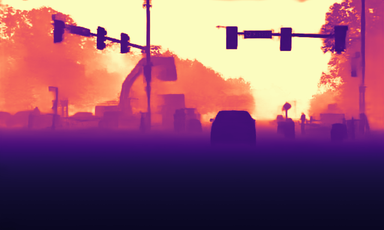}&
\includegraphics[width=\turnwidthnew]{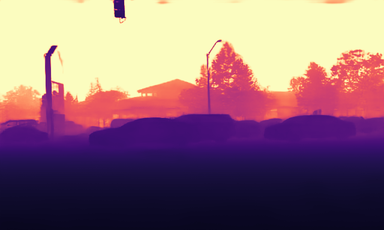}&

\\

\end{tabular}

	

%% file: figures/failures/failures_fig.tex
\newcommand{\turnwidthnew}{0.17 \linewidth}
\newcommand{\len}{2mm}
\newcommand{\STAB}[1]{\begin{tabular}{@{}c@{}}#1\end{tabular}}

\centering

\begin{tabular}{@{\hskip 0mm}c@{\hskip 1mm}c@{\hskip 1mm}c@{\hskip 1mm}c@{\hskip 1mm}c@{\hskip 1mm}c@{\hskip 0mm}}

	\includegraphics[width=\turnwidthnew]{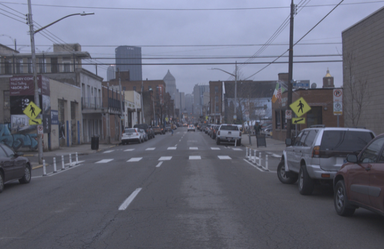}&
	\includegraphics[width=\turnwidthnew]{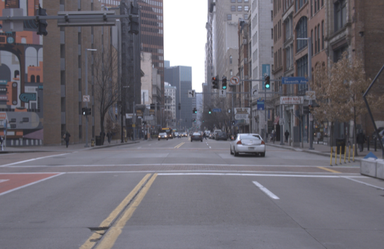}&
	\includegraphics[width=\turnwidthnew]{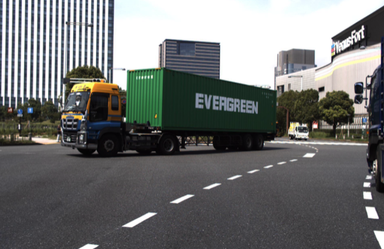}&
	\includegraphics[width=\turnwidthnew]{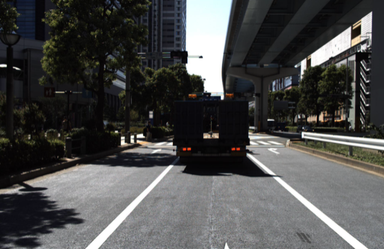}
	\\
	\includegraphics[width=\turnwidthnew]{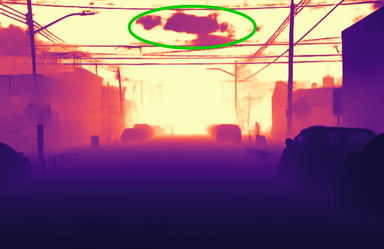}&
	\includegraphics[width=\turnwidthnew]{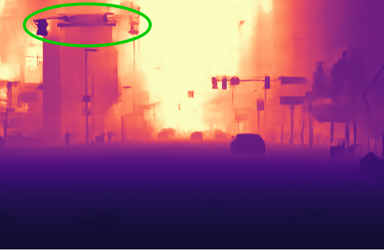}&
	\includegraphics[width=\turnwidthnew]{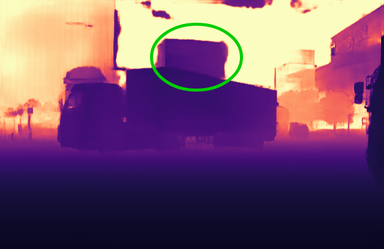}&
	\includegraphics[width=\turnwidthnew]{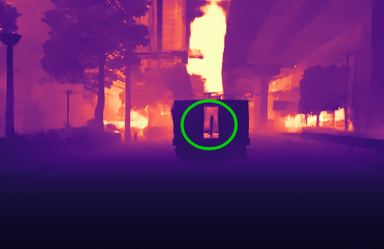}
\end{tabular}

	

%% file: sections/conclusion.tex
In this paper, we present GenDepth, a novel MDE framework developed to overcome perspective geometry bias and provide accurate depth estimation for different vehicle-camera systems. 
After a thorough analysis and discussion of the effects of parameter variations on the MDE and potential problems, we proposed to use highly diverse synthetic data during training, thereby framing the problem as a sim2real adaptation.
To that end, we have collected a bespoke synthetic dataset featuring many different vehicle-camera setups.
To inform the network about the known camera parameters, we propose to embed intrinsics and extrinsics together as the depth of the ground plane. 
This embedding is then carefully integrated with the adversarial domain alignment, eliminating potential generalization limitations due to large domain gaps.
Finally, we have thoroughly validated GenDepth on several real-world datasets, showcasing state-of-the-art generalization capability that is not present in existing approaches.
In addition to the specific beneficial effects of our contributions, our detailed ablation experiments showcase the potential of using synthetically generated ground-truth data, even in relatively simple configurations.
These findings suggest that future research opportunities in MDE lie in the application of domain generalization techniques to a combination of real-world and diverse synthetically generated data, which can mitigate perspective geometry biases and enable the estimation of finer and more accurate depth maps.

%% file: sections/appendix.tex
\subsection*{Evaluation metrics equations}
Given the estimated depths $d_i$ and ground truth depths $\hat{d}_i$ at pixel $i$, the metrics used in this work can be calculated as:
\begin{align}
	\mathrm{Abs Rel:}  & \quad\frac{1}{\lvert\mathbf{\hat{D}}\rvert} \sum_i \frac{\lvert d_i - \hat{d}_i \rvert}{\hat{d}_i}            \\
	\mathrm{Sq Rel:}   & \quad \frac{1}{\lvert\mathbf{\hat{D}}\rvert} \sum_i \frac{(d_i - \hat{d}_i)^2}{\hat{d}_i}                     \\
	\mathrm{RMSE:}     & \quad \sqrt{\frac{1}{\lvert\mathbf{\hat{D}}\rvert} \sum_i (d_i - \hat{d}_i)^2 }                               \\
	\mathrm{RMSE log:} & \quad \sqrt{\frac{1}{\lvert\mathbf{\hat{D}}\rvert} \sum_i (\log_{10}d_i - \log_{10}\hat{d}_i)^2 }              \\
	\delta < \tau:     & \quad\text{\% of $d_i$ s.t. $\mathrm{max}$} \left(\frac{d_i}{\hat{d}_i}, \frac{\hat{d}_i}{d_i}\right) < \tau,
\end{align} where $\lvert\mathbf{\hat{D}}\rvert$ represents the cardinality of the possibly sparse ground truth depth map after the selection of viable pixels. 

\subsection*{Additional quantitative results}
\begin{table}[!h]
	\renewcommand{\arraystretch}{1.2}
	\centering
	\caption{Network architecture details and ablation of computational complexity. All execution times are measured during inference with a Nvidia RTX A5000 GPU on the DDAD dataset. All methods use 640x384 resolution, except MiDaS and LeReS, which use 384x384 and 448x448, respectively.}
	\resizebox{\columnwidth}{!}{
	\begin{tabular}{ccccc}
		\toprule[0.6mm]
		Method & Encoder & Execution time & Params \#
		\\
		\midrule
		Monodepth2 \cite{godard2019digging} & ResNet-18 \cite{he2016deep} & 5.70 ms & 14.84 M \\
		DIFFNet \cite{zhou_diffnet} & HRNet-18 \cite{wang2020deep} & 33.02 ms & 10.87 M \\
		PackNet-SfM \cite{guizilini20203d} & Custom & 55.87 ms & 129.89 M \\
		MonoViT \cite{zhao2022monovit} & Custom  & 37.42 ms & 27.87 M \\
		iDisc \cite{piccinelli2023idisc} & ResNet-101 \cite{he2016deep} & 65.48 ms & 58.94 M \\
		NeWCRFs \cite{yuan2022newcrfs} & Swin-L \cite{liu2021swin} & 59.56 ms & 270.44 M \\
		EPCDepth \cite{peng2021excavating} & ResNet-50 \cite{he2016deep} & 18.28 ms & 26.86 M \\
		PlaneDepth \cite{wang2023planedepth} & ResNet-50 \cite{he2016deep} & 16.72 ms & 39.14 M \\
		MiDaS \cite{ranftl2020towards} & ResNeXt-101 \cite{xie2017aggregated} & 24.56ms & 105.36 M \\
		LeReS \cite{yin2021learning} & ResNet-50 \cite{he2016deep} & 18.90 ms & 52.13 M \\
		\textbf{GenDepth} & ResNet-50 \cite{he2016deep} & 22.67 ms & 35.95 M \\
		\bottomrule[0.6mm]
	\end{tabular}}
	\label{tbl:complexity}
\end{table}

\begin{table*}[!h]
	\renewcommand{\arraystretch}{1.2}
	\centering
	\caption{Depth estimation results on the DDAD dataset \cite{guizilini20203d} after training on the KITTI dataset \cite{geiger2013vision}. Compared to Table \ref{tbl:ddad_results}, we include results with median scaling alignment w.r.t. the ground-truth depth.}
	\label{tbl:med_scale}
	\begin{tabular}{cccccccccccc}
		\toprule[0.6mm]
		
		\multirow{2}{*}[-0.4em]{Method}& \multirow{2}{*}[-0.4em]{Supervision} & \multirow{2}{*}[-0.4em]{Median scaling} &  \multicolumn{4}{c}{Error $\downarrow$} &  \multicolumn{3}{c}{Accuracy $\uparrow$}\\
		\cmidrule(lr){4-7} \cmidrule(lr){8-10} 	
		& & & Abs Rel & Sq Rel & RMSE  & RMSE log & $\delta < 1.25 $ & $\delta < 1.25^{2}$ & $\delta < 1.25^{3}$\\
		\midrule[0.6mm]
		\multirow{2}{*}{Monodepth2\cite{godard2019digging}} & \multirow{2}{*}{M + S} & \xmark & 0.378 & 5.268 & 14.606 & 0.599 & 0.201 & 0.457 & 0.740 \\
		& & \cmark & 0.235 & 2.482 & 9.638 & 0.335 & 0.581 & 0.838 & 0.940 \\
		\midrule
		\multirow{2}{*}{DIFFNet \cite{zhou_diffnet}} & \multirow{2}{*}{M + S} & \xmark & 0.373 & 5.128 & 14.406 & 0.563 & 0.161 & 0.426 & 0.791 \\
		& & \cmark &  0.199 & 1.947 & 8.413 & 0.275 & 0.683 & 0.896 & 0.967 \\
		\midrule
		\multirow{2}{*}{PackNet-SfM \cite{guizilini20203d}} & \multirow{2}{*}{M + v} & \xmark &  0.369 & 5.500 & 15.127 & 0.576 & 0.195 & 0.452 & 0.764 \\
		& &  \cmark & 0.215 & 2.121 & 8.951 & 0.297 & 0.646 & 0.880 & 0.962 \\
		\midrule
		\multirow{2}{*}{Monovit \cite{zhao2022monovit}} & 	\multirow{2}{*}{M + S} & \xmark & 0.218 & 2.912 & 11.469 & 0.357 & 0.556 & 0.831 & 0.928\\
		& & \cmark &  0.197 & 1.932 & 8.912 & 0.276 & 0.661 & 0.916 & 0.968 \\
		
		\midrule
		\multirow{2}{*}{PackNet-SfM \cite{guizilini20203d}} & \multirow{2}{*}{D} & \xmark &  0.322 & 4.834 & 14.478 & 0.536 & 0.305 & 0.617 & 0.821\\
		& & \cmark &  0.250 & 2.672 & 10.054 & 0.350 & 0.574 & 0.837 & 0.937 \\
		\midrule
		\multirow{2}{*}{iDisc \cite{piccinelli2023idisc}} & \multirow{2}{*}{D} & \xmark &  0.274 & 3.993 & 13.304 & 0.461 & 0.384 & 0.743 & 0.884\\
		& & \cmark &  0.215 & 2.244 & 9.407 & 0.312 & 0.656 & 0.886 & 0.949 \\
		\midrule
		\multirow{2}{*}{NeWCRFs \cite{yuan2022newcrfs}} & \multirow{2}{*}{D} & \xmark & 0.337 & 5.450 & 15.572 & 0.571 & 0.302 & 0.629 & 0.792\\
		& & \cmark &  0.356 & 4.091 & 11.911 & 0.449 & 0.436 & 0.704 & 0.861 \\
		
		\midrule

		\multirow{2}{*}{EPCDepth \cite{peng2021excavating}} & \multirow{2}{*}{M + S} & \xmark & 0.301 & 3.956 & 12.820 & 0.469 & 0.293 & 0.686 & 0.889\\
		& & \cmark &  0.226 & 2.310 & 8.924 & 0.311 & 0.648 & 0.866 & 0.952 \\
		\midrule
		\multirow{2}{*}{PlaneDepth \cite{wang2023planedepth}} & \multirow{2}{*}{M + S} & \xmark & 0.327 & 4.760 & 13.955 & 0.529 & 0.379 & 0.467 & 0.792 \\
		& & \cmark &  0.337 & 3.175 & 10.015 & 0.386 & 0.312 & 0.772 & 0.949 \\

		\midrule
		\multirow{2}{*}{\textbf{GenDepth}} &\multirow{2}{*}{S} & \xmark & 0.121 & 1.421 & 6.992 & 0.200 & 0.840 & 0.954 & 0.983\\
		
		& & \cmark & \textbf{0.114 }& \textbf{1.370} & \textbf{6.878} & \textbf{0.185} & \textbf{0.865} & \textbf{0.958} & \textbf{0.984} \\
		\bottomrule[0.6mm]
	\end{tabular}
\end{table*}

In Table \ref{tbl:complexity} we specify the architecture details and corresponding execution times for methods used throughout this work. 
Execution time is averaged for inference on the whole DDAD validation split. 
GenDepth exhibits competitive real-time performance, despite solving a relatively complex generalization problem compared to other methods.
Even though we use a simple ResNet-50 network for feature extraction, GenDepth estimates finer depth maps of higher structural quality when compared to recent methods such as iDisc \cite{piccinelli2023idisc}, which usually use complex architectures with higher execution times.
This challenges the current trend of small incremental improvements in MDE performance via architecture engineering and indicates the value of high-quality diverse ground-truth data without perspective geometry biases.

Furthermore, in Table \ref{tbl:med_scale} we examine the effects of median scaling during inference. 
Even though median scaling generally improves the quantitative performance, none of the methods achieve competitive results compared to GenDepth. 
This means that the domain gap induced by varying camera parameters leads to highly inconsistent depth map estimation, which can not be scale-aligned via fusion with another sensor.
On the contrary, GenDepth achieves accurate metric estimation both with and without scale alignment. Our results generated with median scaling are slightly better, possibly due to the small calibration errors on the DDAD dataset.
\subsection*{Qualitative results on additional datasets}
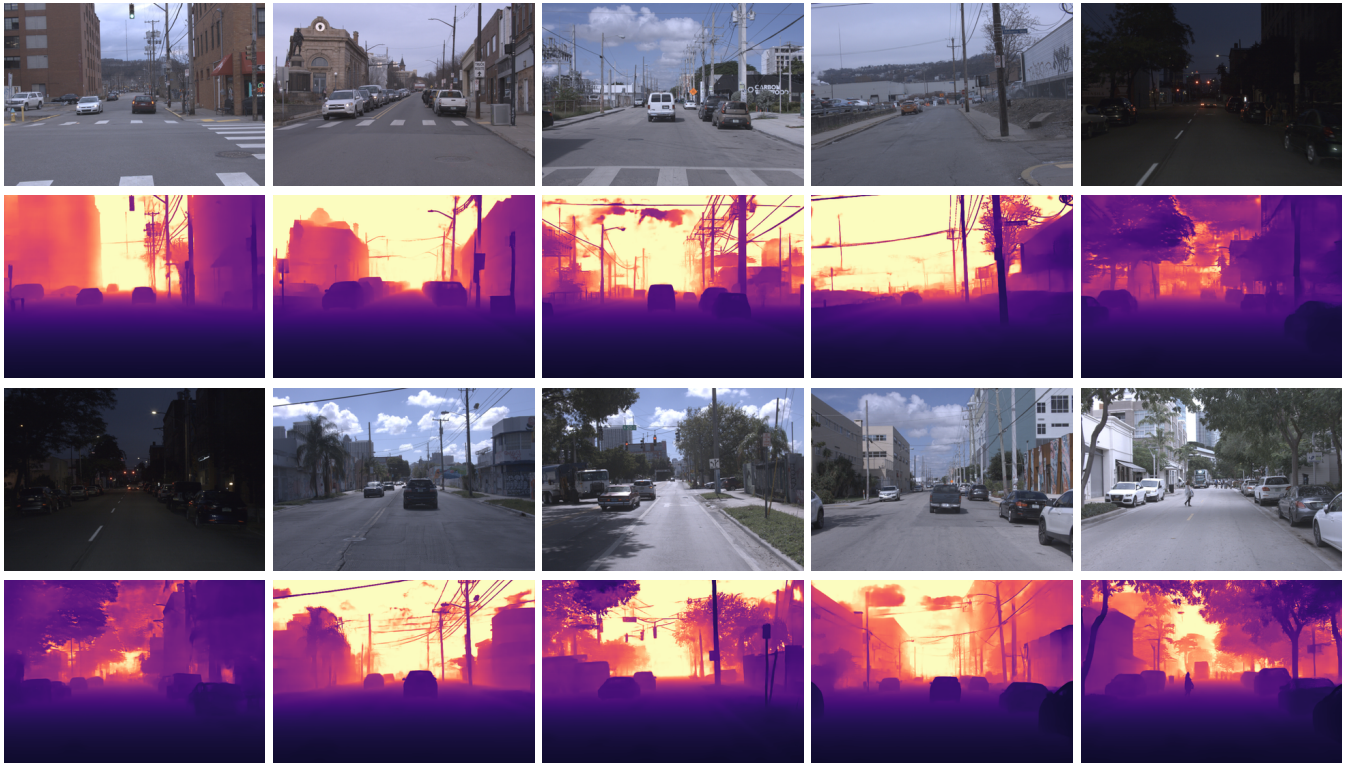
\begin{figure*}[!t]
	\centering
	\resizebox{\linewidth}{!}{\input{figures/appendix_argoverse/appendix_argoverse.tex}}
	\caption{Qualitative results on the Argoverse dataset. We also show results for nigh-time images, exemplifying GenDepth generalization ability for extreme domain gaps which are usually problematic for other methods \cite{gasperini2023robust}.}.
	\label{fig:appendix_argoverse}
\end{figure*}

We show additional qualitative results on the KITTI-360, nuScenes, Waymo, Argoverse and DDAD datasets. 
All depth maps are generated with a single model trained with a data configuration as in Fig. \ref{fig:config}. 
Our depth maps are highly precise, exhibiting extremely fine details when compared to current state-of-the-art approaches, even if they are trained and tested on the same dataset.
Overall, we demonstrate impressive generalization capability on all datasets, despite never using them during training. 
In Fig. \ref{fig:appendix_argoverse} we also include two images captured during night-time, which usually presents a significant issue for MDE methods. 
Even in such scenario, GenDepth accurately estimates key objects, indicating that it learns meaningful geometric priors which are easily transferable across environmental domains.

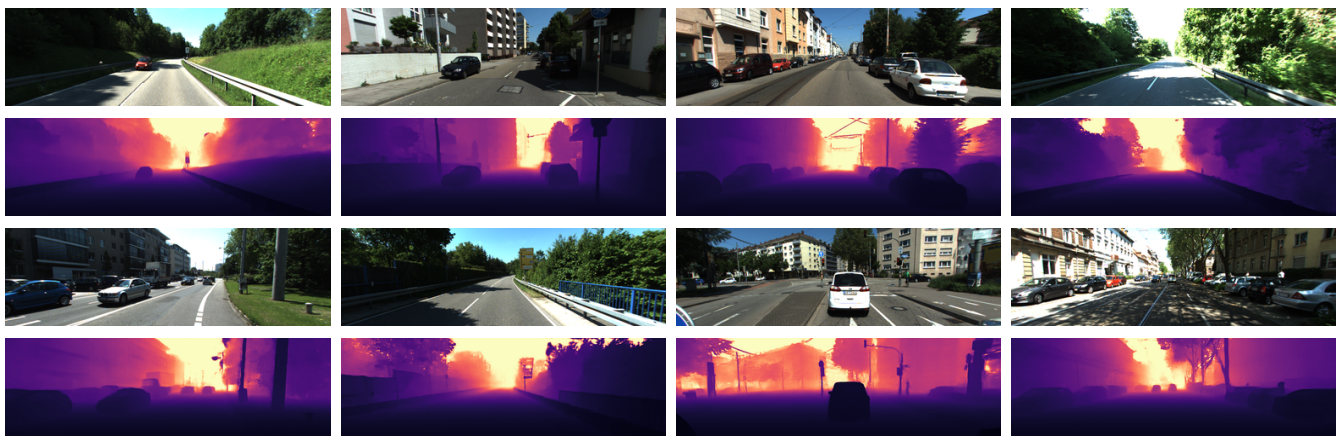
\begin{figure*}[!h]
	\centering
	\resizebox{\linewidth}{!}{\input{figures/appendix_kitti360/appendix_kitti360.tex}}
	\caption{Qualitative results on the KITTI-360 dataset.}
	\label{fig:appendix_kitti360}
\end{figure*}

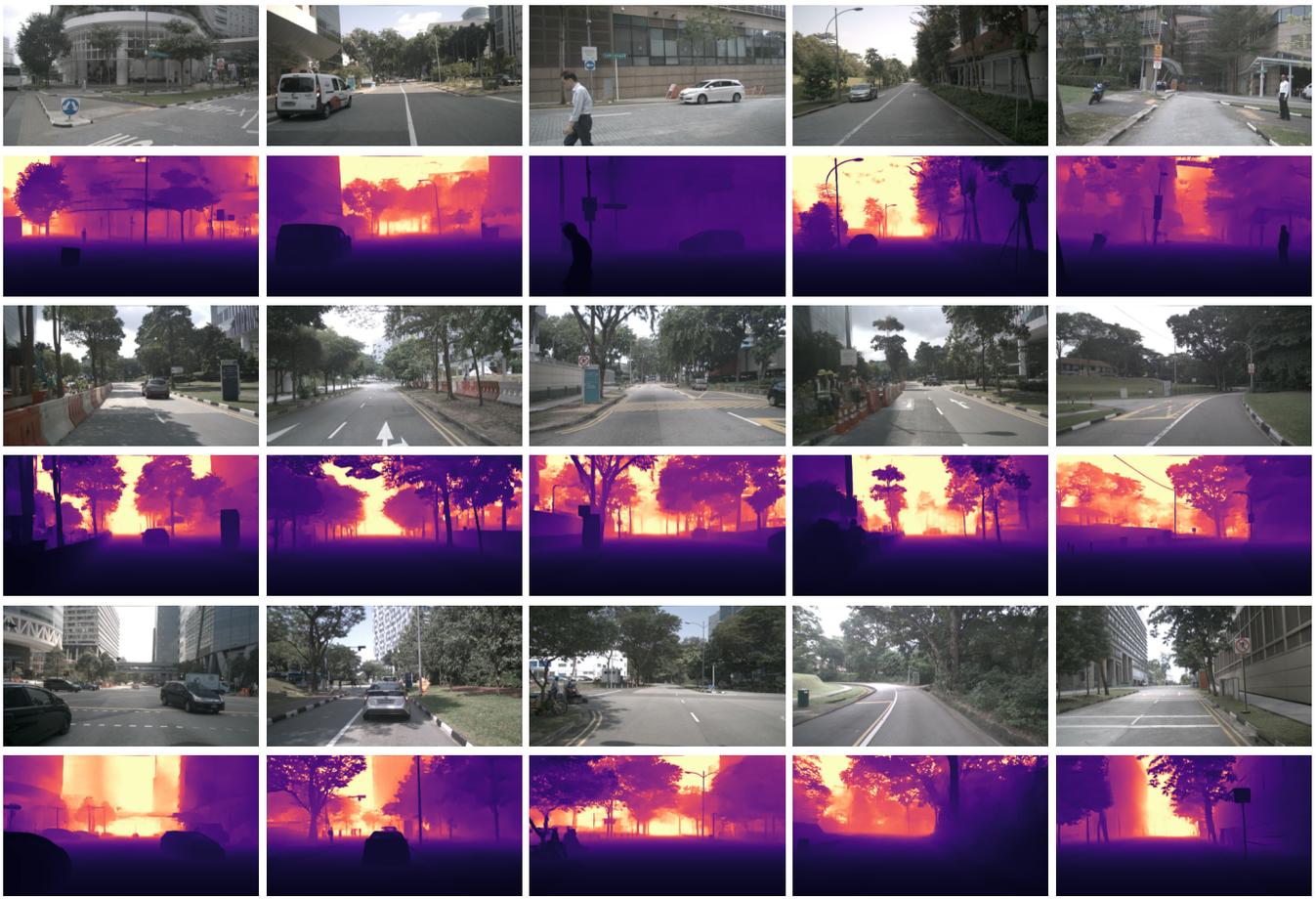
\begin{figure*}[!t]
	\centering
	\resizebox{\linewidth}{!}{\input{figures/appendix_nuscenes/appendix_nuscenes.tex}}
	\caption{Qualitative results on the nuScenes dataset.}
	\label{fig:appendix_nuscenes}
\end{figure*}

\begin{figure*}[!t]
	\centering
	\resizebox{\linewidth}{!}{\input{figures/appendix_waymo/appendix_waymo.tex}}
	\caption{Qualitative results on the Waymo dataset.}
	\label{fig:appendix_waymo}
\end{figure*}
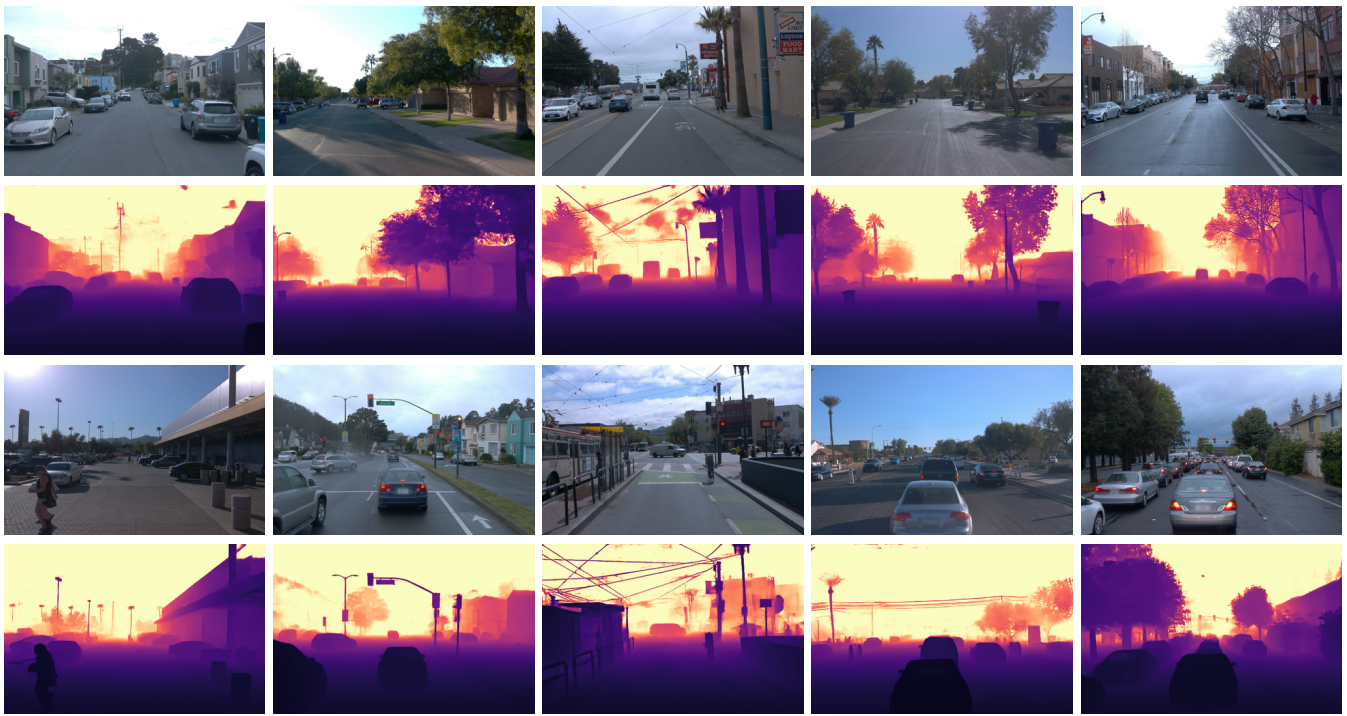

\begin{figure*}[!t]
	\centering
	\resizebox{\linewidth}{!}{\input{figures/appendix_ddad/appendix_ddad.tex}}
	\caption{Additional qualitative results on the DDAD dataset.}
	\label{fig:appendix_ddad}
\end{figure*}
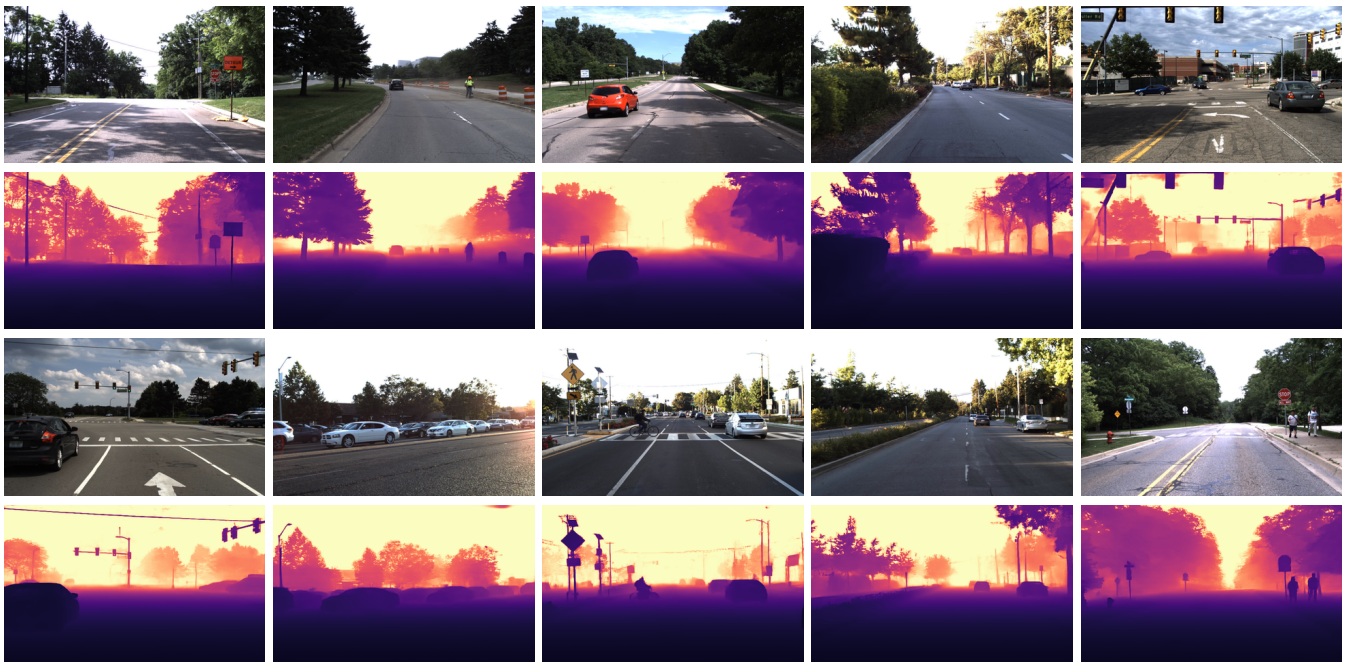

%% file: figures/appendix_kitti360/appendix_kitti360.tex
\newcommand{\turnwidthnew}{0.17 \linewidth}
\newcommand{\len}{2mm}
\newcommand{\STAB}[1]{\begin{tabular}{@{}c@{}}#1\end{tabular}}

\centering

\begin{tabular}{@{\hskip 0mm}c@{\hskip 1mm}c@{\hskip 1mm}c@{\hskip 1mm}c@{\hskip 0mm}}

	\includegraphics[width=\turnwidthnew]{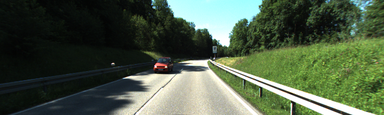}&
	\includegraphics[width=\turnwidthnew]{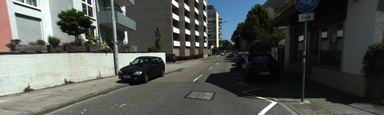}&
	\includegraphics[width=\turnwidthnew]{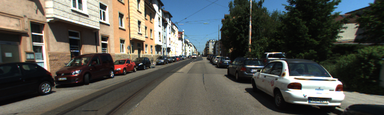}&
	\includegraphics[width=\turnwidthnew]{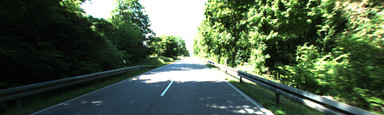}

	\\
	
	\includegraphics[width=\turnwidthnew]{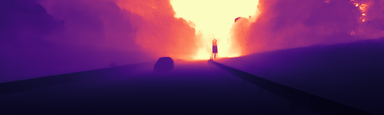}&
	\includegraphics[width=\turnwidthnew]{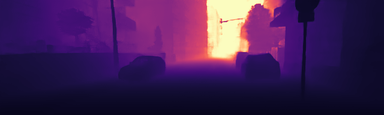}&
	\includegraphics[width=\turnwidthnew]{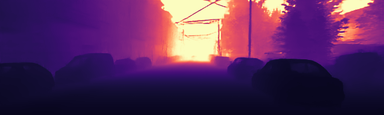}&
	\includegraphics[width=\turnwidthnew]{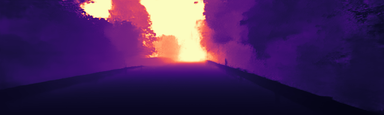}
	
\\
	
	\includegraphics[width=\turnwidthnew]{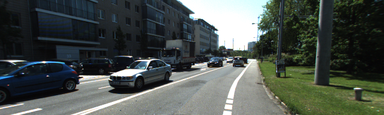}&
	\includegraphics[width=\turnwidthnew]{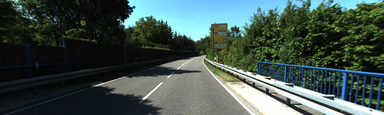}&
	\includegraphics[width=\turnwidthnew]{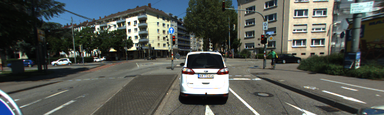}&
	\includegraphics[width=\turnwidthnew]{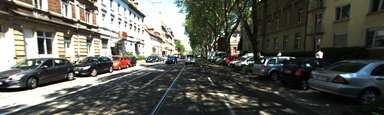}
	\\
	
	\includegraphics[width=\turnwidthnew]{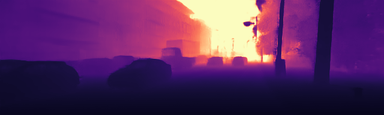}&
	\includegraphics[width=\turnwidthnew]{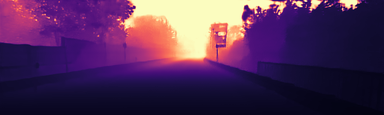}&
	\includegraphics[width=\turnwidthnew]{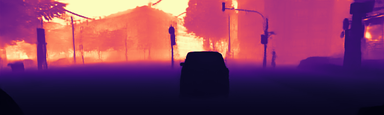}&
	\includegraphics[width=\turnwidthnew]{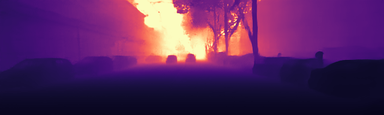}

\end{tabular}

	

%% file: figures/appendix_nuscenes/appendix_nuscenes.tex
\newcommand{\turnwidthnew}{0.17 \linewidth}
\newcommand{\len}{2mm}
\newcommand{\STAB}[1]{\begin{tabular}{@{}c@{}}#1\end{tabular}}

\centering

\begin{tabular}{@{\hskip 0mm}c@{\hskip 1mm}c@{\hskip 1mm}c@{\hskip 1mm}c@{\hskip 1mm}c@{\hskip 0mm}}

	\includegraphics[width=\turnwidthnew]{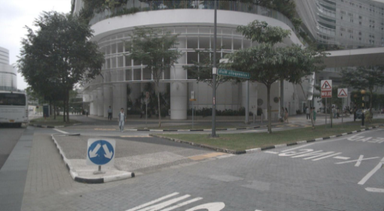}&
	\includegraphics[width=\turnwidthnew]{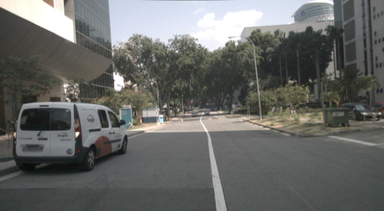}&
	\includegraphics[width=\turnwidthnew]{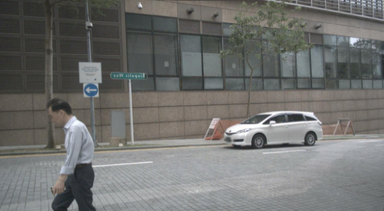}&
	\includegraphics[width=\turnwidthnew]{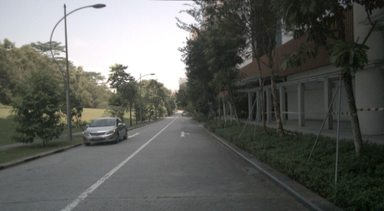}&
	\includegraphics[width=\turnwidthnew]{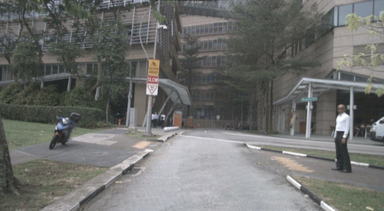}
	\\
	
	\includegraphics[width=\turnwidthnew]{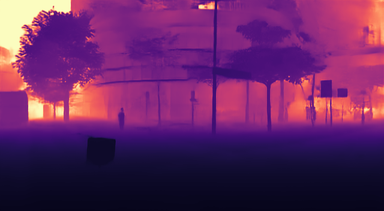}&
	\includegraphics[width=\turnwidthnew]{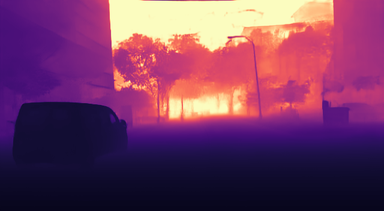}&
	\includegraphics[width=\turnwidthnew]{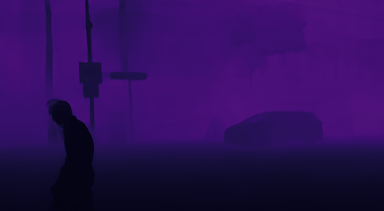}&
	\includegraphics[width=\turnwidthnew]{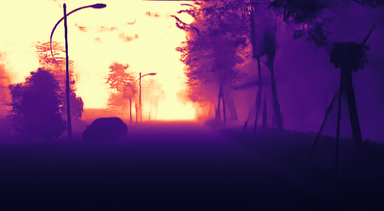}&
	\includegraphics[width=\turnwidthnew]{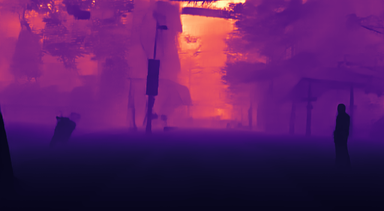}
	\\
	
	\includegraphics[width=\turnwidthnew]{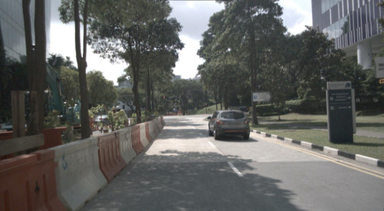}&
	\includegraphics[width=\turnwidthnew]{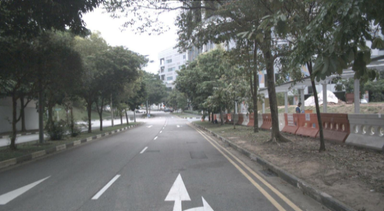}&
	\includegraphics[width=\turnwidthnew]{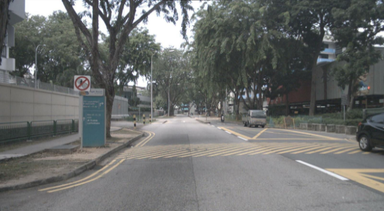}&
	\includegraphics[width=\turnwidthnew]{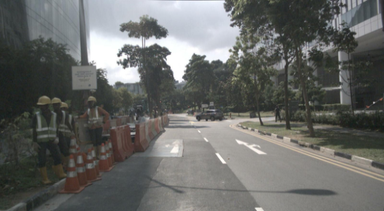}&
	\includegraphics[width=\turnwidthnew]{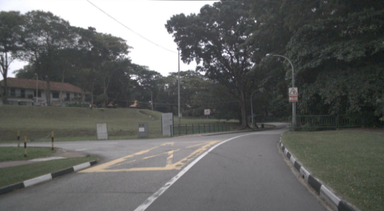}
	\\
	
	\includegraphics[width=\turnwidthnew]{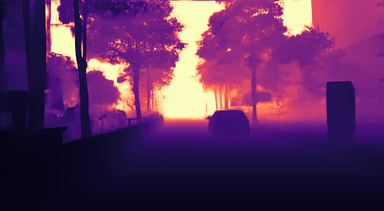}&
	\includegraphics[width=\turnwidthnew]{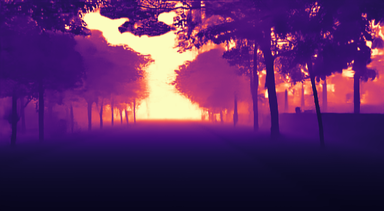}&
	\includegraphics[width=\turnwidthnew]{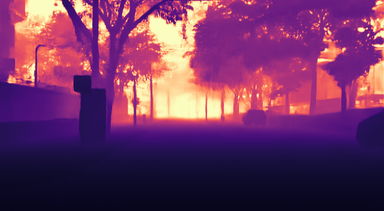}&
	\includegraphics[width=\turnwidthnew]{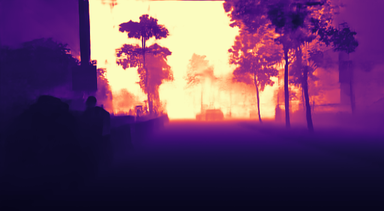}&
	\includegraphics[width=\turnwidthnew]{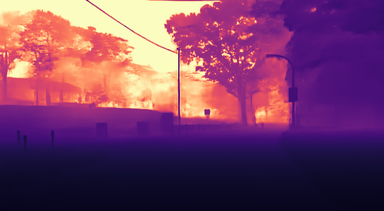}
	
	\\
	
	\includegraphics[width=\turnwidthnew]{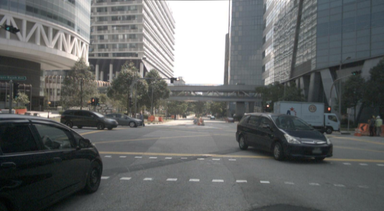}&
	\includegraphics[width=\turnwidthnew]{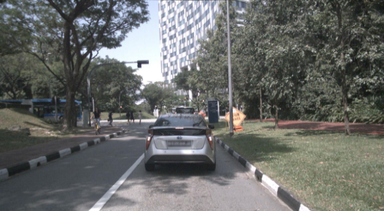}&
	\includegraphics[width=\turnwidthnew]{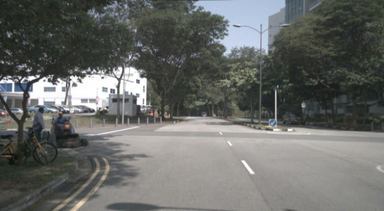}&
	\includegraphics[width=\turnwidthnew]{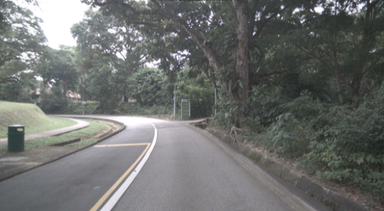}&
	\includegraphics[width=\turnwidthnew]{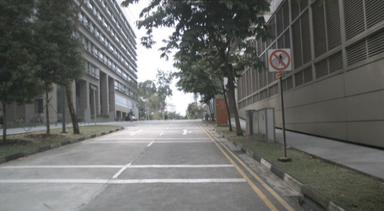}
	\\
	
	\includegraphics[width=\turnwidthnew]{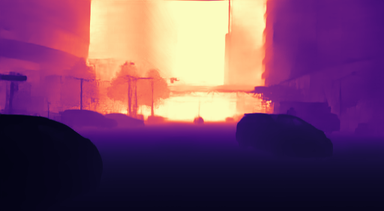}&
	\includegraphics[width=\turnwidthnew]{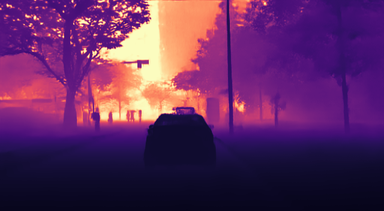}&
	\includegraphics[width=\turnwidthnew]{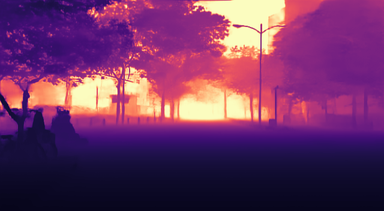}&
	\includegraphics[width=\turnwidthnew]{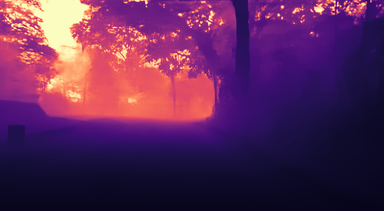}&
	\includegraphics[width=\turnwidthnew]{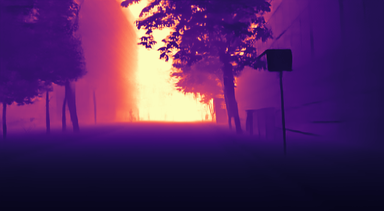}

\end{tabular}

	

%% file: figures/appendix_waymo/appendix_waymo.tex
\newcommand{\turnwidthnew}{0.17 \linewidth}
\newcommand{\len}{2mm}
\newcommand{\STAB}[1]{\begin{tabular}{@{}c@{}}#1\end{tabular}}

\centering

\begin{tabular}{@{\hskip 0mm}c@{\hskip 1mm}c@{\hskip 1mm}c@{\hskip 1mm}c@{\hskip 1mm}c@{\hskip 0mm}}

	\includegraphics[width=\turnwidthnew]{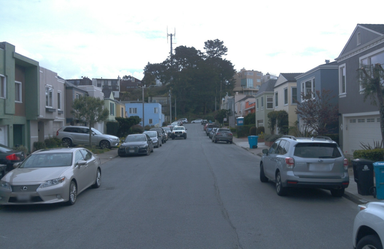}&
	\includegraphics[width=\turnwidthnew]{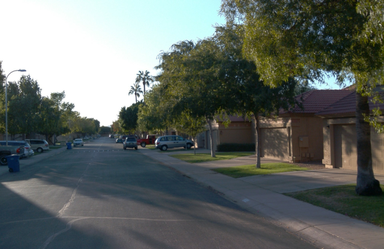}&
	\includegraphics[width=\turnwidthnew]{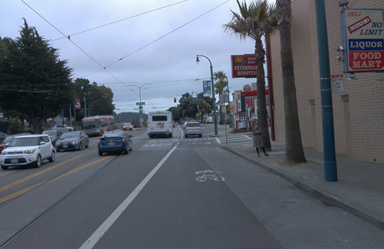}&
	\includegraphics[width=\turnwidthnew]{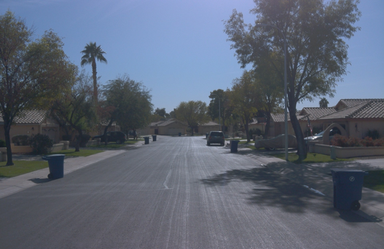}&
	\includegraphics[width=\turnwidthnew]{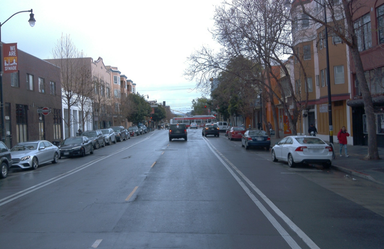}
	\\
	
	\includegraphics[width=\turnwidthnew]{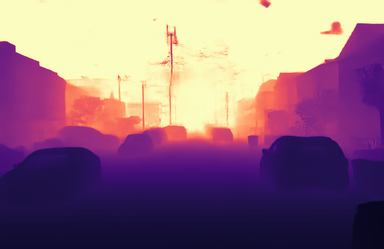}&
	\includegraphics[width=\turnwidthnew]{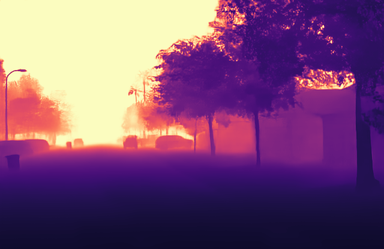}&
	\includegraphics[width=\turnwidthnew]{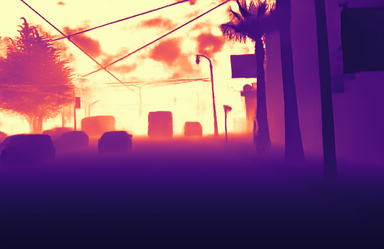}&
	\includegraphics[width=\turnwidthnew]{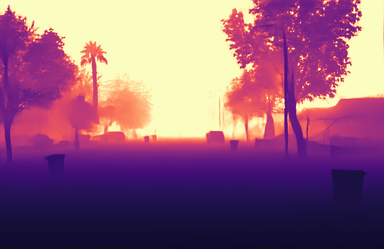}&
	\includegraphics[width=\turnwidthnew]{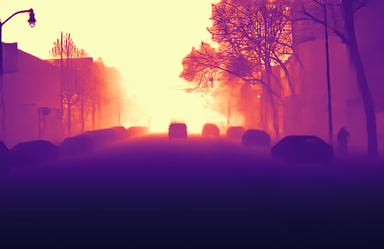}
	\\
	
	\includegraphics[width=\turnwidthnew]{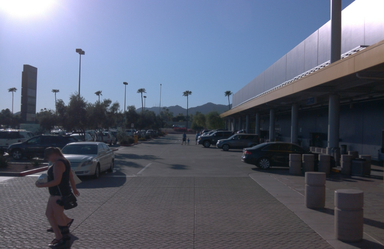}&
	\includegraphics[width=\turnwidthnew]{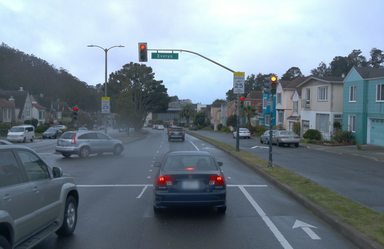}&
	\includegraphics[width=\turnwidthnew]{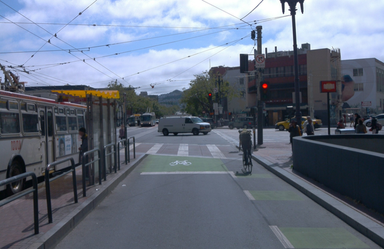}&
	\includegraphics[width=\turnwidthnew]{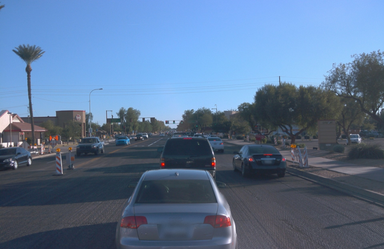}&
	\includegraphics[width=\turnwidthnew]{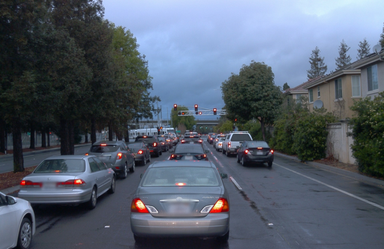}
	\\
	
	\includegraphics[width=\turnwidthnew]{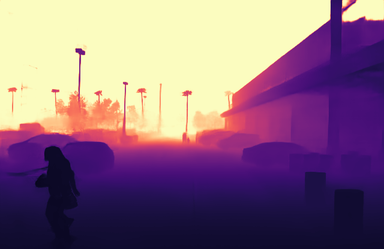}&
	\includegraphics[width=\turnwidthnew]{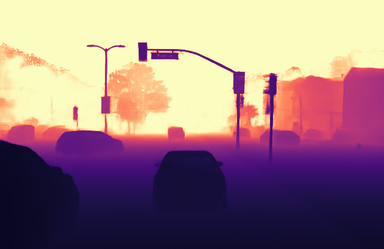}&
	\includegraphics[width=\turnwidthnew]{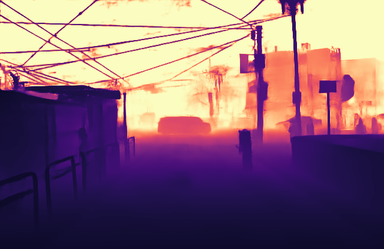}&
	\includegraphics[width=\turnwidthnew]{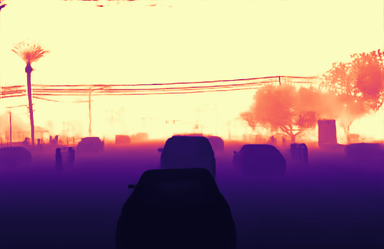}&
	\includegraphics[width=\turnwidthnew]{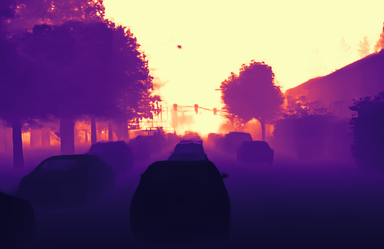}
	
\end{tabular}

	

%% file: figures/appendix_argoverse/appendix_argoverse.tex
\newcommand{\turnwidthnew}{0.17 \linewidth}
\newcommand{\len}{2mm}
\newcommand{\STAB}[1]{\begin{tabular}{@{}c@{}}#1\end{tabular}}

\centering

\begin{tabular}{@{\hskip 0mm}c@{\hskip 1mm}c@{\hskip 1mm}c@{\hskip 1mm}c@{\hskip 1mm}c@{\hskip 0mm}}

	\includegraphics[width=\turnwidthnew]{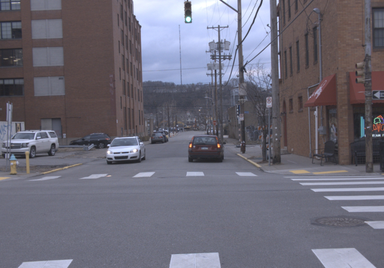}&
	\includegraphics[width=\turnwidthnew]{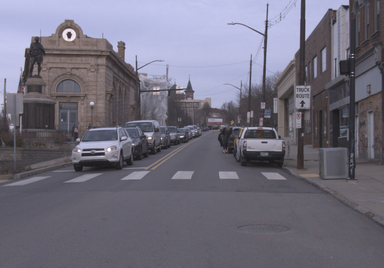}&
	\includegraphics[width=\turnwidthnew]{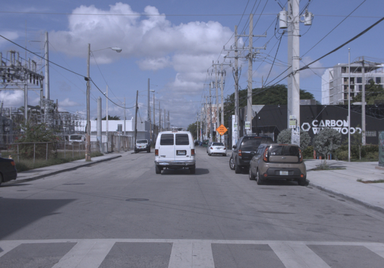}&
	\includegraphics[width=\turnwidthnew]{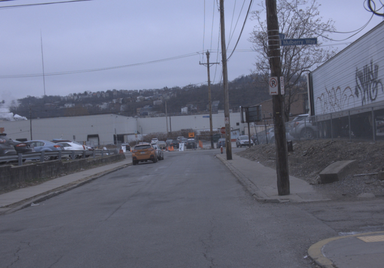}&
	\includegraphics[width=\turnwidthnew]{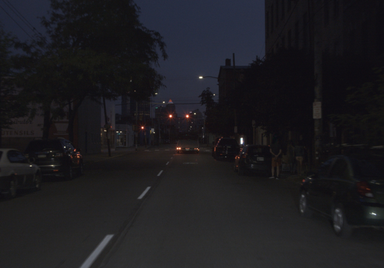}
	\\
	
	\includegraphics[width=\turnwidthnew]{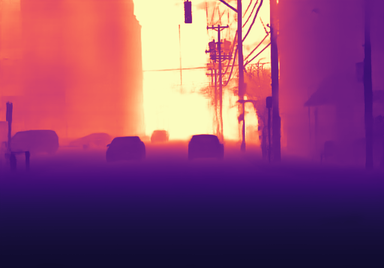}&
	\includegraphics[width=\turnwidthnew]{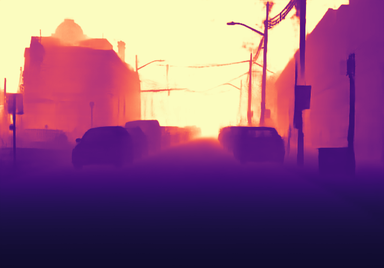}&
	\includegraphics[width=\turnwidthnew]{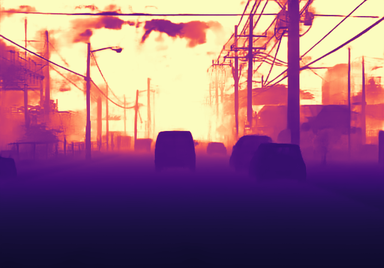}&
	\includegraphics[width=\turnwidthnew]{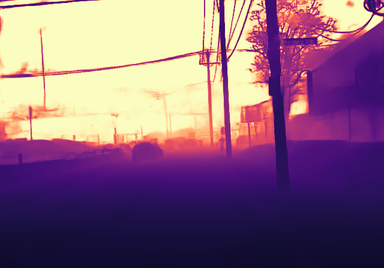}&
	\includegraphics[width=\turnwidthnew]{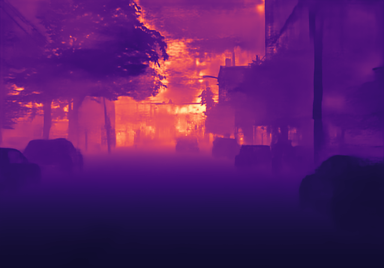}
	\\
	
	\includegraphics[width=\turnwidthnew]{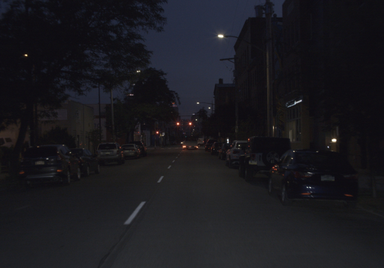}&
	\includegraphics[width=\turnwidthnew]{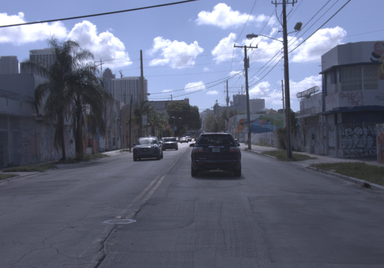}&
	\includegraphics[width=\turnwidthnew]{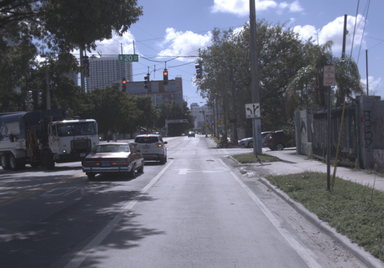}&
	\includegraphics[width=\turnwidthnew]{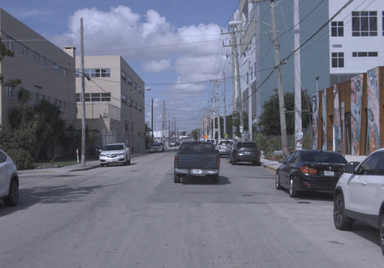}&
	\includegraphics[width=\turnwidthnew]{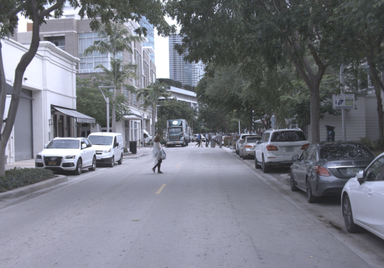}
	\\
	
	\includegraphics[width=\turnwidthnew]{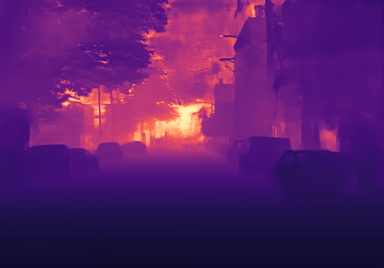}&
	\includegraphics[width=\turnwidthnew]{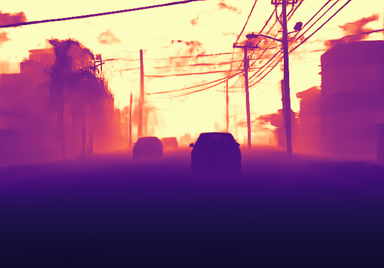}&
	\includegraphics[width=\turnwidthnew]{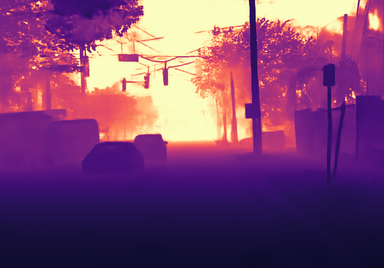}&
	\includegraphics[width=\turnwidthnew]{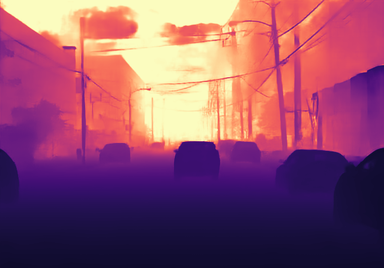}&
	\includegraphics[width=\turnwidthnew]{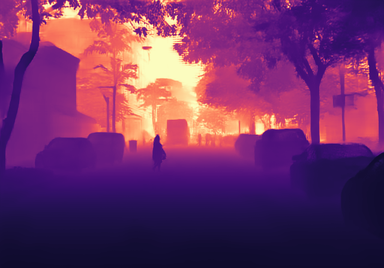}

\end{tabular}

	

%% file: figures/appendix_ddad/appendix_ddad.tex
\newcommand{\turnwidthnew}{0.17 \linewidth}
\newcommand{\len}{2mm}
\newcommand{\STAB}[1]{\begin{tabular}{@{}c@{}}#1\end{tabular}}

\centering

\begin{tabular}{@{\hskip 0mm}c@{\hskip 1mm}c@{\hskip 1mm}c@{\hskip 1mm}c@{\hskip 1mm}c@{\hskip 0mm}}

\includegraphics[width=\turnwidthnew]{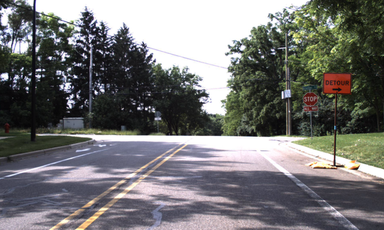}&
\includegraphics[width=\turnwidthnew]{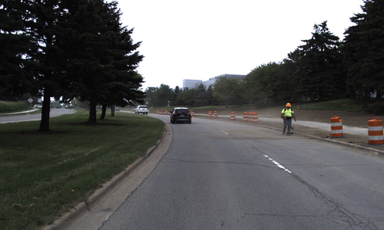}&
\includegraphics[width=\turnwidthnew]{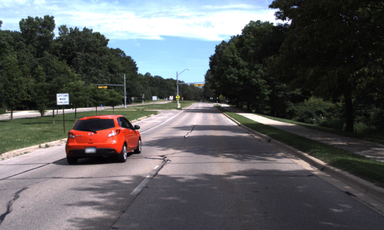}&
\includegraphics[width=\turnwidthnew]{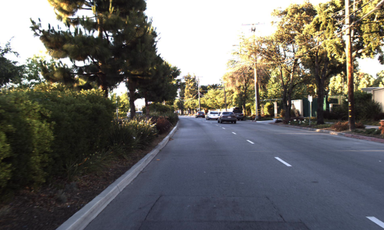}&
\includegraphics[width=\turnwidthnew]{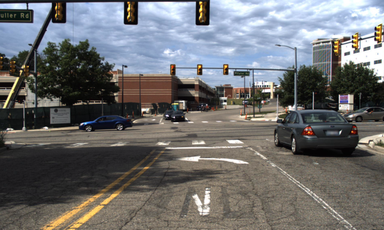}
\\

\includegraphics[width=\turnwidthnew]{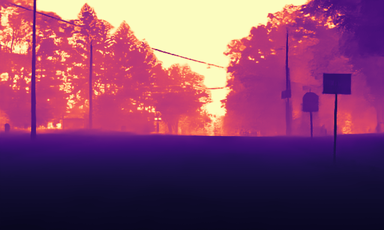}&
\includegraphics[width=\turnwidthnew]{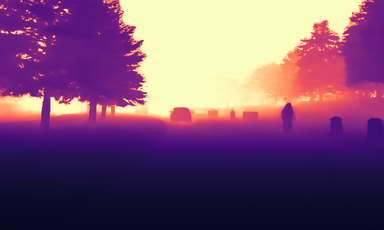}&
\includegraphics[width=\turnwidthnew]{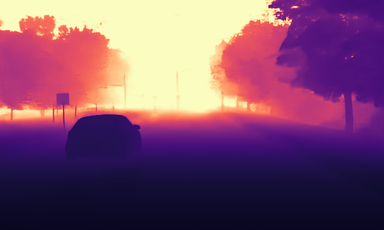}&
\includegraphics[width=\turnwidthnew]{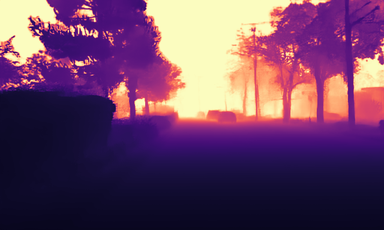}&
\includegraphics[width=\turnwidthnew]{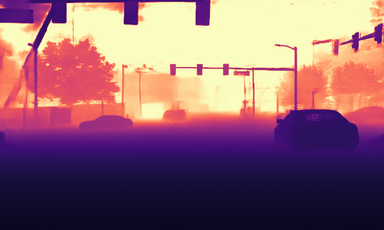}

\\

\includegraphics[width=\turnwidthnew]{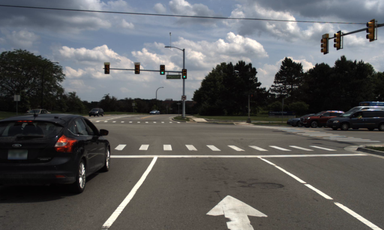}&
\includegraphics[width=\turnwidthnew]{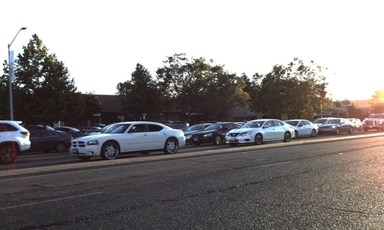}&
\includegraphics[width=\turnwidthnew]{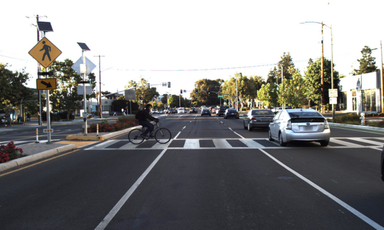}&
\includegraphics[width=\turnwidthnew]{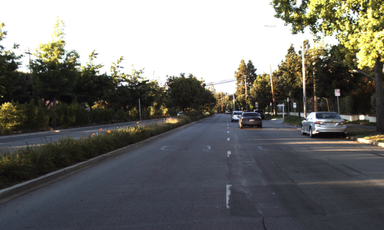}&
\includegraphics[width=\turnwidthnew]{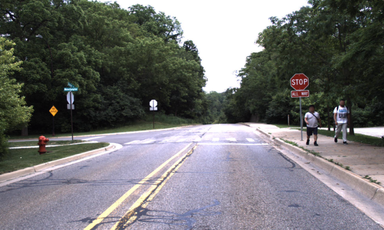}
\\

\includegraphics[width=\turnwidthnew]{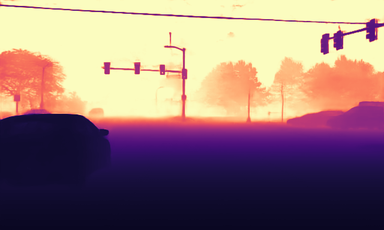}&
\includegraphics[width=\turnwidthnew]{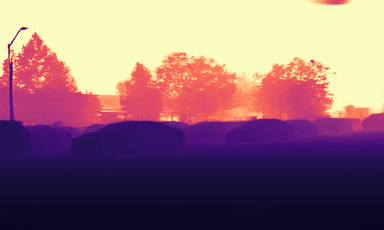}&
\includegraphics[width=\turnwidthnew]{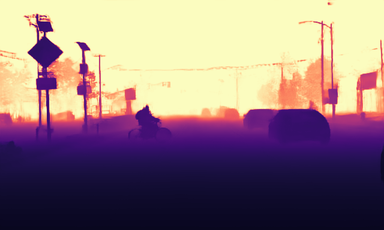}&
\includegraphics[width=\turnwidthnew]{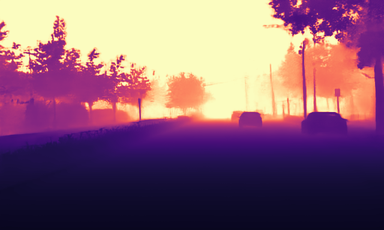}&
\includegraphics[width=\turnwidthnew]{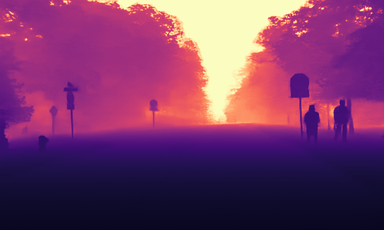}

\end{tabular}

	

%% file: paper.bbl
\begin{thebibliography}{100}
\providecommand{\url}[1]{#1}
\csname url@rmstyle\endcsname
\providecommand{\newblock}{\relax}
\providecommand{\bibinfo}[2]{#2}
\providecommand\BIBentrySTDinterwordspacing{\spaceskip=0pt\relax}
\providecommand\BIBentryALTinterwordstretchfactor{4}
\providecommand\BIBentryALTinterwordspacing{\spaceskip=\fontdimen2\font plus
\BIBentryALTinterwordstretchfactor\fontdimen3\font minus
  \fontdimen4\font\relax}
\providecommand\BIBforeignlanguage[2]{{%
\expandafter\ifx\csname l@#1\endcsname\relax
\typeout{** WARNING: IEEEtran.bst: No hyphenation pattern has been}%
\typeout{** loaded for the language `#1'. Using the pattern for}%
\typeout{** the default language instead.}%
\else
\language=\csname l@#1\endcsname
\fi
#2}}

\bibitem{eigen2014depth}
D.~Eigen, C.~Puhrsch, and R.~Fergus, ``Depth map prediction from a single image
  using a multi-scale deep network,'' \emph{Advances in neural information
  processing systems}, vol.~27, 2014.

\bibitem{laina2016deeper}
I.~Laina, C.~Rupprecht, V.~Belagiannis, F.~Tombari, and N.~Navab, ``Deeper
  depth prediction with fully convolutional residual networks,'' in \emph{2016
  Fourth international conference on 3D vision (3DV)}.\hskip 1em plus 0.5em
  minus 0.4em\relax IEEE, 2016, pp. 239--248.

\bibitem{cs2018depthnet}
A.~CS~Kumar, S.~M. Bhandarkar, and M.~Prasad, ``Depthnet: A recurrent neural
  network architecture for monocular depth prediction,'' in \emph{Proceedings
  of the IEEE Conference on Computer Vision and Pattern Recognition Workshops},
  2018, pp. 283--291.

\bibitem{mancini2017toward}
M.~Mancini, G.~Costante, P.~Valigi, T.~A. Ciarfuglia, J.~Delmerico, and
  D.~Scaramuzza, ``Toward domain independence for learning-based monocular
  depth estimation,'' \emph{IEEE Robotics and Automation Letters}, vol.~2,
  no.~3, pp. 1778--1785, 2017.

\bibitem{li2015depth}
B.~Li, C.~Shen, Y.~Dai, A.~Van Den~Hengel, and M.~He, ``Depth and surface
  normal estimation from monocular images using regression on deep features and
  hierarchical crfs,'' in \emph{Proceedings of the IEEE conference on computer
  vision and pattern recognition}, 2015, pp. 1119--1127.

\bibitem{yin2019enforcing}
W.~Yin, Y.~Liu, C.~Shen, and Y.~Yan, ``Enforcing geometric constraints of
  virtual normal for depth prediction,'' in \emph{Proceedings of the IEEE/CVF
  International Conference on Computer Vision}, 2019, pp. 5684--5693.

\bibitem{yuan2022newcrfs}
W.~Yuan, X.~Gu, Z.~Dai, S.~Zhu, and P.~Tan, ``Newcrfs: Neural window
  fully-connected crfs for monocular depth estimation,'' in \emph{Proceedings
  of the IEEE Conference on Computer Vision and Pattern Recognition}, 2022.

\bibitem{bhat2021adabins}
S.~F. Bhat, I.~Alhashim, and P.~Wonka, ``Adabins: Depth estimation using
  adaptive bins,'' in \emph{Proceedings of the IEEE/CVF Conference on Computer
  Vision and Pattern Recognition}, 2021, pp. 4009--4018.

\bibitem{agarwal2023attention}
A.~Agarwal and C.~Arora, ``Attention attention everywhere: Monocular depth
  prediction with skip attention,'' in \emph{Proceedings of the IEEE/CVF Winter
  Conference on Applications of Computer Vision}, 2023, pp. 5861--5870.

\bibitem{godard2019digging}
C.~Godard, O.~Mac~Aodha, M.~Firman, and G.~J. Brostow, ``Digging into
  self-supervised monocular depth estimation,'' in \emph{Proceedings of the
  IEEE/CVF International Conference on Computer Vision}, 2019, pp. 3828--3838.

\bibitem{zhan2018unsupervised}
H.~Zhan, R.~Garg, C.~S. Weerasekera, K.~Li, H.~Agarwal, and I.~Reid,
  ``Unsupervised learning of monocular depth estimation and visual odometry
  with deep feature reconstruction,'' in \emph{Proceedings of the IEEE
  conference on computer vision and pattern recognition}, 2018, pp. 340--349.

\bibitem{li2018undeepvo}
R.~Li, S.~Wang, Z.~Long, and D.~Gu, ``Undeepvo: Monocular visual odometry
  through unsupervised deep learning,'' in \emph{2018 IEEE international
  conference on robotics and automation (ICRA)}.\hskip 1em plus 0.5em minus
  0.4em\relax IEEE, 2018, pp. 7286--7291.

\bibitem{zhou_diffnet}
H.~Zhou, D.~Greenwood, and S.~Taylor, ``Self-supervised monocular depth
  estimation with internal feature fusion,'' in \emph{British Machine Vision
  Conference (BMVC)}, 2021.

\bibitem{guizilini20203d}
V.~Guizilini, R.~Ambrus, S.~Pillai, A.~Raventos, and A.~Gaidon, ``3d packing
  for self-supervised monocular depth estimation,'' in \emph{Proceedings of the
  IEEE/CVF conference on computer vision and pattern recognition}, 2020, pp.
  2485--2494.

\bibitem{wang2023planedepth}
R.~Wang, Z.~Yu, and S.~Gao, ``Planedepth: Self-supervised depth estimation via
  orthogonal planes,'' in \emph{Proceedings of the IEEE/CVF Conference on
  Computer Vision and Pattern Recognition}, 2023, pp. 21\,425--21\,434.

\bibitem{zhao2022monovit}
C.~Zhao, Y.~Zhang, M.~Poggi, F.~Tosi, X.~Guo, Z.~Zhu, G.~Huang, Y.~Tang, and
  S.~Mattoccia, ``Monovit: Self-supervised monocular depth estimation with a
  vision transformer,'' in \emph{2022 International Conference on 3D Vision
  (3DV)}.\hskip 1em plus 0.5em minus 0.4em\relax IEEE, 2022, pp. 668--678.

\bibitem{zhou2017unsupervised}
T.~Zhou, M.~Brown, N.~Snavely, and D.~G. Lowe, ``Unsupervised learning of depth
  and ego-motion from video,'' in \emph{Proceedings of the IEEE conference on
  computer vision and pattern recognition}, 2017, pp. 1851--1858.

\bibitem{piccinelli2023idisc}
L.~Piccinelli, C.~Sakaridis, and F.~Yu, ``idisc: Internal discretization for
  monocular depth estimation,'' in \emph{Proceedings of the IEEE/CVF Conference
  on Computer Vision and Pattern Recognition}, 2023, pp. 21\,477--21\,487.

\bibitem{koledic2023moft}
K.~Koledi{\'c}, I.~Cvi{\v{s}}i{\'c}, I.~Markovi{\'c}, and I.~Petrovi{\'c},
  ``Moft: Monocular odometry based on deep depth and careful feature selection
  and tracking,'' in \emph{2023 IEEE International Conference on Robotics and
  Automation (ICRA)}.\hskip 1em plus 0.5em minus 0.4em\relax IEEE, 2023, pp.
  6175--6181.

\bibitem{zhao2021camera}
Y.~Zhao, S.~Kong, and C.~Fowlkes, ``Camera pose matters: Improving depth
  prediction by mitigating pose distribution bias,'' in \emph{Proceedings of
  the IEEE/CVF Conference on Computer Vision and Pattern Recognition}, 2021,
  pp. 15\,759--15\,768.

\bibitem{facil2019cam}
J.~M. Facil, B.~Ummenhofer, H.~Zhou, L.~Montesano, T.~Brox, and J.~Civera,
  ``Cam-convs: Camera-aware multi-scale convolutions for single-view depth,''
  in \emph{Proceedings of the IEEE/CVF Conference on Computer Vision and
  Pattern Recognition}, 2019, pp. 11\,826--11\,835.

\bibitem{he2018learning}
L.~He, G.~Wang, and Z.~Hu, ``Learning depth from single images with deep neural
  network embedding focal length,'' \emph{IEEE Transactions on Image
  Processing}, vol.~27, no.~9, pp. 4676--4689, 2018.

\bibitem{chen2016single}
W.~Chen, Z.~Fu, D.~Yang, and J.~Deng, ``Single-image depth perception in the
  wild,'' \emph{Advances in neural information processing systems}, vol.~29,
  2016.

\bibitem{yin2021learning}
W.~Yin, J.~Zhang, O.~Wang, S.~Niklaus, L.~Mai, S.~Chen, and C.~Shen, ``Learning
  to recover 3d scene shape from a single image,'' in \emph{Proceedings of the
  IEEE/CVF Conference on Computer Vision and Pattern Recognition}, 2021, pp.
  204--213.

\bibitem{ranftl2020towards}
R.~Ranftl, K.~Lasinger, D.~Hafner, K.~Schindler, and V.~Koltun, ``Towards
  robust monocular depth estimation: Mixing datasets for zero-shot
  cross-dataset transfer,'' \emph{IEEE transactions on pattern analysis and
  machine intelligence}, vol.~44, no.~3, pp. 1623--1637, 2020.

\bibitem{xian2018monocular}
K.~Xian, C.~Shen, Z.~Cao, H.~Lu, Y.~Xiao, R.~Li, and Z.~Luo, ``Monocular
  relative depth perception with web stereo data supervision,'' in
  \emph{Proceedings of the IEEE Conference on Computer Vision and Pattern
  Recognition}, 2018, pp. 311--320.

\bibitem{yin2020diversedepth}
W.~Yin, X.~Wang, C.~Shen, Y.~Liu, Z.~Tian, S.~Xu, C.~Sun, and D.~Renyin,
  ``Diversedepth: Affine-invariant depth prediction using diverse data,''
  \emph{arXiv preprint arXiv:2002.00569}, 2020.

\bibitem{wang2019web}
C.~Wang, S.~Lucey, F.~Perazzi, and O.~Wang, ``Web stereo video supervision for
  depth prediction from dynamic scenes,'' in \emph{2019 International
  Conference on 3D Vision (3DV)}.\hskip 1em plus 0.5em minus 0.4em\relax IEEE,
  2019, pp. 348--357.

\bibitem{xian2020structure}
K.~Xian, J.~Zhang, O.~Wang, L.~Mai, Z.~Lin, and Z.~Cao, ``Structure-guided
  ranking loss for single image depth prediction,'' in \emph{Proceedings of the
  IEEE/CVF Conference on Computer Vision and Pattern Recognition}, 2020, pp.
  611--620.

\bibitem{li2018megadepth}
Z.~Li and N.~Snavely, ``Megadepth: Learning single-view depth prediction from
  internet photos,'' in \emph{Proceedings of the IEEE conference on computer
  vision and pattern recognition}, 2018, pp. 2041--2050.

\bibitem{dosovitskiy2017carla}
A.~Dosovitskiy, G.~Ros, F.~Codevilla, A.~Lopez, and V.~Koltun, ``Carla: An open
  urban driving simulator,'' in \emph{Conference on robot learning}.\hskip 1em
  plus 0.5em minus 0.4em\relax PMLR, 2017, pp. 1--16.

\bibitem{ronneberger2015u}
O.~Ronneberger, P.~Fischer, and T.~Brox, ``U-net: Convolutional networks for
  biomedical image segmentation,'' in \emph{International Conference on Medical
  image computing and computer-assisted intervention}.\hskip 1em plus 0.5em
  minus 0.4em\relax Springer, 2015, pp. 234--241.

\bibitem{xu2018structured}
D.~Xu, W.~Wang, H.~Tang, H.~Liu, N.~Sebe, and E.~Ricci, ``Structured attention
  guided convolutional neural fields for monocular depth estimation,'' in
  \emph{Proceedings of the IEEE conference on computer vision and pattern
  recognition}, 2018, pp. 3917--3925.

\bibitem{liu2015deep}
F.~Liu, C.~Shen, and G.~Lin, ``Deep convolutional neural fields for depth
  estimation from a single image,'' in \emph{Proceedings of the IEEE conference
  on computer vision and pattern recognition}, 2015, pp. 5162--5170.

\bibitem{jung2017depth}
H.~Jung, Y.~Kim, D.~Min, C.~Oh, and K.~Sohn, ``Depth prediction from a single
  image with conditional adversarial networks,'' in \emph{2017 IEEE
  International Conference on Image Processing (ICIP)}.\hskip 1em plus 0.5em
  minus 0.4em\relax IEEE, 2017, pp. 1717--1721.

\bibitem{lore2018generative}
K.~G. Lore, K.~Reddy, M.~Giering, and E.~A. Bernal, ``Generative adversarial
  networks for depth map estimation from rgb video,'' in \emph{2018 IEEE/CVF
  Conference on Computer Vision and Pattern Recognition Workshops
  (CVPRW)}.\hskip 1em plus 0.5em minus 0.4em\relax IEEE, 2018, pp.
  1258--12\,588.

\bibitem{dosovitskiy2020image}
A.~Dosovitskiy, L.~Beyer, A.~Kolesnikov, D.~Weissenborn, X.~Zhai,
  T.~Unterthiner, M.~Dehghani, M.~Minderer, G.~Heigold, S.~Gelly,
  \emph{et~al.}, ``An image is worth 16x16 words: Transformers for image
  recognition at scale,'' \emph{arXiv preprint arXiv:2010.11929}, 2020.

\bibitem{li2022depthformer}
Z.~Li, Z.~Chen, X.~Liu, and J.~Jiang, ``Depthformer: Exploiting long-range
  correlation and local information for accurate monocular depth estimation,''
  \emph{arXiv preprint arXiv:2203.14211}, 2022.

\bibitem{zhao2023unleashing}
W.~Zhao, Y.~Rao, Z.~Liu, B.~Liu, J.~Zhou, and J.~Lu, ``Unleashing text-to-image
  diffusion models for visual perception,'' \emph{arXiv preprint
  arXiv:2303.02153}, 2023.

\bibitem{oquab2023dinov2}
M.~Oquab, T.~Darcet, T.~Moutakanni, H.~Vo, M.~Szafraniec, V.~Khalidov,
  P.~Fernandez, D.~Haziza, F.~Massa, A.~El-Nouby, \emph{et~al.}, ``Dinov2:
  Learning robust visual features without supervision,'' \emph{arXiv preprint
  arXiv:2304.07193}, 2023.

\bibitem{garg2016unsupervised}
R.~Garg, V.~K. Bg, G.~Carneiro, and I.~Reid, ``Unsupervised cnn for single view
  depth estimation: Geometry to the rescue,'' in \emph{Computer Vision--ECCV
  2016: 14th European Conference, Amsterdam, The Netherlands, October 11-14,
  2016, Proceedings, Part VIII 14}.\hskip 1em plus 0.5em minus 0.4em\relax
  Springer, 2016, pp. 740--756.

\bibitem{godard2017unsupervised}
C.~Godard, O.~Mac~Aodha, and G.~J. Brostow, ``Unsupervised monocular depth
  estimation with left-right consistency,'' in \emph{Proceedings of the IEEE
  conference on computer vision and pattern recognition}, 2017, pp. 270--279.

\bibitem{han2019deepvio}
L.~Han, Y.~Lin, G.~Du, and S.~Lian, ``Deepvio: Self-supervised deep learning of
  monocular visual inertial odometry using 3d geometric constraints,'' in
  \emph{2019 IEEE/RSJ International Conference on Intelligent Robots and
  Systems (IROS)}.\hskip 1em plus 0.5em minus 0.4em\relax IEEE, 2019, pp.
  6906--6913.

\bibitem{koledic2023towards}
K.~Koledi{\'c}, I.~Markovi{\'c}, and I.~Petrovi{\'c}, ``Towards camera
  parameters invariant monocular depth estimation in autonomous driving,'' in
  \emph{2023 European Conference on Mobile Robots (ECMR)}.\hskip 1em plus 0.5em
  minus 0.4em\relax IEEE, 2023, pp. 1--7.

\bibitem{hoffman2016fcns}
J.~Hoffman, D.~Wang, F.~Yu, and T.~Darrell, ``Fcns in the wild: Pixel-level
  adversarial and constraint-based adaptation,'' \emph{arXiv preprint
  arXiv:1612.02649}, 2016.

\bibitem{tzeng2017adversarial}
E.~Tzeng, J.~Hoffman, K.~Saenko, and T.~Darrell, ``Adversarial discriminative
  domain adaptation,'' in \emph{Proceedings of the IEEE conference on computer
  vision and pattern recognition}, 2017, pp. 7167--7176.

\bibitem{atapour2018real}
A.~Atapour-Abarghouei and T.~P. Breckon, ``Real-time monocular depth estimation
  using synthetic data with domain adaptation via image style transfer,'' in
  \emph{Proceedings of the IEEE conference on computer vision and pattern
  recognition}, 2018, pp. 2800--2810.

\bibitem{hoffman2018cycada}
J.~Hoffman, E.~Tzeng, T.~Park, J.-Y. Zhu, P.~Isola, K.~Saenko, A.~Efros, and
  T.~Darrell, ``Cycada: Cycle-consistent adversarial domain adaptation,'' in
  \emph{International conference on machine learning}.\hskip 1em plus 0.5em
  minus 0.4em\relax Pmlr, 2018, pp. 1989--1998.

\bibitem{kundu2018adadepth}
J.~N. Kundu, P.~K. Uppala, A.~Pahuja, and R.~V. Babu, ``Adadepth: Unsupervised
  content congruent adaptation for depth estimation,'' in \emph{Proceedings of
  the IEEE conference on computer vision and pattern recognition}, 2018, pp.
  2656--2665.

\bibitem{zheng2018t2net}
C.~Zheng, T.-J. Cham, and J.~Cai, ``T2net: Synthetic-to-realistic translation
  for solving single-image depth estimation tasks,'' in \emph{Proceedings of
  the European conference on computer vision (ECCV)}, 2018, pp. 767--783.

\bibitem{chen2019crdoco}
Y.-C. Chen, Y.-Y. Lin, M.-H. Yang, and J.-B. Huang, ``Crdoco: Pixel-level
  domain transfer with cross-domain consistency,'' in \emph{Proceedings of the
  IEEE/CVF conference on computer vision and pattern recognition}, 2019, pp.
  1791--1800.

\bibitem{zhao2019geometry}
S.~Zhao, H.~Fu, M.~Gong, and D.~Tao, ``Geometry-aware symmetric domain
  adaptation for monocular depth estimation,'' in \emph{Proceedings of the
  IEEE/CVF Conference on Computer Vision and Pattern Recognition}, 2019, pp.
  9788--9798.

\bibitem{akada2022self}
H.~Akada, S.~F. Bhat, I.~Alhashim, and P.~Wonka, ``Self-supervised learning of
  domain invariant features for depth estimation,'' in \emph{Proceedings of the
  IEEE/CVF Winter Conference on Applications of Computer Vision}, 2022, pp.
  3377--3387.

\bibitem{lo2022learning}
S.-Y. Lo, W.~Wang, J.~Thomas, J.~Zheng, V.~M. Patel, and C.-H. Kuo, ``Learning
  feature decomposition for domain adaptive monocular depth estimation,'' in
  \emph{2022 IEEE/RSJ International Conference on Intelligent Robots and
  Systems (IROS)}.\hskip 1em plus 0.5em minus 0.4em\relax IEEE, 2022, pp.
  8376--8382.

\bibitem{gurram2021monocular}
A.~Gurram, A.~F. Tuna, F.~Shen, O.~Urfalioglu, and A.~M. L{\'o}pez, ``Monocular
  depth estimation through virtual-world supervision and real-world sfm
  self-supervision,'' \emph{IEEE Transactions on Intelligent Transportation
  Systems}, vol.~23, no.~8, pp. 12\,738--12\,751, 2021.

\bibitem{pnvr2020sharingan}
K.~PNVR, H.~Zhou, and D.~Jacobs, ``Sharingan: Combining synthetic and real data
  for unsupervised geometry estimation,'' in \emph{Proceedings of the IEEE/CVF
  Conference on Computer Vision and Pattern Recognition}, 2020, pp.
  13\,974--13\,983.

\bibitem{cheng2020s}
B.~Cheng, I.~S. Saggu, R.~Shah, G.~Bansal, and D.~Bharadia, ``S 3 net:
  Semantic-aware self-supervised depth estimation with monocular videos and
  synthetic data,'' in \emph{European Conference on Computer Vision}.\hskip 1em
  plus 0.5em minus 0.4em\relax Springer, 2020, pp. 52--69.

\bibitem{lopez2023desc}
A.~Lopez-Rodriguez and K.~Mikolajczyk, ``Desc: Domain adaptation for depth
  estimation via semantic consistency,'' \emph{International Journal of
  Computer Vision}, vol. 131, no.~3, pp. 752--771, 2023.

\bibitem{saha2021learning}
S.~Saha, A.~Obukhov, D.~P. Paudel, M.~Kanakis, Y.~Chen, S.~Georgoulis, and
  L.~Van~Gool, ``Learning to relate depth and semantics for unsupervised domain
  adaptation,'' in \emph{Proceedings of the IEEE/CVF Conference on Computer
  Vision and Pattern Recognition}, 2021, pp. 8197--8207.

\bibitem{dijk2019neural}
T.~v. Dijk and G.~d. Croon, ``How do neural networks see depth in single
  images?'' in \emph{Proceedings of the IEEE/CVF International Conference on
  Computer Vision}, 2019, pp. 2183--2191.

\bibitem{peng2021excavating}
R.~Peng, R.~Wang, Y.~Lai, L.~Tang, and Y.~Cai, ``Excavating the potential
  capacity of self-supervised monocular depth estimation,'' in
  \emph{Proceedings of the IEEE/CVF International Conference on Computer
  Vision}, 2021, pp. 15\,560--15\,569.

\bibitem{he2022ra}
M.~He, L.~Hui, Y.~Bian, J.~Ren, J.~Xie, and J.~Yang, ``Ra-depth: Resolution
  adaptive self-supervised monocular depth estimation,'' in \emph{European
  Conference on Computer Vision}.\hskip 1em plus 0.5em minus 0.4em\relax
  Springer, 2022, pp. 565--581.

\bibitem{geiger2013vision}
A.~Geiger, P.~Lenz, C.~Stiller, and R.~Urtasun, ``Vision meets robotics: The
  kitti dataset,'' \emph{The International Journal of Robotics Research},
  vol.~32, no.~11, pp. 1231--1237, 2013.

\bibitem{maddern20171}
W.~Maddern, G.~Pascoe, C.~Linegar, and P.~Newman, ``1 year, 1000 km: The oxford
  robotcar dataset,'' \emph{The International Journal of Robotics Research},
  vol.~36, no.~1, pp. 3--15, 2017.

\bibitem{cordts2016cityscapes}
M.~Cordts, M.~Omran, S.~Ramos, T.~Rehfeld, M.~Enzweiler, R.~Benenson,
  U.~Franke, S.~Roth, and B.~Schiele, ``The cityscapes dataset for semantic
  urban scene understanding,'' in \emph{Proceedings of the IEEE conference on
  computer vision and pattern recognition}, 2016, pp. 3213--3223.

\bibitem{Chang_2019_CVPR}
M.-F. Chang, J.~Lambert, P.~Sangkloy, J.~Singh, S.~Bak, A.~Hartnett, D.~Wang,
  P.~Carr, S.~Lucey, D.~Ramanan, and J.~Hays, ``Argoverse: 3d tracking and
  forecasting with rich maps,'' in \emph{Proceedings of the IEEE/CVF Conference
  on Computer Vision and Pattern Recognition (CVPR)}, June 2019.

\bibitem{nuscenes2019}
H.~Caesar, V.~Bankiti, A.~H. Lang, S.~Vora, V.~E. Liong, Q.~Xu, A.~Krishnan,
  Y.~Pan, G.~Baldan, and O.~Beijbom, ``nuscenes: A multimodal dataset for
  autonomous driving,'' \emph{arXiv preprint arXiv:1903.11027}, 2019.

\bibitem{Sun_2020_CVPR}
P.~Sun, H.~Kretzschmar, X.~Dotiwalla, A.~Chouard, V.~Patnaik, P.~Tsui, J.~Guo,
  Y.~Zhou, Y.~Chai, B.~Caine, V.~Vasudevan, W.~Han, J.~Ngiam, H.~Zhao,
  A.~Timofeev, S.~Ettinger, M.~Krivokon, A.~Gao, A.~Joshi, Y.~Zhang, J.~Shlens,
  Z.~Chen, and D.~Anguelov, ``Scalability in perception for autonomous driving:
  Waymo open dataset,'' in \emph{Proceedings of the IEEE/CVF Conference on
  Computer Vision and Pattern Recognition (CVPR)}, June 2020.

\bibitem{mildenhall2021nerf}
B.~Mildenhall, P.~P. Srinivasan, M.~Tancik, J.~T. Barron, R.~Ramamoorthi, and
  R.~Ng, ``Nerf: Representing scenes as neural radiance fields for view
  synthesis,'' \emph{Communications of the ACM}, vol.~65, no.~1, pp. 99--106,
  2021.

\bibitem{he2016deep}
K.~He, X.~Zhang, S.~Ren, and J.~Sun, ``Deep residual learning for image
  recognition,'' in \emph{Proceedings of the IEEE conference on computer vision
  and pattern recognition}, 2016, pp. 770--778.

\bibitem{liu2022convnet}
Z.~Liu, H.~Mao, C.-Y. Wu, C.~Feichtenhofer, T.~Darrell, and S.~Xie, ``A convnet
  for the 2020s,'' in \emph{Proceedings of the IEEE/CVF conference on computer
  vision and pattern recognition}, 2022, pp. 11\,976--11\,986.

\bibitem{vaswani2017attention}
A.~Vaswani, N.~Shazeer, N.~Parmar, J.~Uszkoreit, L.~Jones, A.~N. Gomez,
  {\L}.~Kaiser, and I.~Polosukhin, ``Attention is all you need,''
  \emph{Advances in neural information processing systems}, vol.~30, 2017.

\bibitem{dosovitskiy2015flownet}
A.~Dosovitskiy, P.~Fischer, E.~Ilg, P.~Hausser, C.~Hazirbas, V.~Golkov, P.~Van
  Der~Smagt, D.~Cremers, and T.~Brox, ``Flownet: Learning optical flow with
  convolutional networks,'' in \emph{Proceedings of the IEEE international
  conference on computer vision}, 2015, pp. 2758--2766.

\bibitem{ranftl2021vision}
R.~Ranftl, A.~Bochkovskiy, and V.~Koltun, ``Vision transformers for dense
  prediction,'' in \emph{Proceedings of the IEEE/CVF International Conference
  on Computer Vision}, 2021, pp. 12\,179--12\,188.

\bibitem{liu2021swin}
Z.~Liu, Y.~Lin, Y.~Cao, H.~Hu, Y.~Wei, Z.~Zhang, S.~Lin, and B.~Guo, ``Swin
  transformer: Hierarchical vision transformer using shifted windows,'' in
  \emph{Proceedings of the IEEE/CVF international conference on computer
  vision}, 2021, pp. 10\,012--10\,022.

\bibitem{wang2021pyramid}
W.~Wang, E.~Xie, X.~Li, D.-P. Fan, K.~Song, D.~Liang, T.~Lu, P.~Luo, and
  L.~Shao, ``Pyramid vision transformer: A versatile backbone for dense
  prediction without convolutions,'' in \emph{Proceedings of the IEEE/CVF
  international conference on computer vision}, 2021, pp. 568--578.

\bibitem{li2022next}
J.~Li, X.~Xia, W.~Li, H.~Li, X.~Wang, X.~Xiao, R.~Wang, M.~Zheng, and X.~Pan,
  ``Next-vit: Next generation vision transformer for efficient deployment in
  realistic industrial scenarios,'' \emph{arXiv preprint arXiv:2207.05501},
  2022.

\bibitem{johnson2016perceptual}
J.~Johnson, A.~Alahi, and L.~Fei-Fei, ``Perceptual losses for real-time style
  transfer and super-resolution,'' in \emph{Computer Vision--ECCV 2016: 14th
  European Conference, Amsterdam, The Netherlands, October 11-14, 2016,
  Proceedings, Part II 14}.\hskip 1em plus 0.5em minus 0.4em\relax Springer,
  2016, pp. 694--711.

\bibitem{huang2017arbitrary}
X.~Huang and S.~Belongie, ``Arbitrary style transfer in real-time with adaptive
  instance normalization,'' in \emph{Proceedings of the IEEE international
  conference on computer vision}, 2017, pp. 1501--1510.

\bibitem{islam2020much}
M.~A. Islam, S.~Jia, and N.~D. Bruce, ``How much position information do
  convolutional neural networks encode?'' \emph{arXiv preprint
  arXiv:2001.08248}, 2020.

\bibitem{ba2016layer}
J.~L. Ba, J.~R. Kiros, and G.~E. Hinton, ``Layer normalization,'' \emph{arXiv
  preprint arXiv:1607.06450}, 2016.

\bibitem{chollet2017xception}
F.~Chollet, ``Xception: Deep learning with depthwise separable convolutions,''
  in \emph{Proceedings of the IEEE conference on computer vision and pattern
  recognition}, 2017, pp. 1251--1258.

\bibitem{li2021localvit}
Y.~Li, K.~Zhang, J.~Cao, R.~Timofte, and L.~Van~Gool, ``Localvit: Bringing
  locality to vision transformers,'' \emph{arXiv preprint arXiv:2104.05707},
  2021.

\bibitem{lin2013network}
M.~Lin, Q.~Chen, and S.~Yan, ``Network in network,'' \emph{arXiv preprint
  arXiv:1312.4400}, 2013.

\bibitem{wang2004image}
Z.~Wang, A.~C. Bovik, H.~R. Sheikh, and E.~P. Simoncelli, ``Image quality
  assessment: from error visibility to structural similarity,'' \emph{IEEE
  transactions on image processing}, vol.~13, no.~4, pp. 600--612, 2004.

\bibitem{goodfellow2014generative}
I.~Goodfellow, J.~Pouget-Abadie, M.~Mirza, B.~Xu, D.~Warde-Farley, S.~Ozair,
  A.~Courville, and Y.~Bengio, ``Generative adversarial nets,'' \emph{Advances
  in neural information processing systems}, vol.~27, 2014.

\bibitem{eigen2015predicting}
D.~Eigen and R.~Fergus, ``Predicting depth, surface normals and semantic labels
  with a common multi-scale convolutional architecture,'' in \emph{Proceedings
  of the IEEE international conference on computer vision}, 2015, pp.
  2650--2658.

\bibitem{pontes2020scene}
J.~K. Pontes, J.~Hays, and S.~Lucey, ``Scene flow from point clouds with or
  without learning,'' in \emph{2020 international conference on 3D vision
  (3DV)}.\hskip 1em plus 0.5em minus 0.4em\relax IEEE, 2020, pp. 261--270.

\bibitem{caesar2020nuscenes}
H.~Caesar, V.~Bankiti, A.~H. Lang, S.~Vora, V.~E. Liong, Q.~Xu, A.~Krishnan,
  Y.~Pan, G.~Baldan, and O.~Beijbom, ``nuscenes: A multimodal dataset for
  autonomous driving,'' in \emph{Proceedings of the IEEE/CVF conference on
  computer vision and pattern recognition}, 2020, pp. 11\,621--11\,631.

\bibitem{liao2022kitti}
Y.~Liao, J.~Xie, and A.~Geiger, ``Kitti-360: A novel dataset and benchmarks for
  urban scene understanding in 2d and 3d,'' \emph{IEEE Transactions on Pattern
  Analysis and Machine Intelligence}, 2022.

\bibitem{hendrycks2016gaussian}
D.~Hendrycks and K.~Gimpel, ``Gaussian error linear units (gelus),''
  \emph{arXiv preprint arXiv:1606.08415}, 2016.

\bibitem{russakovsky2015imagenet}
O.~Russakovsky, J.~Deng, H.~Su, J.~Krause, S.~Satheesh, S.~Ma, Z.~Huang,
  A.~Karpathy, A.~Khosla, M.~Bernstein, \emph{et~al.}, ``Imagenet large scale
  visual recognition challenge,'' \emph{International journal of computer
  vision}, vol. 115, pp. 211--252, 2015.

\bibitem{kingma2014adam}
D.~P. Kingma and J.~Ba, ``Adam: A method for stochastic optimization,''
  \emph{arXiv preprint arXiv:1412.6980}, 2014.

\bibitem{teed2021droid}
Z.~Teed and J.~Deng, ``Droid-slam: Deep visual slam for monocular, stereo, and
  rgb-d cameras,'' \emph{Advances in neural information processing systems},
  vol.~34, pp. 16\,558--16\,569, 2021.

\bibitem{roessle2022dense}
B.~Roessle, J.~T. Barron, B.~Mildenhall, P.~P. Srinivasan, and M.~Nie{\ss}ner,
  ``Dense depth priors for neural radiance fields from sparse input views,'' in
  \emph{Proceedings of the IEEE/CVF Conference on Computer Vision and Pattern
  Recognition}, 2022, pp. 12\,892--12\,901.

\bibitem{gaidon2016virtual}
A.~Gaidon, Q.~Wang, Y.~Cabon, and E.~Vig, ``Virtual worlds as proxy for
  multi-object tracking analysis,'' in \emph{Proceedings of the IEEE conference
  on computer vision and pattern recognition}, 2016, pp. 4340--4349.

\bibitem{mohan2021efficientps}
R.~Mohan and A.~Valada, ``Efficientps: Efficient panoptic segmentation,''
  \emph{International Journal of Computer Vision}, vol. 129, no.~5, pp.
  1551--1579, 2021.

\bibitem{deng2022depth}
K.~Deng, A.~Liu, J.-Y. Zhu, and D.~Ramanan, ``Depth-supervised nerf: Fewer
  views and faster training for free,'' in \emph{Proceedings of the IEEE/CVF
  Conference on Computer Vision and Pattern Recognition}, 2022, pp.
  12\,882--12\,891.

\bibitem{yang2018deep}
N.~Yang, R.~Wang, J.~Stuckler, and D.~Cremers, ``Deep virtual stereo odometry:
  Leveraging deep depth prediction for monocular direct sparse odometry,'' in
  \emph{Proceedings of the European Conference on Computer Vision (ECCV)},
  2018, pp. 817--833.

\bibitem{yang2020d3vo}
N.~Yang, L.~v. Stumberg, R.~Wang, and D.~Cremers, ``D3vo: Deep depth, deep pose
  and deep uncertainty for monocular visual odometry,'' in \emph{Proceedings of
  the IEEE/CVF Conference on Computer Vision and Pattern Recognition}, 2020,
  pp. 1281--1292.

\bibitem{cvivsic2021recalibrating}
I.~Cvi{\v{s}}i{\'c}, I.~Markovi{\'c}, and I.~Petrovi{\'c}, ``Recalibrating the
  kitti dataset camera setup for improved odometry accuracy,'' in \emph{2021
  European Conference on Mobile Robots (ECMR)}.\hskip 1em plus 0.5em minus
  0.4em\relax IEEE, 2021, pp. 1--6.

\bibitem{peng2022semantic}
D.~Peng, Y.~Lei, M.~Hayat, Y.~Guo, and W.~Li, ``Semantic-aware domain
  generalized segmentation,'' in \emph{Proceedings of the IEEE/CVF Conference
  on Computer Vision and Pattern Recognition}, 2022, pp. 2594--2605.

\bibitem{zhao2023style}
Y.~Zhao, Z.~Zhong, N.~Zhao, N.~Sebe, and G.~H. Lee, ``Style-hallucinated dual
  consistency learning: A unified framework for visual domain generalization,''
  \emph{International Journal of Computer Vision}, pp. 1--17, 2023.

\bibitem{wang2020deep}
J.~Wang, K.~Sun, T.~Cheng, B.~Jiang, C.~Deng, Y.~Zhao, D.~Liu, Y.~Mu, M.~Tan,
  X.~Wang, \emph{et~al.}, ``Deep high-resolution representation learning for
  visual recognition,'' \emph{IEEE transactions on pattern analysis and machine
  intelligence}, vol.~43, no.~10, pp. 3349--3364, 2020.

\bibitem{xie2017aggregated}
S.~Xie, R.~Girshick, P.~Doll{\'a}r, Z.~Tu, and K.~He, ``Aggregated residual
  transformations for deep neural networks,'' in \emph{Proceedings of the IEEE
  conference on computer vision and pattern recognition}, 2017, pp. 1492--1500.

\bibitem{gasperini2023robust}
S.~Gasperini, N.~Morbitzer, H.~Jung, N.~Navab, and F.~Tombari, ``Robust
  monocular depth estimation under challenging conditions,'' in
  \emph{Proceedings of the IEEE/CVF international conference on computer
  vision}, 2023, pp. 8177--8186.

\end{thebibliography}
